# Autonomous Droplet Microfluidic Design Framework with Large Language Models

Dinh-Nguyen Nguyen[1], Raymond Kai-Yu Tong[1], Ngoc-Duy Dinh[1*]

**Abstract**

Droplet-based microfluidic devices have substantial promise as cost-effective alternatives to current assessment tools in biological research. Moreover, machine learning models that leverage tabular data, including input design parameters and their corresponding efficiency outputs, are increasingly utilised to automate the design process of these devices and to predict their performance. However, these models fail to fully leverage the data presented in the tables, neglecting crucial contextual information, including column headings and their associated descriptions. This study presents µ-Fluidic-LLMs, a framework designed for processing and feature extraction, which effectively captures contextual information from tabular data formats. µ-Fluidic-LLMs overcomes processing challenges by transforming the content into a linguistic format and leveraging pre-trained large language models (LLMs) for analysis. We evaluate our µ-Fluidic-LLMs framework on 11 prediction tasks, covering aspects such as geometry, flow conditions, regimes, and performance, utilising a publicly available dataset on flow-focusing droplet microfluidics. We demonstrate that our µ-Fluidic-LLMs framework can empower deep neural network models to be highly effective and straightforward while minimising the need for extensive data preprocessing. Moreover, the exceptional performance of deep neural network models, particularly when combined with advanced natural language processing models such as DistilBERT and GPT-2, *reduces the mean absolute error in the droplet diameter and generation rate by nearly 5- and 7-fold*, respectively, and enhances the regime classification accuracy by *over 4%*, compared with the performance reported in a previous study. This study lays the foundation for the huge potential applications of LLMs and machine learning in a wider spectrum of microfluidic applications.

[1]Department of Biomedical Engineering, The Chinese University of Hong Kong, Shatin, N.T., Hong Kong, China

*Corresponding Author, Corresponding Email: ngocduydinh@cuhk.edu.hk, dinhngocduy@u.nus.edu

# 1. Introduction

Droplet-based microfluidics has been recognised as a groundbreaking technology for miniaturising biological and chemical experiments. It has significantly advanced biotechnology[1–5] by enabling techniques such as next-generation sequencing,[6–8] single-cell RNA sequencing,[8–10] droplet digital PCR,[11–14] and liquid biopsies diagnostics.[15] However, the impact of microfluidics remains largely confined to single-use cartridges, integrated bench-top devices, and specialised lab setups.[16,17] In addition, the complex design and fabrication of custom microfluidic devices have limited their widespread adoption and general use.[18,19] Moreover, microfluidic design and operation can take months or years of iterative testing to optimise, even if fabrication is outsourced at a high cost.[20] To overcome these limitations, machine learning, which predicts patterns and behaviour, has been employed.

Machine learning models are becoming more popular for predicting performance and automating design in microfluidic droplet generation. For instance, Mahdi *et al.*[21] used machine learning to predict the size of water droplets in glycerine oil from a T-junction setup. The model, trained with 742 data points, accurately predicted droplet size using Reynolds and capillary numbers across different flow rates and fluid properties within one geometry. Furthermore, in Lashkaripour *et al.*,[22] neural networks were used to predict droplet size, generation rate, and flow regime based on design geometry and flow conditions. The neural networks, which were trained on 888 data points with varying capillary numbers, flow rate ratio, and six geometric parameters, accurately predicted the droplet generation regime (95.1% accuracy), size (error < 10 μm), and generation rate (error < 20 Hz) for droplets ranging from 25 to 250 μm in size and 5 to 500 Hz in rate. Elsewhere, in Damiati *et al.*,[23] a machine learning model predicted the size of poly(lactic-co-glycolic acid) (PLGA) microparticles produced by flow-focusing droplet generators and dichloromethane solvent evaporation. The model was trained on data from 223 combinations of flow rates, PLGA concentrations, device types, and sizes to predict PLGA particle size ($R^2$ > 0.94) accurately. Furthermore, Hong *et al.*[24] applied machine learning models to automate the design of concentration gradient generators. A neural network trained on 9 million data points from a verified physics model was able to map desired concentration profiles to inlet settings, achieving an 8.5% error rate. Meanwhile, *Ji et al.*[25] applied machine learning to automate the iterative design of grid micromixers. Neural networks were trained on 4,320 simulated chips to map channel lengths to output concentrations. The designs met outlet concentration targets within 0.01 mol/m³, compared with simulations, for 91.5% of benchmarks. However, these machine learning models are limited to processing the explicit content within tables, without considering the surrounding contextual information, such as column headers and accompanying descriptions. Furthermore, the data processing becomes more complex when inconsistencies in units of measurement or data types are present across varying tabular data systems. These challenges could be mitigated by harnessing the power of language-based approaches, because language is a highly versatile data modality capable of representing information across diverse data points without the need for structural consistency. Additionally, recent advancements in large language models (LLMs) utilising Transformer architecture[26] have enabled state-of-the-art performance across a diverse array of language tasks, including translation, sentence completion, and question answering. These pre-trained models are frequently built using vast and diverse datasets, which allows them to leverage prior knowledge and make accurate predictions even with minimal training data. Some LLMs are specifically trained to address domain-specific knowledge and technical complexities, enhancing their utility in relevant applications. For instance, LLMs including ClinicalBERT,[27] BioBERT,[28] and BioGPT[29] have been used to fine-tune medical records and biomedical datasets, providing substantial benefits for healthcare-related tasks.

In this study, we establish *µ-Fluidic-LLMs*, a systematic framework that utilises linguistic techniques to derive contextual information related to droplet microfluidic design parameters within tabular structures. This approach produces more comprehensive data representations. We handle tabular data by converting each data sample into a corresponding textual representation. This representation is integrated into the column attributes with their respective values while incorporating any relevant contextual information that may be present. Subsequently, the complete text is fed into a pre-trained LLM to generate embeddings with a consistent fixed dimensionality. These embeddings are then utilised as input for standard machine learning models to perform downstream tasks. Then, we evaluate the effect on final task performance by comparing the performance when inputting the embeddings into standard machine learning models against the performance when inputting the original tabular features into the same models. Finally, we demonstrate that integrating deep neural networks (DNNs) with LLMs substantially enhances performance on downstream tasks. We also demonstrate the adaptability of this framework, enabling it to incorporate new LLMs as they become available. An overview of this study is illustrated in Fig. 1.

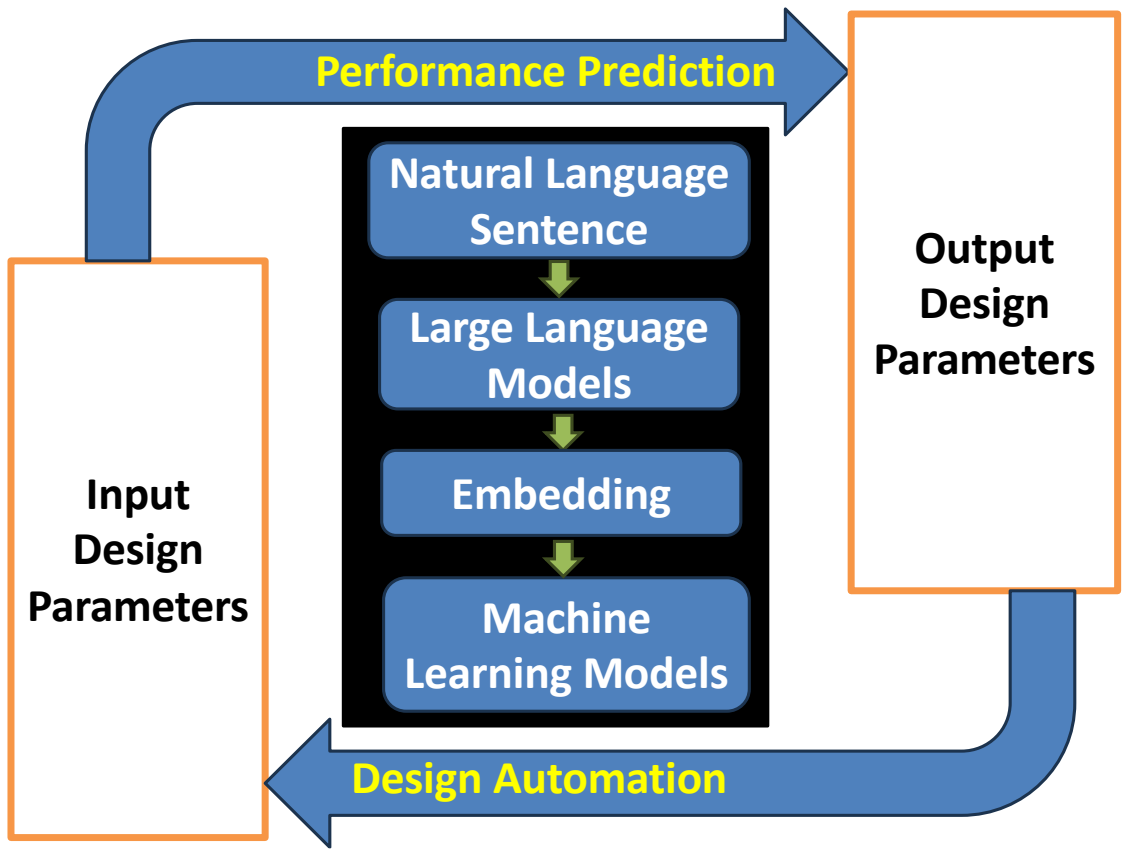

Fig. 1 μ-Fluidic-LLMs framework. Regarding performance prediction, μ-Fluidic-LLMs transforms input design parameters from tabular structures into natural language sentences, which are then processed by LLMs to generate embeddings. These embeddings are subsequently used as input for traditional machine learning models. In the context of design automation, μ-Fluidic-LLMs follows the same steps; however, the output design parameters serve as the input parameters.

**2. Experimental**

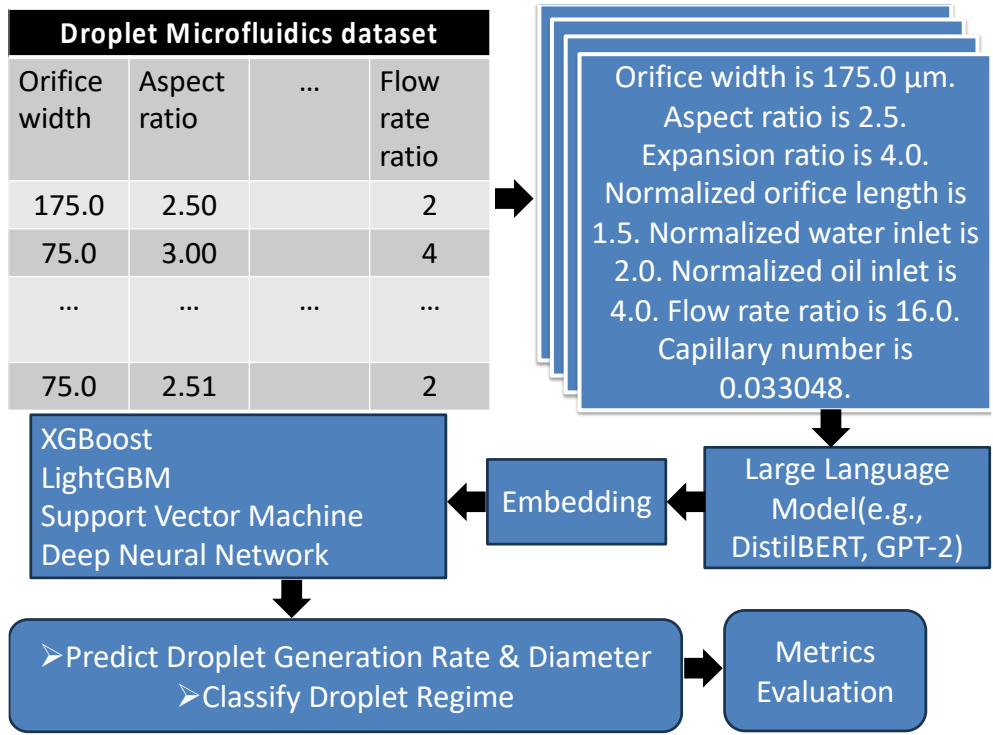

(A) Performance Prediction

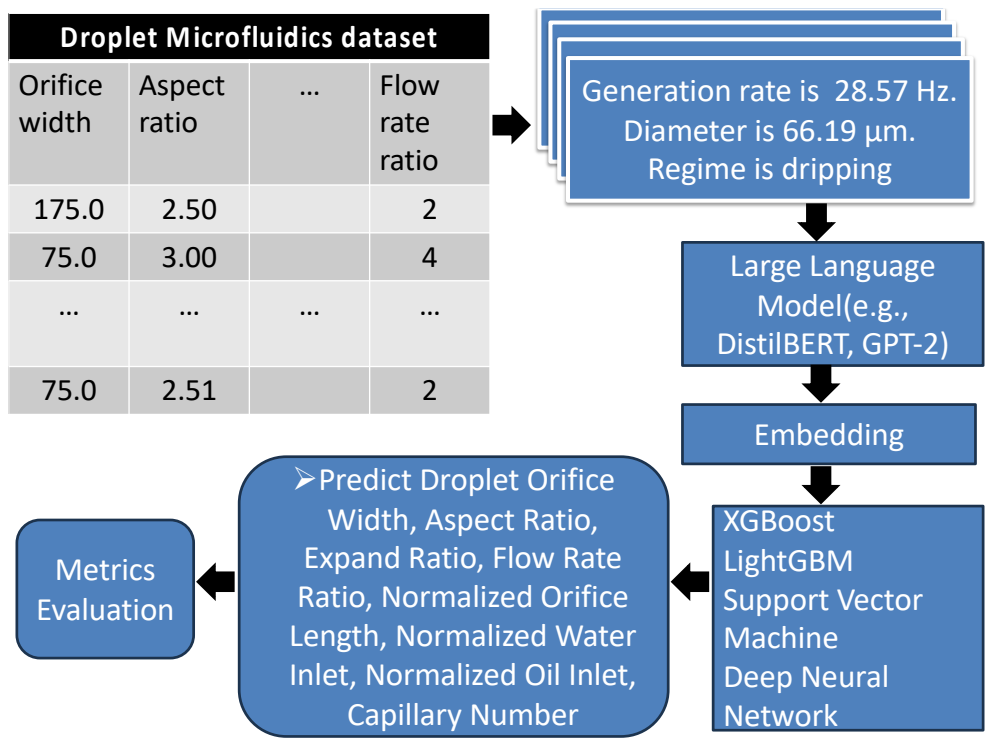

(B) Design Automation

Fig. 2 Summary of the overall methodology. Fig. 2A illustrates the process of converting each row of tabular data into a paragraph that includes the eight column headers and their corresponding values. This paragraph is then input into large language models to generate embeddings, which are subsequently used in machine learning models for performance prediction. Fig. 2B demonstrates a similar procedure for design automation, with the paragraph containing only three column headers and their corresponding values.

## 2.1 Data and Tasks

We utilised a dataset on flow-focusing droplet microfluidics,[22] structured in tabular form and comprising 998 data points. This dataset includes 11 features: orifice width, normalised orifice length, normalised water inlet width, normalised oil inlet width, expansion ratio, aspect ratio, flow rate ratio, capillary number, droplet diameter, generation rate, and regime. Using this dataset, we conducted two regression tasks and one classification task focused on performance prediction, as well as eight regression tasks related to design automation. For the design automation aspect, we present the results of one regression task, with the remaining results in the supplementary file.

## 2.2 Text Serialisation for Performance Prediction and Design Automation

Text serialisation is the process of converting structured data into a linear, text-based format that can be easily stored, transmitted, and reconstructed. In LLMs, text serialisation plays a vital role in processing and managing vast amounts of data[30]. These models rely on serialised text data to train on diverse datasets, often including a mix of natural language, structured information, and metadata. By converting complex input data into a serialised text format, LLMs can uniformly process this information, enabling them to generate meaningful predictions, responses, or completions based on the input. Furthermore, the serialised text is essential for fine-tuning models with specific data, allowing the LLMs to adapt to new tasks or domains by processing serialised input and output pairs.[31,32] Our framework implements manual text serialisation for individual data samples, as shown in Fig. 2. By leveraging the column headers of tabular data, the framework systematically transforms each table row into a sequence of key–value pairs, with the keys originating from the column headers and the values corresponding to the respective data entries. For performance prediction, as shown in Fig. 2A, the eight column headers and their corresponding values are transformed into a paragraph. This text is then input into pre-trained LLMs to generate embeddings. These embeddings are subsequently used as input for baseline machine learning models, which predict output values such as droplet diameter, generation rate, and regime. For design automation, as illustrated in Fig. 2B, the three column headers droplet diameter, generation rate, and regime, along with their respective values, are transformed into a paragraph. This paragraph is processed by pre-trained LLMs to produce embeddings, which are then utilised as inputs for the baseline machine learning models to predict the remaining eight design parameters. This strategy is designed to maintain both the integrity and the contextual meaning of the original tabular data during serialisation.

**2.3 Text Embeddings**

Text embedding is a technique used to convert textual data into a dense, fixed-size vector representation. These vectors capture the semantic meaning of the text by mapping words, phrases, or entire documents into a continuous vector space, where similar pieces of text are represented by vectors that are close to each other. The primary advantage of text embedding lies in its ability to represent complex linguistic relationships and contextual meanings in a numerical format that can be efficiently processed by machine learning algorithms. This makes text embedding an essential component in various natural language processing (NLP) tasks, such as sentiment analysis, text classification, and machine translation.[33]

**2.4 Large Language Model Selection**

LLMs play a pivotal role in generating high-quality text embeddings by leveraging their understanding of linguistic patterns and contextual relationships. Pre-trained language models, such as BERT,[34] GPT-2,[35] and their variants, are particularly effective in producing embeddings because they are trained on vast text corpora and can capture nuanced semantic information. These models transform input text into embeddings by processing it through multiple layers of neural networks, where each layer refines the representation by focusing on different aspects of the language, such as syntax, semantics, and context. The resulting embeddings are highly informative and can be used as input for downstream NLP tasks, enhancing the performance of models in applications like question-answering document retrieval and text summarisation.

In our study, it was essential to carefully select the language model backbones that would be employed. While there are many potential options, we focused on DistilBERT,[36]

SentenceTransformer, and GPT-2 for generating text embeddings in downstream tasks. DistilBERT, a more compact version of BERT, is particularly suitable for scenarios where computational efficiency is paramount, as it effectively balances speed and accuracy. SentenceTransformer[37] is specifically designed for creating sentence-level embeddings, making it ideal for tasks such as semantic similarity, clustering, and information retrieval. This model builds upon the BERT architecture, which has been fine-tuned for these particular applications. Finally, GPT-2, although more computationally demanding, produces contextually rich embeddings by considering a broader contextual scope. This makes it a robust choice for tasks that require deep comprehension and generative capabilities, such as text generation and complex language understanding. The choice of these three language models highlights the versatility of our framework, as it can be adapted to integrate any available language model.

**2.5 Baseline Machine Learning Models**

To assess the effectiveness of text embeddings compared with non-text embeddings in downstream tasks, we employed a range of standard machine learning models, including XGBoost,[38] LightGBM,[39] and support vector machine (SVM).[40] Additionally, DNNs, an advanced form of traditional neural networks characterised by the addition of multiple hidden layers,[41] were designed and applied in this study. Subsequently, we optimised the hyperparameters for all baseline models, as well as for their respective combinations with pre-trained language models.

**2.6 Evaluation Metrics**

After finalising the models, we performed 10-fold cross-validation 15 times to obtain more robust and reliable mean performance estimates, thereby reducing the variability introduced by random data partitioning.[42] For regression tasks, the evaluation utilised four metrics: mean absolute error (MAE), mean squared error (MSE), root mean squared error (RMSE), and the coefficient of determination ($R^2$), along with their associated standard errors. For the classification task, five metrics were employed: accuracy, F1 score, precision, recall, and area under the receiver operating characteristic curve (ROC AUC), each accompanied by standard errors. Here, we present line graphs comparing MAE across models for regression and accuracy across models for classification. Line graphs for the remaining metrics are included in the supplementary file. In addition, we provide tables in the 'Results & discussion' section demonstrating the minimisation of standard error and the stabilisation of mean estimated performance for metrics across different models. Specifically, the values presented in the table are those obtained from repetitions in which the discrepancy between the mean and median of the corresponding metric is minimised, indicating that the data likely follows a normal distribution.[42,43]

## 3. Results & discussion

To verify the effectiveness of our μ-Fluidic-LLMs framework, we highlight selected results in this section, with the remaining findings provided in the supplementary file.

### 3.1 Enhanced Efficiency for Performance Prediction

#### 3.1.1 Performance in Prediction of Droplet Diameter

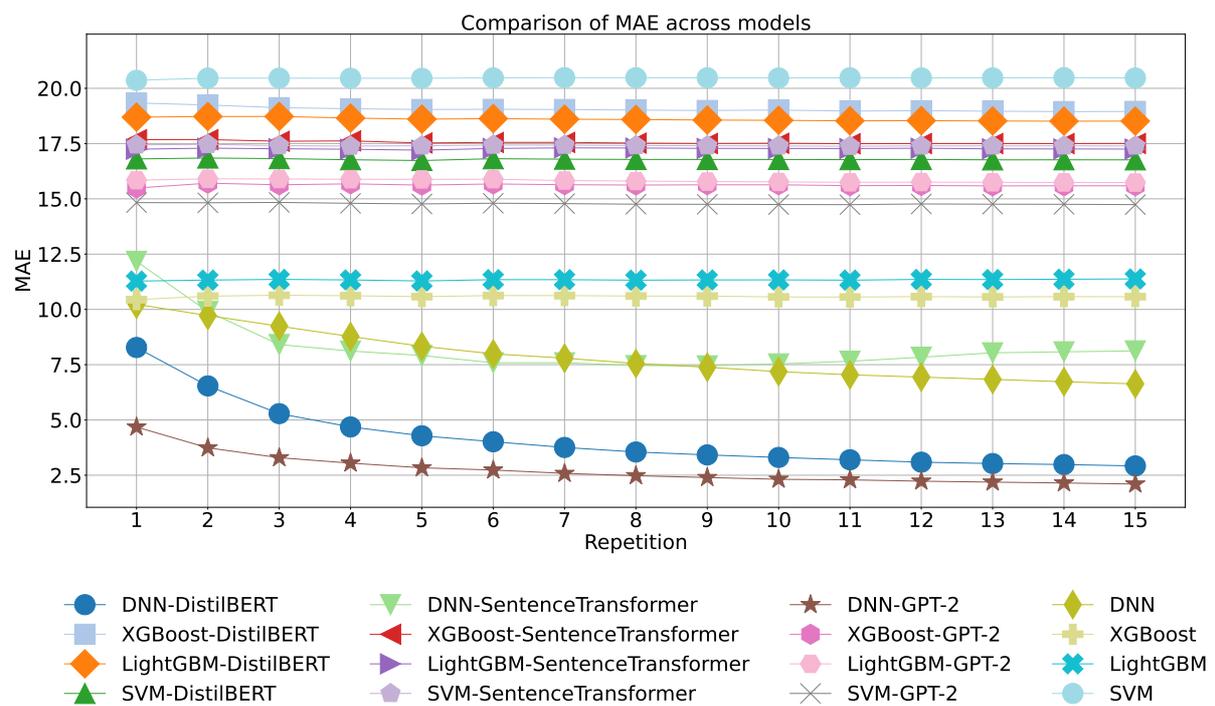

Fig. 3 Comparison of mean absolute error (MAE) across models for droplet diameter

To assess the trends in accuracy in droplet diameter prediction across different models, a detailed comparison of 16 models over 15 repetitions was conducted, using MAE as the performance metric, as illustrated in Fig. 3. DNNs integrated with DistilBERT and GPT-2 exhibited the most substantial improvement, with MAE decreasing from approximately 7.5 to around 2.5, demonstrating their adaptability and accuracy. In contrast, XGBoost and LightGBM, when paired with DistilBERT, consistently displayed high MAE values, approaching 20, indicating persistent inefficiencies and minimal improvement in performance. However, XGBoost and LightGBM achieved moderate performance, with MAE stabilising between 10 and 12.5 after initial reductions. This variability in model performance highlights the critical need for iterative testing and careful model selection to optimise predictive accuracy and reliability. Across the remaining metrics, the combination of DNNs with DistilBERT and GPT-2 consistently showed the best performance (see Supplementary Figs. 1, 2, and 3).

| Droplet Diameter (µm) | | | | |
|---|---|---|---|---|
| Model | Metrics | | | |
| | MAE | MSE | RMSE | $R^2$ |
| DNN | 6.9375 ± 0.1782 | 98.6397 ± 6.5203 | 10.4152 ± 0.3664 | 0.9675 ± 0.0025 |
| DNN–DistilBERT | 2.9814 ± 0.1623 | 22.6262 ± 3.5155 | 3.9286 ± 0.219 | 0.9619 ± 0.0063 |
| DNN–GPT-2 | 2.3941 ± 0.1159 | 12.8634 ± 1.3313 | 3.512 ± 0.1573 | 0.9966 ± 0.0004 |
| DNN–SentenceTransformer | 12.1736 ± 1.147 | 98.5915 ± 6.2758 | 15.112 ± 1.384 | 0.978 ± 0.0023 |
| LightGBM | 11.344 ± 0.217 | 399.2599 ± 13.8805 | 19.489 ± 0.5123 | 0.9019 ± 0.0031 |
| LightGBM–DistilBERT | 18.6042 ± 0.3312 | 835.1644 ± 52.8888 | 28.1189 ± 0.3305 | 0.8028 ± 0.0032 |
| LightGBM–GPT-2 | 15.8265 ± 0.2479 | 642.6034 ± 17.0149 | 25.0282 ± 0.3484 | 0.8436 ± 0.0034 |
| LightGBM–SentenceTransformer | 17.2428 ± 0.3663 | 752.7167 ± 20.0303 | 27.1305 ± 0.3931 | 0.8161 ± 0.0033 |
| SVM | 20.4542 ± 0.3696 | 1136.2003 ± 35.1405 | 33.309 ± 0.4373 | 0.7203 ± 0.0058 |
| SVM–DistilBERT | 16.8047 ± 0.9945 | 851.9856 ± 34.0729 | 28.7854 ± 0.578 | 0.7916 ± 0.0059 |
| SVM–GPT-2 | 14.7982 ± 0.3302 | 649.6983 ± 18.0926 | 25.1214 ± 0.368 | 0.8405 ± 0.0058 |
| SVM–SentenceTransformer | 17.3951 ± 0.2184 | 908.6298 ± 23.5036 | 29.7663 ± 0.3882 | 0.7779 ± 0.0069 |
| XGBoost | 10.6225 ± 0.1797 | 307.5357 ± 13.8743 | 17.3687 ± 0.3621 | 0.9233 ± 0.0019 |
| XGBoost–DistilBERT | 19.0411 ± 0.2699 | 742.1614 ± 17.4954 | 27.1867 ± 0.4374 | 0.8138 ± 0.0047 |
| XGBoost–GPT-2 | 15.6338 ± 0.1782 | 545.6909 ± 12.6868 | 23.1538 ± 0.2716 | 0.862 ± 0.0046 |
| XGBoost–SentenceTransformer | 17.5008 ± 0.1626 | 679.0781 ± 13.921 | 25.8538 ± 0.2868 | 0.8298 ± 0.0044 |

Table 1. Metrics of evaluation for droplet diameter prediction, compared across models

To demonstrate the minimal standard error and the stabilisation of mean estimated performance across metrics of droplet diameter prediction for different models, a comprehensive evaluation of various machine learning models combined with language models is provided. This evaluation focuses on MAE, MSE, RMSE, and $R^2$, along with their corresponding standard errors, as shown in Table 1. The table presents values derived from the repetitions in which the discrepancy between the mean and median of the respective metric was minimised. The models assessed include DNNs, LightGBM, SVM, and XGBoost, each tested with DistilBERT, GPT-2, and SentenceTransformer. DNN paired with GPT-2 emerged as the most effective model across all metrics. It achieved the lowest MAE of 2.3941 – approximately a factor of five smaller than the MAE of 10 reported in a previous study[22] – an MSE of 12.8634, and RMSE of 3.512, indicating minimal prediction error. Moreover, it recorded the highest R² value of 0.9966, demonstrating an exceptionally strong correlation between predicted and actual values, with a very low standard error of 0.0004, underscoring its reliability and stability. This combination significantly outperformed other models, highlighting the potential of GPT-2 in enhancing predictive accuracy when integrated with DNNs. In contrast, SVM models, particularly when combined with SentenceTransformer and DistilBERT, displayed the weakest performance. The SVM model without any language model integration showed a notably high MAE of 20.4542 and MSE of 1136.2003, alongside a low R² of 0.7203. Even with DistilBERT, the SVM model's performance remained suboptimal, exhibiting considerable prediction errors. This suggests that SVM might not be as well-suited for the types of tasks or data considered in this evaluation, especially compared with other model and language model combinations. Another critical observation is the performance variability of LightGBM and XGBoost models when integrated with different language models. LightGBM showed poor performance with DistilBERT (MAE of 18.6042 and MSE of 835.1644) but significantly better results with GPT-2 (MAE of 15.8265 and MSE of 642.6034), although still not reaching the efficacy of DNN–GPT-2. XGBoost, while generally outperforming LightGBM and SVM, still lagged behind DNN in accuracy, particularly in its combination with SentenceTransformer, where it registered an RMSE of 25.8538 and an R² of 0.8298.

**3.1.2 Performance in Prediction of Droplet Generation Rate**

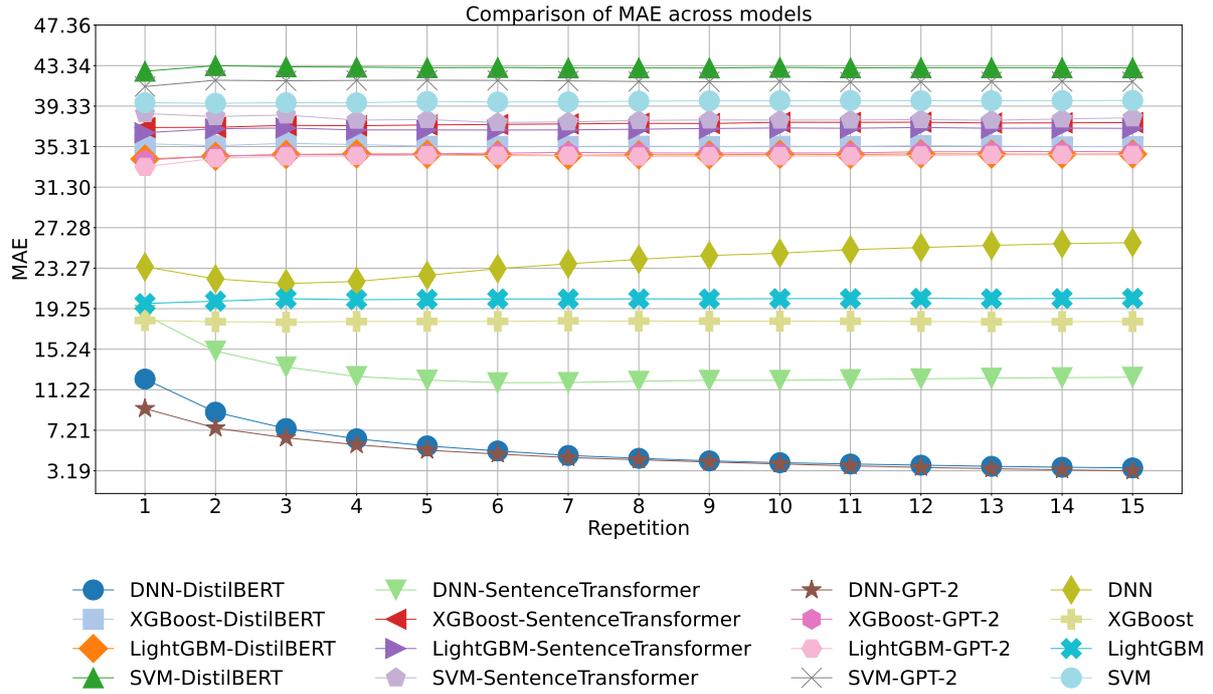

Fig. 4 Comparison of MAE across models for droplet generation rate

To evaluate the trends in accuracy in droplet generation rate prediction across various models, a thorough comparison of the 16 models over 15 repetitions was carried out, utilising MAE as the performance metric, as depicted in Fig. 4. DNNs combined with DistilBERT and GPT-2 demonstrated outstanding performance, with MAE values of around 3, indicating high accuracy and effective learning. This reflects robust model architecture and well-optimised hyperparameters, which contribute to their stability and precision. In contrast, XGBoost, LightGBM, and SVM, when paired with GPT-2, consistently exhibited MAE values above 35, indicating significant limitations in predictive capability. DNN paired with SentenceTransformer, XGBoost paired with SentenceTransformer, and standalone DNN models showed moderate performance, with MAE values between 12 and 25. Across the remaining metrics, DNN integrated with DistilBERT and GPT-2 consistently outperformed other models, showing significant improvements (see Supplementary Figs. 4, 5, and 6).

| Droplet Generation Rate (Hz) | | | | |
|---|---|---|---|---|
| Model | Metrics | | | |
| | MAE | MSE | RMSE | R² |
| DNN | 25.3095 ± 0.4555 | 1624.4079 ± 139.5017 | 44.6364 ± 1.1816 | 0.8718 ± 0.0041 |
| DNN–DistilBERT | 3.478 ± 0.2272 | 49.2167 ± 11.2906 | 20.5089 ± 2.09 | 0.9969 ± 0.0008 |
| DNN–GPT-2 | 3.1921 ± 0.178 | 39.8099 ± 7.8973 | 6.2534 ± 0.4721 | 0.9975 ± 0.0005 |
| DNN–SentenceTransformer | 12.4193 ± 0.3476 | 509.0871 ± 40.3895 | 21.3771 ± 0.8071 | 0.968 ± 0.0015 |
| LightGBM | 20.1445 ± 0.5134 | 1559.2185 ± 96.3836 | 38.9632 ± 0.6701 | 0.9089 ± 0.0089 |
| LightGBM–DistilBERT | 34.531 ± 1.023 | 4156.6292 ± 337.1268 | 63.9626 ± 1.2642 | 0.747 ± 0.016 |
| LightGBM–GPT-2 | 33.3094 ± 0.72 | 3954.8462 ± 142.4064 | 61.8214 ± 1.1531 | 0.7634 ± 0.0055 |
| LightGBM–SentenceTransformer | 36.9546 ± 0.7657 | 5373.4073 ± 253.7448 | 71.6789 ± 2.0648 | 0.6851 ± 0.0066 |
| SVM | 39.6055 ± 0.9626 | 4613.4453 ± 131.3044 | 67.0637 ± 1.2258 | 0.7132 ± 0.035 |
| SVM–DistilBERT | 42.7807 ± 1.2286 | 7303.8329 ± 477.9589 | 84.2248 ± 1.8198 | 0.5687 ± 0.0102 |
| SVM–GPT-2 | 41.7472 ± 0.5623 | 6699.5807 ± 226.8325 | 80.7062 ± 1.8152 | 0.6032 ± 0.0061 |
| SVM–SentenceTransformer | 38.0529 ± 0.4146 | 4218.9963 ± 337.7307 | 63.8702 ± 2.6419 | 0.7398 ± 0.0299 |
| XGBoost | 17.9172 ± 0.3871 | 1105.9582 ± 56.9718 | 32.3535 ± 0.5636 | 0.9325 ± 0.002 |
| XGBoost–DistilBERT | 35.2952 ± 0.4083 | 3444.8867 ± 100.9941 | 57.8354 ± 0.845 | 0.7891 ± 0.006 |
| XGBoost–GPT-2 | 34.3704 ± 0.9647 | 3319.7046 ± 113.926 | 57.1386 ± 0.7216 | 0.792 ± 0.0052 |
| XGBoost–SentenceTransformer | 37.622 ± 0.6225 | 4099.8887 ± 359.9063 | 63.5527 ± 1.0436 | 0.7529 ± 0.0169 |

Table 2. Metrics of evaluation for droplet generation rate, compared across models

To illustrate the minimal standard error and the stabilisation of mean estimated performance across metrics of droplet generation rate prediction for different models, we present a comprehensive evaluation of the various machine learning models integrated with LLMs. This assessment again uses MAE, MSE, RMSE, and R², along with their associated standard errors, as presented in Table 2. The table displays values obtained from the repetitions in which the difference between the mean and median of the corresponding metric was minimised. The DNN models integrated with NLP models—particularly DistilBERT and GPT-2—consistently outperformed other models, achieving significantly lower error metrics and higher R² values. For example, the DNN–GPT-2 model exhibited an MAE of 3.1921, roughly a factor of seven smaller than the MAE of 20 reported in a previous study,[22] and an R² of 0.9975, indicating an exceptional fit to the data with minimal prediction error. This strong performance underscores the effectiveness of combining deep learning with advanced NLP models in enhancing predictive accuracy. In contrast, traditional machine learning models such as SVM and LightGBM showed substantially higher error rates and lower R² values, particularly when combined with NLP models. For instance, the SVM–DistilBERT model, with an MAE of 42.7807 and an R² of 0.5687, demonstrated inadequate predictive power, suggesting that the integration of DistilBERT with SVM does not yield substantial improvements. Similarly, the LightGBM–SentenceTransformer combination resulted in a high MAE of 36.9546 and a relatively low R² of 0.6851, reflecting its limited efficacy. The analysis also reveals notable inconsistencies in the performance of models when combined with NLP techniques. While DNN-based models saw significant improvements when integrated with NLP models, traditional models like XGBoost and SVM did not consistently benefit from such integration. For instance, the XGBoost–GPT-2 combination, despite achieving a respectable R² of 0.792, showed only marginal improvement over XGBoost alone, indicating that the choice of model architecture plays a crucial role in determining the success of NLP integration.

### 3.1.3 Performance in Prediction of Droplet Regime

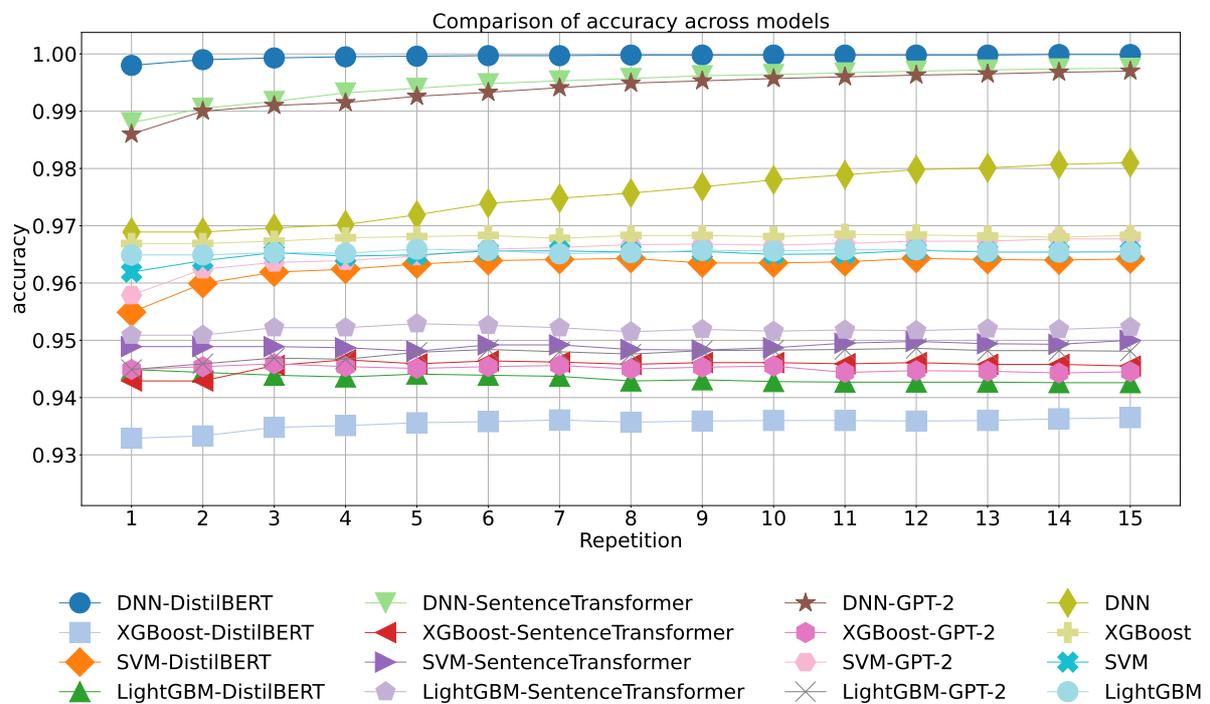

Fig. 5 Comparison of accuracy across models for droplet regime

To evaluate the trends in accuracy in droplet regime predictions across various models, a detailed comparison of 16 models across 15 repetitions was performed, with accuracy serving as the primary performance metric, as shown in Fig. 5. DNNs paired with DistilBERT, GPT-2, and SentenceTransformer consistently demonstrated exceptional accuracy, with values approaching 1.00. This indicates that these models are highly effective, likely due to advanced algorithms or well-optimised training processes. Their consistent performance across repetitions suggests robustness and reliability, making them well-suited for tasks requiring high precision. In contrast, XGBoost paired with DistilBERT consistently showed lower accuracy, around 0.93. Meanwhile, DNN and LightGBM paired with SentenceTransformer, SVM paired with DistilBERT, and XGBoost exhibited moderate accuracy, generally ranging from 0.95 to 0.98, with slight improvements observed over repetitions. Across the remaining metrics, the combination of DNN and DistilBERT consistently outperformed other models, exhibiting superior predictive performance (see Supplementary Figs. 7, 8, 9, and 10).

| Droplet Regime | | | | | |
|---|---|---|---|---|---|
| Model | Metrics | | | | |
| | Accuracy | F1 Score | Precision | Recall | ROC AUC |
| DNN | 0.9801 ± 0.0014 | 0.9778 ± 0.0016 | 0.9778 ± 0.0024 | 0.9756 ± 0.0022 | 0.9682 ± 0.0034 |
| DNN–DistilBERT | 0.9999 ± 0.0001 | 0.9998 ± 0.0002 | 1.0 ± 0.0 | 0.9997 ± 0.0002 | 0.9998 ± 0.0002 |
| DNN–GPT-2 | 0.99 ± 0.0022 | 0.9886 ± 0.0025 | 0.9969 ± 0.0007 | 0.9964 ± 0.0009 | 0.9908 ± 0.0019 |
| DNN–SentenceTransformer | 0.9905 ± 0.0024 | 0.9892 ± 0.0027 | 0.9975 ± 0.0007 | 0.997 ± 0.0007 | 0.9904 ± 0.0024 |
| LightGBM | 0.9653 ± 0.0027 | 0.9594 ± 0.0033 | 0.9767 ± 0.0026 | 0.9441 ± 0.0055 | 0.9628 ± 0.003 |
| LightGBM–DistilBERT | 0.9444 ± 0.0051 | 0.9337 ± 0.0021 | 0.9419 ± 0.004 | 0.9303 ± 0.0055 | 0.9422 ± 0.0029 |
| LightGBM–GPT-2 | 0.9449 ± 0.007 | 0.9368 ± 0.0079 | 0.95 ± 0.0028 | 0.9317 ± 0.0115 | 0.9459 ± 0.0027 |
| LightGBM–SentenceTransformer | 0.9509 ± 0.0046 | 0.9438 ± 0.0026 | 0.9572 ± 0.0065 | 0.9318 ± 0.0042 | 0.9504 ± 0.0031 |
| SVM | 0.965 ± 0.0017 | 0.9599 ± 0.0026 | 0.9761 ± 0.0034 | 0.947 ± 0.0029 | 0.9635 ± 0.0023 |
| SVM–DistilBERT | 0.9549 ± 0.0061 | 0.9545 ± 0.0049 | 0.9546 ± 0.005 | 0.9566 ± 0.0056 | 0.9596 ± 0.0042 |
| SVM–GPT-2 | 0.9647 ± 0.0023 | 0.9597 ± 0.0026 | 0.9567 ± 0.0075 | 0.9554 ± 0.0059 | 0.9642 ± 0.0024 |
| SVM–SentenceTransformer | 0.95 ± 0.0016 | 0.9432 ± 0.0019 | 0.9377 ± 0.0032 | 0.9453 ± 0.0117 | 0.948 ± 0.0026 |
| XGBoost | 0.9685 ± 0.0017 | 0.9633 ± 0.002 | 0.977 ± 0.003 | 0.9521 ± 0.0094 | 0.9657 ± 0.0023 |
| XGBoost–DistilBERT | 0.9351 ± 0.0034 | 0.9247 ± 0.0041 | 0.9384 ± 0.0043 | 0.9072 ± 0.0111 | 0.9327 ± 0.0036 |
| XGBoost–GPT-2 | 0.9449 ± 0.0072 | 0.937 ± 0.0028 | 0.9434 ± 0.0042 | 0.9317 ± 0.0143 | 0.944 ± 0.0029 |
| XGBoost–SentenceTransformer | 0.9429 ± 0.0057 | 0.9339 ± 0.0066 | 0.9536 ± 0.003 | 0.9224 ± 0.0036 | 0.9407 ± 0.006 |

Table 3. Metrics of evaluation for droplet regime, compared across models

To demonstrate the minimal standard error and the stabilisation of mean estimated performance across metrics of droplet regime prediction for various models, we present an extensive comparative analysis of DNNs, LightGBM, SVM, and XGBoost. The performance of each model was evaluated when integrated with advanced NLP models, as outlined in Table 3. Key metrics of accuracy, F1 score, precision, recall, and ROC AUC are employed to examine the efficacy of these models. The table reports values obtained from the repetitions in which the difference between the mean and median of the corresponding metric was minimised. DNN models, particularly when combined with DistilBERT, GPT-2, and SentenceTransformer, demonstrated superior performance across all metrics. Notably, the DNN–DistilBERT model achieved an exceptional accuracy of 0.9999 – representing an improvement of more than 4% over the accuracy of 0.951 reported in a previous study[22] – and an F1 score of 0.9998 and precision of 1.0, underscoring its exceptional ability to correctly classify instances with minimal error. This model also achieved a recall of 0.9997, reflecting its high sensitivity in detecting true positives. In contrast, other model combinations, such as LightGBM with DistilBERT and GPT-2, underperformed relative to DNN-based models. For instance, LightGBM–DistilBERT achieved a lower accuracy of 0.9444 and F1 score of 0.9337, indicating weaker overall performance and a greater likelihood of misclassifications. Similarly, SVM models paired with various NLP techniques exhibited varying degrees of performance, with the SVM–DistilBERT combination showing an accuracy of 0.9549 and an F1 score of 0.9545, but with increased variability in precision and recall, suggesting instability in classification consistency. XGBoost, although traditionally strong in structured data scenarios, did not perform as well when integrated with these NLP models, particularly in comparison with DNN-based approaches. For example, XGBoost–DistilBERT yielded an accuracy of 0.9351 and an F1 score of 0.9247, lower than the best-performing DNN combinations. Additionally, the larger standard errors observed in models like LightGBM–GPT-2 (e.g., recall standard error of 0.0115) indicate higher uncertainty.

**3.2 Enhanced Efficiency for Design Automation**

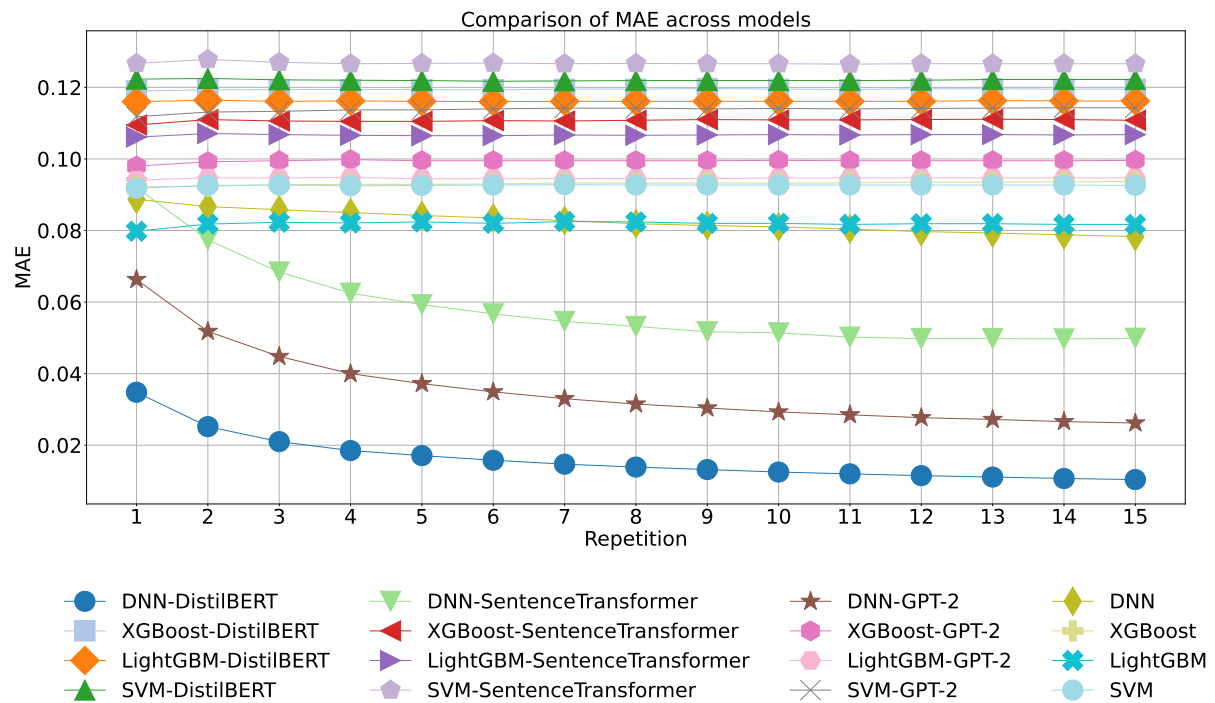

Fig. 6 Comparison of MAE across models for droplet capillary number

To analyse the trends in accuracy in droplet capillary number predictions across various models, a detailed comparison of 16 models was conducted over 15 repetitions, employing MAE as the performance metric, as shown in Fig. 6. The DNN paired with DistilBERT exhibited exceptional accuracy, with MAE decreasing significantly from approximately 0.04 to below 0.02, indicating robust learning and adaptability. Similarly, the DNN paired with GPT-2 and the DNN paired with SentenceTransformer showed improvements, although their MAE levels stabilised at higher values, around 0.03 to 0.05, suggesting moderate effectiveness. In contrast, SVM paired with DistilBERT, XGBoost paired with DistilBERT, LightGBM paired with DistilBERT, and SVM paired with SentenceTransformer consistently displayed high MAE values near 0.12. DNN and LightGBM maintained stable MAE values around 0.08, indicating average performance. DNN integrated with DistilBERT demonstrated exceptional performance across all metrics (Supplementary Figs. 11, 12, and 13).

| Droplet Capillary Number | | | | |
|---|---|---|---|---|
| Model | Metrics | | | |
| | MAE | MSE | RMSE | R² |
| DNN | 0.0828 ± 0.0012 | 0.0185 ± 0.0004 | 0.1327 ± 0.0014 | 0.8301 ± 0.0042 |
| DNN–DistilBERT | 0.0115 ± 0.0009 | 0.0005 ± 0.0002 | 0.0163 ± 0.0016 | 0.9947 ± 0.0023 |
| DNN–GPT-2 | 0.0663 ± 0.0057 | 0.0017 ± 0.0003 | 0.0354 ± 0.0017 | 0.9842 ± 0.0025 |
| DNN–SentenceTransformer | 0.092 ± 0.0086 | 0.0041 ± 0.0003 | 0.0603 ± 0.0018 | 0.9611 ± 0.0034 |
| LightGBM | 0.0819 ± 0.0010 | 0.0249 ± 0.0005 | 0.1565 ± 0.0016 | 0.7714 ± 0.0042 |
| LightGBM–DistilBERT | 0.1161 ± 0.0014 | 0.0341 ± 0.0006 | 0.1839 ± 0.0016 | 0.686 ± 0.0049 |
| LightGBM–GPT-2 | 0.0945 ± 0.0016 | 0.0265 ± 0.0005 | 0.162 ± 0.0016 | 0.7546 ± 0.0107 |
| LightGBM–SentenceTransformer | 0.1061 ± 0.0021 | 0.0307 ± 0.0006 | 0.1741 ± 0.0016 | 0.7182 ± 0.0043 |
| SVM | 0.0918 ± 0.0020 | 0.0223 ± 0.0009 | 0.1487 ± 0.0030 | 0.7919 ± 0.0042 |
| SVM–DistilBERT | 0.1218 ± 0.0012 | 0.0316 ± 0.0006 | 0.1771 ± 0.0020 | 0.7061 ± 0.0055 |
| SVM–GPT-2 | 0.1137 ± 0.0015 | 0.0296 ± 0.0007 | 0.1717 ± 0.0018 | 0.7289 ± 0.0073 |
| SVM–SentenceTransformer | 0.1267 ± 0.0022 | 0.0325 ± 0.0009 | 0.1795 ± 0.0026 | 0.6995 ± 0.0094 |
| XGBoost | 0.0936 ± 0.0009 | 0.0234 ± 0.0004 | 0.1522 ± 0.0013 | 0.7929 ± 0.0122 |
| XGBoost–DistilBERT | 0.1194 ± 0.0025 | 0.0341 ± 0.0007 | 0.1832 ± 0.0022 | 0.6893 ± 0.0083 |
| XGBoost–GPT-2 | 0.0995 ± 0.0011 | 0.0269 ± 0.0015 | 0.1642 ± 0.0031 | 0.7479 ± 0.012 |
| XGBoost–SentenceTransformer | 0.1108 ± 0.0011 | 0.0308 ± 0.0011 | 0.1748 ± 0.0033 | 0.7151 ± 0.0051 |

Table 4. Metrics of evaluation for droplet capillary number, compared across models

To highlight the low standard error and the stabilisation of mean estimation performance across metrics of prediction for droplet capillary number for different models, we present an extensive comparative analysis of DNN, LightGBM, SVM, and XGBoost paired with the various NLP models, i.e., DistilBERT, GPT-2, and SentenceTransformer, as detailed in Table 4. The performances of the models are rigorously evaluated using MAE, MSE, RMSE, and R². The table presents values derived from the repetitions in which the difference between the mean and median of the respective metric was minimised. This analysis provides key insights into the efficacy of these models in handling complex predictive tasks. The DNN model combined with DistilBERT emerged as the most effective, achieving an MAE of 0.0115, MSE of 0.0005, RMSE of 0.0163, and an R² value of 0.9947. These results indicate that DNN–DistilBERT has a remarkable ability to minimise error and maximise predictive accuracy, suggesting a strong synergy between deep learning architectures and transformer-based embeddings. The high R² value of almost 1 implies that the model can explain nearly all the variability in the data, making it highly reliable for predictive tasks. In contrast, other model–embedding combinations showed significant weaknesses in performance. For instance, LightGBM, while known for its efficiency in handling structured data, showed limitations when paired with NLP embeddings. The combination of LightGBM and SentenceTransformer resulted in a relatively high MAE of 0.1061 and a low R² of 0.7182, reflecting a less accurate prediction capability. Similarly, SVM, a model traditionally strong in classification tasks, demonstrated suboptimal performance with DistilBERT (MAE: 0.1218, R²: 0.7061), indicating that SVM may not fully capitalise on the rich semantic features offered by transformers. A critical observation is the consistent underperformance of SentenceTransformer across all models. Despite its theoretical promise, SentenceTransformer consistently yielded higher error rates and lower R² values than DistilBERT and GPT-2. This suggests that SentenceTransformer may not capture the linguistic nuances required for high-accuracy predictions in these scenarios. The underperformance of the models with this embedding highlights the importance of choosing the right embedding based on the specific task and dataset characteristics. Moreover, the data reveals that advanced transformers like DistilBERT and GPT-2 significantly enhance model performance, especially in deep learning contexts. For instance, the DNN model with GPT-2 achieved a competitive MAE of 0.0663 and an R² of 0.9842, although slightly inferior to DNN–DistilBERT. This reinforces the idea that the architecture of the embedding model plays a crucial role in determining the overall performance, with simpler embeddings potentially lacking the depth needed for nuanced predictions. XGBoost, often favoured for its robustness and speed in structured data tasks, also showed variability depending on the embedding used. While XGBoost–DistilBERT achieved a reasonable performance (MAE: 0.1194, R²: 0.6893), it still lagged behind DNN-based approaches, particularly in terms of error minimisation. This suggests that while XGBoost remains a strong contender in traditional settings, its performance may be somewhat poorer when tasked with complex NLP tasks. The design automation for the remaining parameters is detailed in the supplementary file.

## 4. Conclusion

In this paper, we present µ-Fluidic-LLMs, an innovative framework designed to process tabular data associated with droplet microfluidics by transforming it into a text-based representation that effectively incorporates essential contextual information, including column descriptions. Our findings underscore the critical role of model selection and language model integration in achieving optimal efficiency of performance prediction and design automation. We emphasise the superior performance of DNN models, particularly when integrated with advanced NLP models such as DistilBERT and GPT-2. In comparison, traditional ensemble methods like LightGBM and XGBoost, even when incorporating state-of-the-art NLP approaches, may struggle to achieve comparable performance levels.

Multiple research directions can be pursued to further advance and optimise our framework. For example, exploration of various text serialisation methods, regularisation techniques, dataset augmentation, or advanced optimisation algorithms could provide means to enhance model performance and reduce error rates. Furthermore, given the flexibility of our framework to accommodate different language models, it can be utilised with the latest LLMs, such as GPT-4,[44,45] PaLM 2,[46] Gemini 1.5,[47] and LLaMA 3,[48] to assess variations in overall performance. Moreover, µ-Fluidic-LLMs is a general framework that can be adapted for use with other microfluidic components, such as micromixers,[49] and extended to non-microfluidic structures like 3D-printed lattices[50], thereby enabling the automation of intricate design processes. In addition, µ-Fluidic-LLMs can be seamlessly integrated with existing microfluidic computer-aided design tools[51–59] to enable more sophisticated and advanced design automation. We anticipate that this work will inspire further research into the application of language technologies across a broader range of microfluidic applications.

**Conflicts of interest**

There are no conflicts to declare.

**Data availability**

https://github.com/duydinhlab/MicrofluidicLLMs

**Acknowledgments**

We gratefully acknowledge the funding provided by the Research Grant Council of Hong Kong, General Research Fund (Ref No. 14211223).

## 5. References

1	G. M. Whitesides, *Nat. 2006 4427101*, 2006, **442**, 368–373.

2	Y. Ding, P. D. Howes and A. J. Demello, *Anal. Chem.*, 2020, **92**, 132–149.

3	E. Y. u. Basova and F. Foret, *Analyst*, 2014, **140**, 22–38.

4	S. H. Han, Y. Choi, J. Kim and D. Lee, *ACS Appl. Mater. Interfaces*, 2020, **12**, 3936–3944.

5	T. Moragues, D. Arguijo, T. Beneyton, C. Modavi, K. Simutis, A. R. Abate, J. C. Baret, A. J. deMello, D. Densmore and A. D. Griffiths, *Nat. Rev. Methods Prim. 2023 31*, 2023, **3**, 1–22.

6	R. Zilionis, J. Nainys, A. Veres, V. Savova, D. Zemmour, A. M. Klein and L. Mazutis, *Nat. Protoc. 2016 121*, 2016, **12**, 44–73.

7	M. Pellegrino, A. Sciambi, S. Treusch, R. Durruthy-Durruthy, K. Gokhale, J. Jacob, T. X. Chen, J. A. Geis, W. Oldham, J. Matthews, H. Kantarjian, P. A. Futreal, K. Patel, K. W. Jones, K. Takahashi and D. J. Eastburn, *Genome Res.*, 2018, **28**, 1345–1352.

8	X. Zhang, T. Li, F. Liu, Y. Chen, J. Yao, Z. Li, Y. Huang and J. Wang, *Mol. Cell*, 2019, **73**, 130-142.e5.

9	K. Matuła, F. Rivello and W. T. S. Huck, *Adv. Biosyst.*, 2020, **4**, 1900188.

10	H. M. Kang, M. Subramaniam, S. Targ, M. Nguyen, L. Maliskova, E. McCarthy, E. Wan, S. Wong, L. Byrnes, C. M. Lanata, R. E. Gate, S. Mostafavi, A. Marson, N. Zaitlen, L. A. Criswell and C. J. Ye, *Nat. Biotechnol. 2017 361*, 2017, **36**, 89–94.

11	H. Yin, Z. Wu, N. Shi, Y. Qi, X. Jian, L. Zhou, Y. Tong, Z. Cheng, J. Zhao and H.

Mao, *Biosens. Bioelectron.*, 2021, **188**, 113282.

12  Y. Hou, S. Chen, Y. Zheng, X. Zheng and J. M. Lin, *TrAC Trends Anal. Chem.*, 2023, **158**, 116897.

13  C. Wei, C. Yu, S. Li, J. Meng, T. Li, J. Cheng and J. Li, *Sensors Actuators B Chem.*, 2022, **371**, 132473.

14  F. M. Galogahi, M. Christie, A. S. Yadav, H. An, H. Stratton and N. T. Nguyen, *Analyst*, 2023, **148**, 4064–4071.

15  Y. Belotti and C. T. Lim, *Anal. Chem.*, 2021, **93**, 4727–4738.

16  T. S. Kaminski, O. Scheler and P. Garstecki, *Lab Chip*, 2016, **16**, 2168–2187.

17  D. T. Chiu, A. J. deMello, D. Di Carlo, P. S. Doyle, C. Hansen, R. M. Maceiczyk and R. C. R. Wootton, *Chem*, 2017, **2**, 201–223.

18  F. Su, K. Chakrabarty and R. B. Fair, *IEEE Trans. Comput. Des. Integr. Circuits Syst.*, 2006, **25**, 211–223.

19  S. Battat, D. A. Weitz and G. M. Whitesides, *Lab Chip*, 2022, **22**, 530–536.

20  D. McIntyre, A. Lashkaripour, P. Fordyce and D. Densmore, *Lab Chip*, 2022, **22**, 2925–2937.

21  Y. Mahdi and K. Daoud, *J. Dispers. Sci. Technol.*, 2017, **38**, 1501–1508.

22  A. Lashkaripour, C. Rodriguez, N. Mehdipour, R. Mardian, D. McIntyre, L. Ortiz, J. Campbell and D. Densmore, *Nat. Commun. 2021 121*, 2021, **12**, 1–14.

23  S. A. Damiati, D. Rossi, H. N. Joensson and S. Damiati, *Sci. Reports 2020 101*, 2020, **10**, 1–11.

24  S. H. Hong, H. Yang and Y. Wang, *Microfluid. Nanofluidics*, 2020, **24**, 1–20.

25  W. Ji, T. Y. Ho, J. Wang and H. Yao, *IEEE Trans. Comput. Des. Integr. Circuits Syst.*, 2020, **39**, 2544–2557.

26  A. Vaswani, G. Brain, N. Shazeer, N. Parmar, J. Uszkoreit, L. Jones, A. N. Gomez, Ł. Kaiser and I. Polosukhin, arXiv:1706.03762, 2017

27  E. Alsentzer, J. R. Murphy, W. Boag, W.-H. Weng, D. Jin, T. Naumann and M. B. A. Mcdermott, 2019, 72–78.

28  J. Lee, W. Yoon, S. Kim, D. Kim, S. Kim, C. H. So and J. Kang, *Bioinformatics*, 2020,

**36**, 1234–1240.

29  R. Luo, L. Sun, Y. Xia, T. Qin, S. Zhang, H. Poon and T. Y. Liu, *Brief. Bioinform.*, 2022, **23**, 1–11.

30  P. Yin, G. Neubig, W. T. Yih and S. Riedel, *Proc. Annu. Meet. Assoc. Comput. Linguist.*, 2020, 8413–8426.

31  T. Dinh, Y. Zeng, R. Zhang, Z. Lin, M. Gira, S. Rajput, J. Sohn, D. Papailiopoulos and K. Lee, *Adv. Neural Inf. Process. Syst.*, 2022, **35**, 11763–11784.

32  M. Sahakyan, Z. Aung and T. Rahwan, *IEEE Access*, 2021, **9**, 135392–135422.

33  T. Sun, Y. Shao, X. Qiu, Q. Guo, Y. Hu, X. Huang and Z. Zhang, *COLING 2020 - 28th Int. Conf. Comput. Linguist. Proc. Conf.*, 2020, 3660–3670.

34  J. Devlin, M.-W. Chang, K. Lee, K. T. Google and A. I. Language, *Proc. 2019 Conf. North*, 2019, 4171–4186.

35  A. Radford, J. Wu, R. Child, D. Luan, D. Amodei and I. Sutskever, *OpenAI blog*, p. 9, 2019

36  V. Sanh, L. Debut, J. Chaumond and T. Wolf, arXiv:1910.01108, 2019

37  N. Reimers and I. Gurevych, arXiv:1908.10084, 2019

38  T. Chen and C. Guestrin, *Proc. ACM SIGKDD Int. Conf. Knowl. Discov. Data Min.*, 2016, 13-17-August-2016, 785–794.

39  G. Ke, Q. Meng, T. Finley, T. Wang, W. Chen, W. Ma, Q. Ye and T.-Y. Liu, *Proceedings of the 31st International Conference on Neural Information Processing Systems (2017), pp. 3149-3157*

40  C. Cortes, V. Vapnik and L. Saitta, *Mach. Learn. 1995 203*, 1995, **20**, 273–297.

41  Y. Lecun, Y. Bengio and G. Hinton, *Nat. 2015 5217553*, 2015, **521**, 436–444.

42  M. Kuhn and K. Johnson, *Applied Predictive Modeling*, Springer (2013), ISBN: 9781461468493.

43  T. T. . Soong, *Fundamentals of Probability and Statistics for Engineers*, New York, NY: John Wiley & Sons (2004).

44  OpenAI *et al.*, arXiv:2303.08774v6, 2024.

45  S. Bubeck *et al.*, arXiv:2303.12712v5, 2023.

46   R. Anil *et al*., arXiv:2305.10403v3, 2023.

47   G. Team *et al.*, arXiv: 2403.05530v4, 2024.

48   A. Dubey *et al.*, arXiv: 2407.21783v2, 2024.

49   M. R. Rasouli and M. Tabrizian, *Lab Chip*, 2019, **19**, 3316–3325.

50   J. Mueller, J. R. Raney, K. Shea and J. A. Lewis, *Adv Mater, 30 (12) (2018), p. 1705001*.

51   H. Huang and D. Densmore, *ACM J. Emerg. Technol. Comput. Syst. 11, 26 (2014)*.

52   B. Crites, K. Kong and P. Brisk, *ACM J. Emerg. Technol. Comput. Syst.*, 2019, **16**, 14.

53   Y. Moradi, M. Ibrahim, K. Chakrabarty and U. Schlichtmann, *Proc. 2018 Des. Autom. Test Eur. Conf. Exhib. DATE 2018*, 2018, **2018**-**January**, 1484–1487.

54   Y. Zhu, X. Huang, B. Li, T. Y. Ho, Q. Wang, H. Yao, R. Wille and U. Schlichtmann, *IEEE Trans. Comput. Des. Integr. Circuits Syst.*, 2020, **39**, 2489–2502.

55   J. Wang, N. Zhang, J. Chen, V. G. J. Rodgers, P. Brisk and W. H. Grover, *Lab Chip*, 2019, **19**, 3618–3627.

56   M. Ibrahim, K. Chakrabarty and U. Schlichtmann, *Proc. 2017 Des. Autom. Test Eur. DATE 2017*, 2017, 1673–1678.

57   J. Wang, P. Brisk and W. H. Grover, *Lab Chip*, 2016, **16**, 4212–4219.

58   J. Wang, V. G. J. Rodgers, P. Brisk and W. H. Grover, *Biomicrofluidics*, 2017, 11, 034121.

59   A. Lashkaripour, D. P. McIntyre, S. G. K. Calhoun, K. Krauth, D. M. Densmore and P. M. Fordyce, *Nat. Commun. 2024 151*, 2024, **15**, 1–16.

**Electronic Supplementary Information (ESI)**

1. Performance Prediction

1.1 Droplet diameter

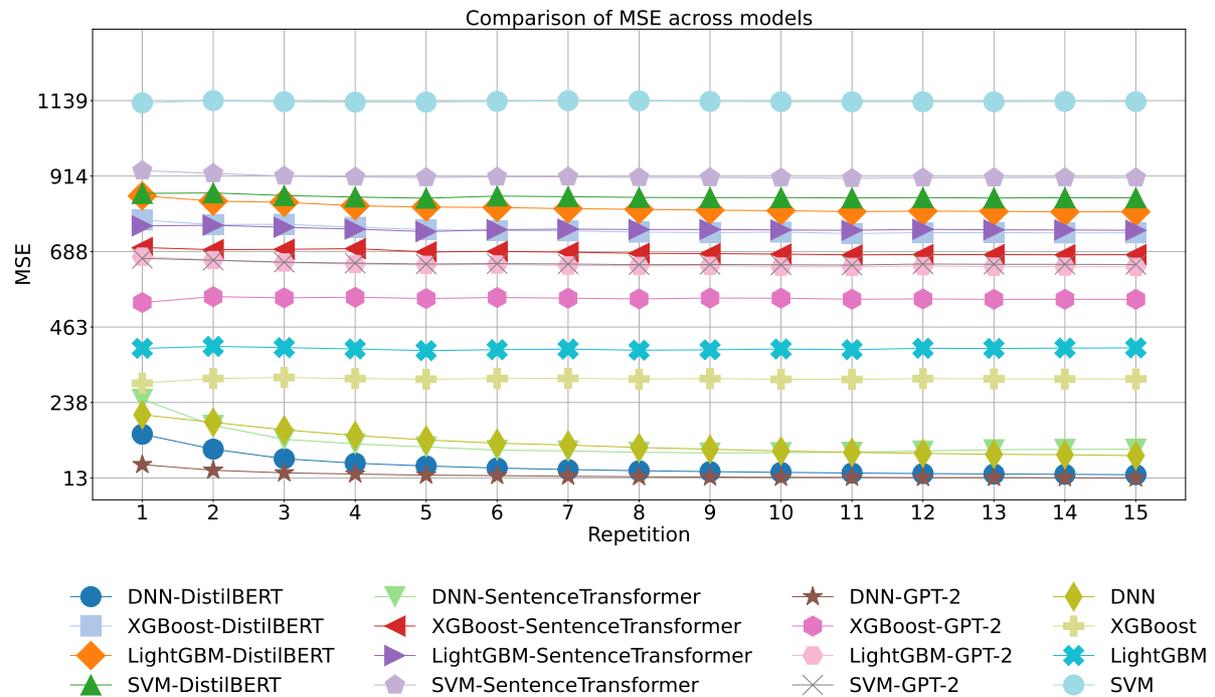

Supplementary Fig.1. Comparison of mean squared error (MSE) across models for droplet diameter

Supplementary Fig.1 presents a comprehensive comparison of 16 models evaluated over 15 iterations, using mean squared error (MSE) as the performance metric. The combination of deep neural network (DNN) with DistilBERT and GPT-2 demonstrates a notable improvement, reducing the mean square error to approximately 13. The DNN model, along with its integration with SentenceTransformer, initially shows an error near 238, which gradually stabilizes around 150, indicating moderate improvement. In contrast, the support vector machine (SVM) exhibits poor performance, with an error persisting around 1139. Other models exhibit minor fluctuations.

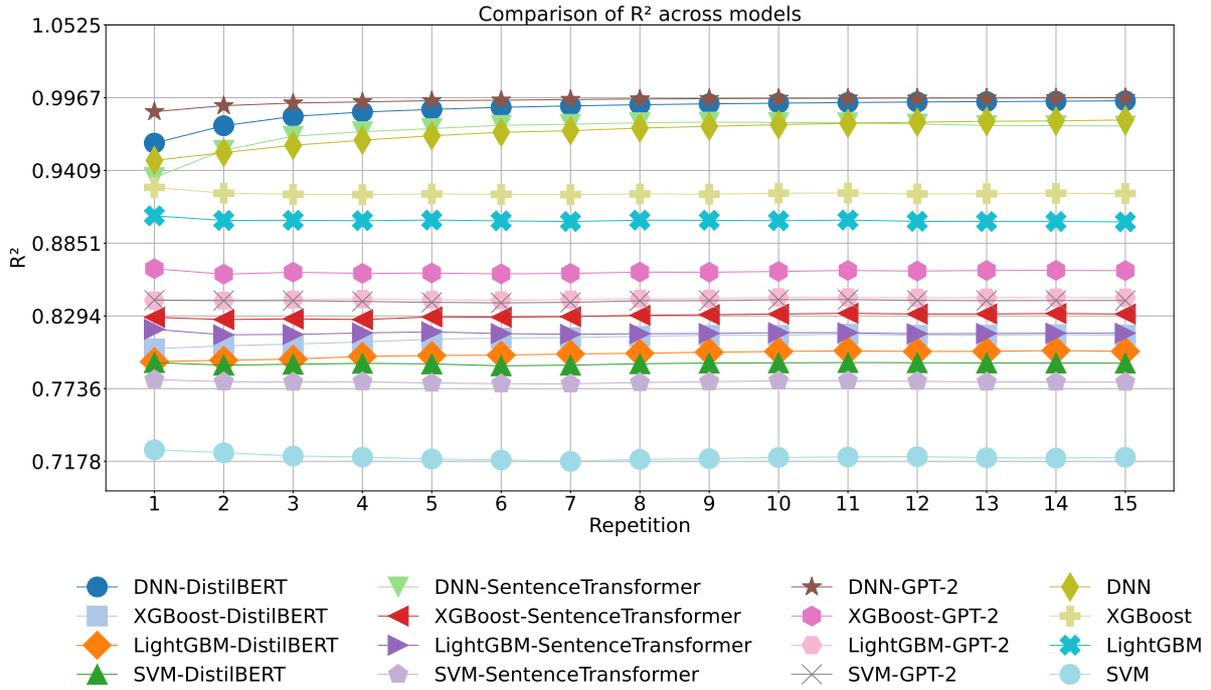

Supplementary Fig.2. Comparison of coefficient of determination (R²) across models for droplet diameter

Supplementary Fig.2 provides a detailed comparison of 16 models evaluated over 15 iterations, using the coefficient of determination ($R^2$) as the performance metric. The combination of the deep neural network (DNN) with DistilBERT and GPT-2 achieves high $R^2$ values, approaching 0.9967, indicating an excellent fit. The DNN model, as well as its integration with SentenceTransformer, achieves $R^2$ values around 0.98, indicating solid performance. In contrast, the support vector machine (SVM) exhibits a significantly lower $R^2$ of approximately 0.7178, reflecting poor predictive capability.

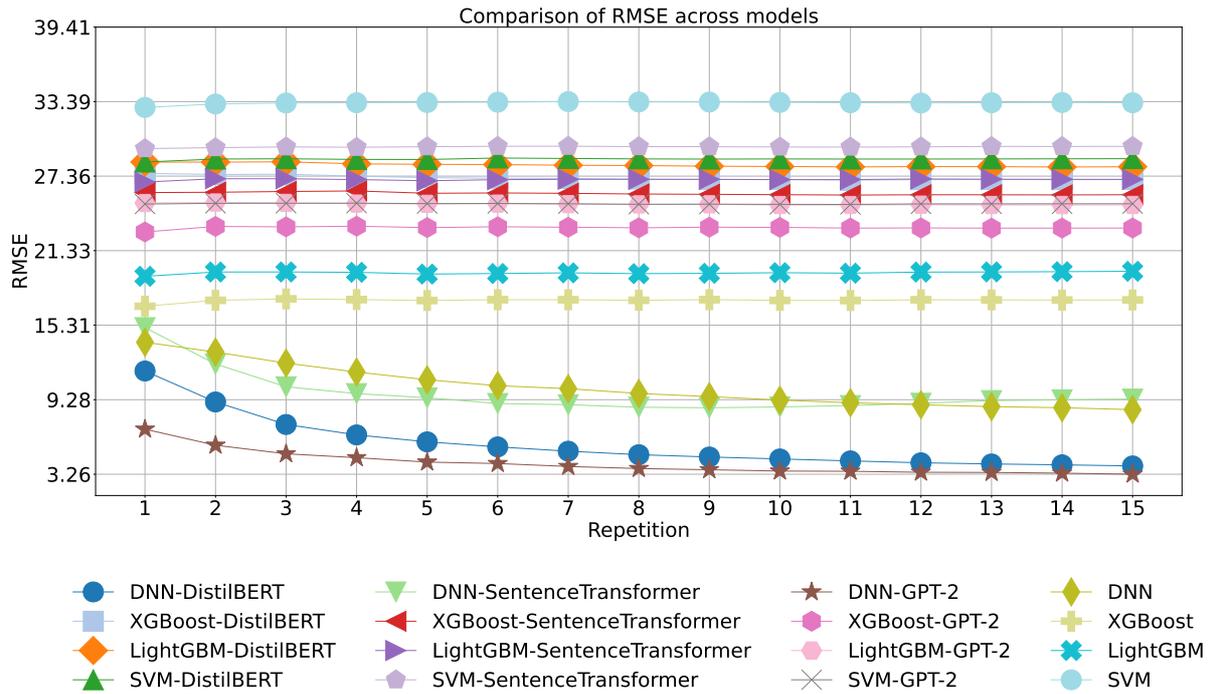

Supplementary Fig.3. Comparison of root mean squared error (RMSE) across models for droplet diameter

Supplementary Fig.3 presents a detailed comparison of 16 models assessed over 15 iterations, using root mean squared error (RMSE) as the performance metric. The combination of deep neural networks (DNN) with DistilBERT and GPT-2 demonstrates substantial accuracy improvement, with RMSE decreasing from approximately 9.28 to 3.26. In contrast, the support vector machine (SVM) exhibits poor performance, maintaining a high RMSE of around 33.39. The DNN model and its integration with SentenceTransformer show RMSE values starting below 15.31 and stabilizing around 10, reflecting moderate improvement. Other models display minor fluctuations.

Overall, the combination of DNN with DistilBERT and GPT-2 consistently shows the best performance across all metrics. In contrast, support vector machine (SVM) consistently performs poorly.

1.2 Droplet Generation Rate

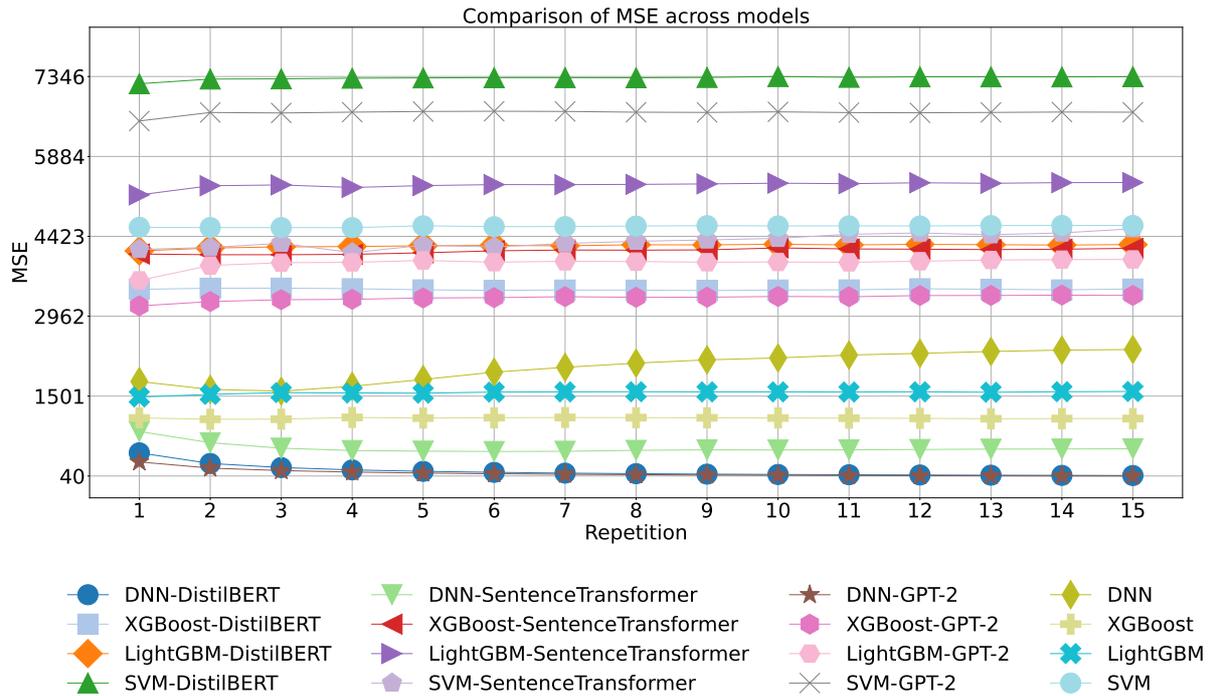

Supplementary Fig.4. Comparison of mean squared error (MSE) across models for droplet generation rate

Supplementary Fig.4 provides a detailed comparison of 16 models evaluated over 15 iterations, using mean squared error (MSE) as the performance metric. The combination of the deep neural network (DNN) with DistilBERT and GPT-2 achieves a significant reduction in MSE to approximately 40, demonstrating substantial improvement. The DNN integrated with DistilBERT shows a moderate improvement, with a MSE around 750. In contrast, the support vector machine (SVM) combined with DistilBERT maintains a high MSE of approximately 7346, indicating poor performance. Other models exhibit minor variations.

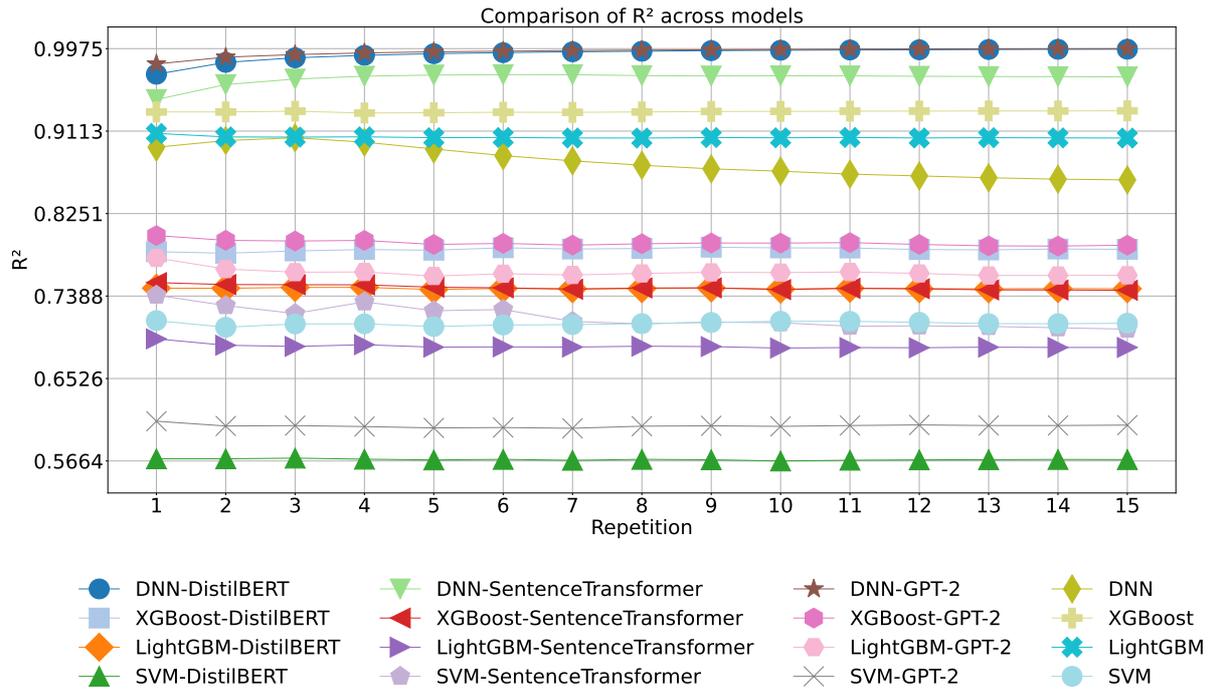

Supplementary Fig.5. Comparison of coefficient of determination (R²) across models for droplet generation rate

Supplementary Fig.5 offers a detailed comparison of 16 models evaluated over 15 iterations, using the coefficient of determination ($R^2$) as the performance metric. The deep neural network (DNN) combined with DistilBERT, GPT-2, and SentenceTransformer achieves high $R^2$ values nearing 0.9975, indicating an excellent fit. In contrast, the support vector machine (SVM) paired with DistilBERT exhibits a low $R^2$ of approximately 0.5664, reflecting poor predictive capability. Both XGBoost and LightGBM demonstrate decent performance, with $R^2$ values around 0.9113. Other models display moderate $R^2$ values.

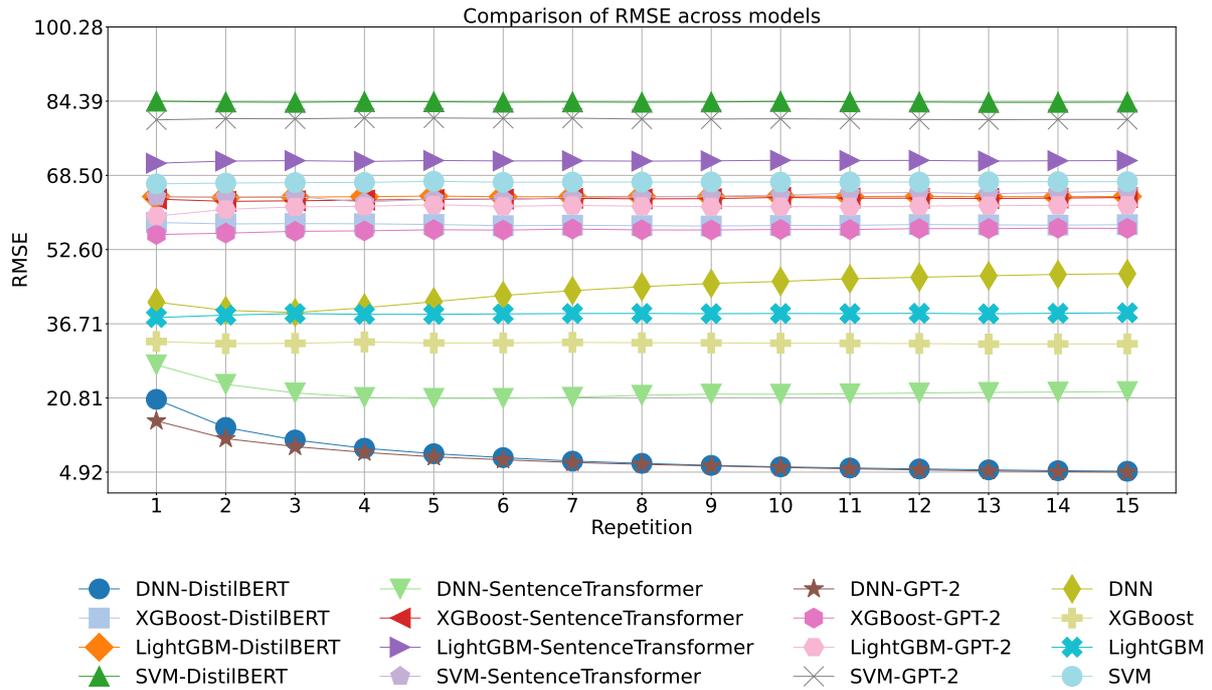

Supplementary Fig.6. Comparison of root mean squared error (RMSE) across models for droplet generation rate.

Supplementary Fig.6 presents a detailed comparison of 16 models assessed over 15 iterations, using root mean squared error (RMSE) as the performance metric. The deep neural network (DNN) integrated with DistilBERT and GPT-2 demonstrates substantial accuracy improvement, with RMSE decreasing from around 20 to 4.92. In contrast, the combination of support vector machine (SVM) with DistilBERT exhibits poor accuracy, maintaining a high RMSE of around 84. The DNN paired with SentenceTransformer remains around 20, indicating moderate improvement. Other models show slight variations.

Overall, the deep neural network (DNN) integrated with DistilBERT and GPT-2 consistently demonstrates superior performance across all metrics, exhibiting significant improvements. SVM paired with DistilBERT consistently performs poorly.

1.3 Droplet Regime

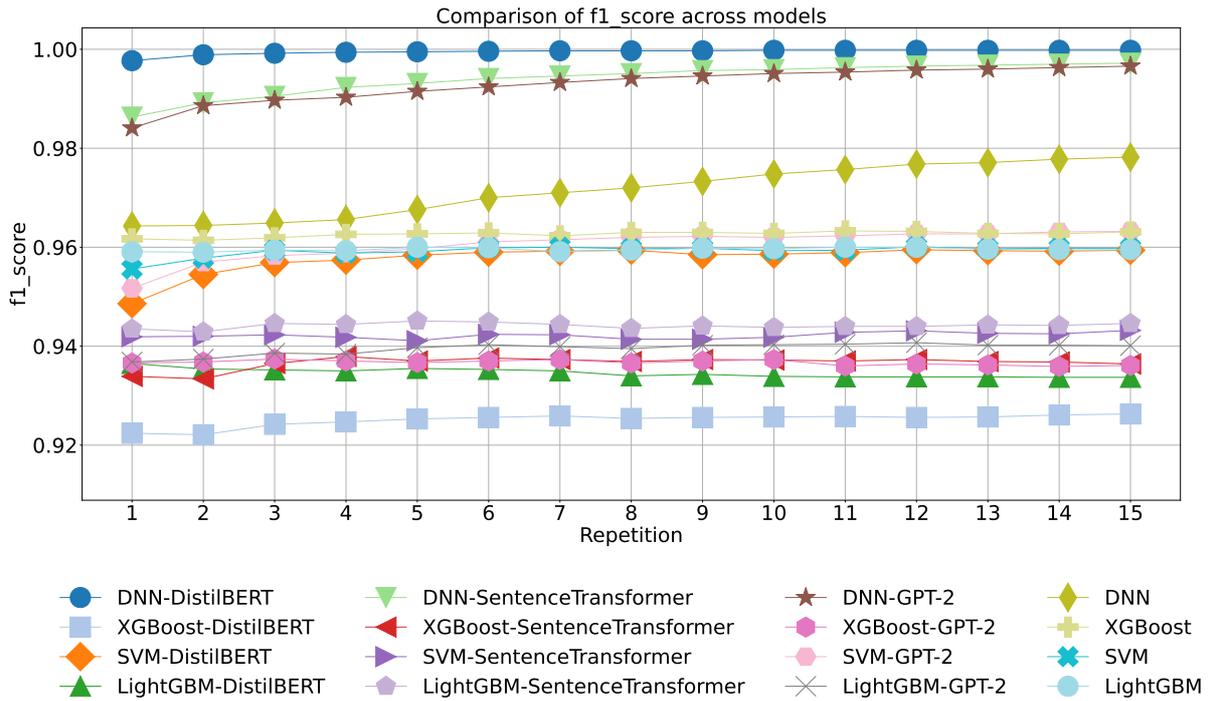

Supplementary Fig.7. Comparison of F1 score across models for droplet regime

Supplementary Fig.7 provides a detailed comparison of 16 models evaluated over 15 iterations, using F1 score as the performance metric. The deep neural network (DNN) combined with DistilBERT consistently achieves the highest F1 score, approaching 1.0, indicating excellent performance. DNN paired with SentenceTransformer and GPT-2 also demonstrate strong performance, with F1 scores around 0.98. In contrast, the combination of XGBoost with DistilBERT shows the lowest performance, with an F1 score of approximately 0.92.

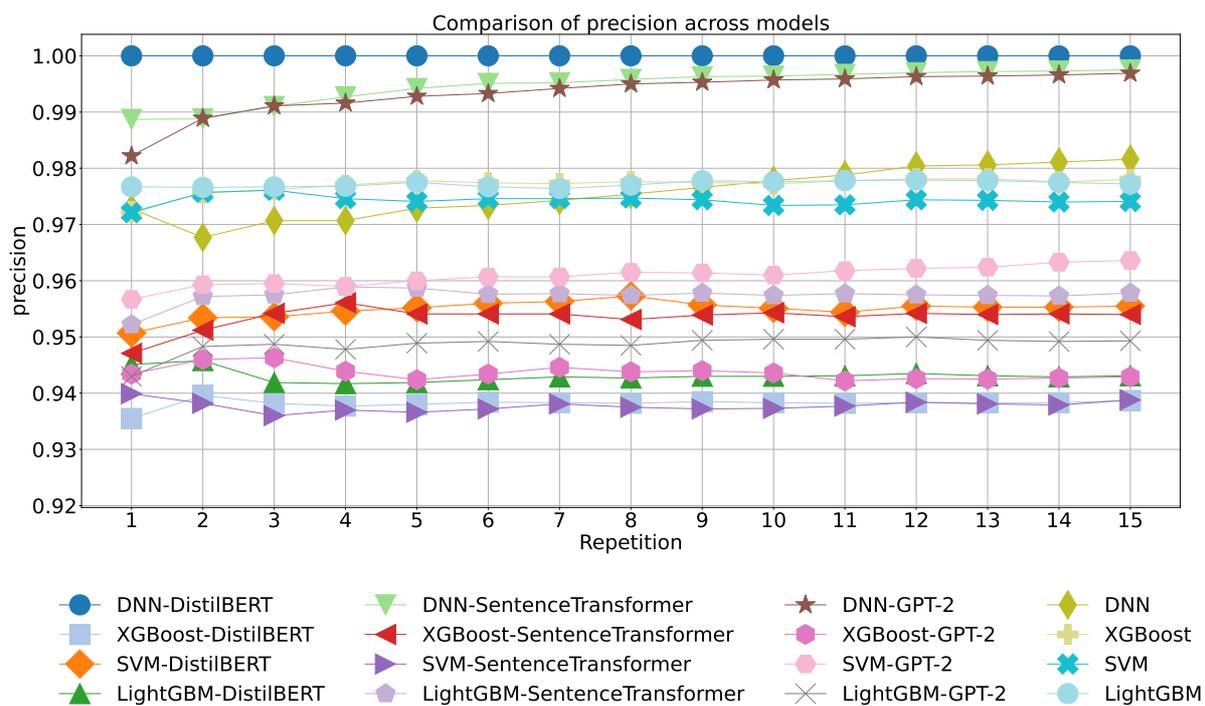

Supplementary Fig.8. Comparison of precision across models for droplet regime

Supplementary Fig.8 provides a detailed comparison of 16 models evaluated over 15 iterations, using precision as the performance metric. The deep neural network paired with DistilBERT maintains perfect precision across repetitions, indicating that its predictions are highly accurate. DNN paired with GPT-2 and DNN show strong precision as well. In contrast, XGBoost integrated with DistilBERT and SVM integrated with SentenceTransformer show the lowest precision, around 0.94.

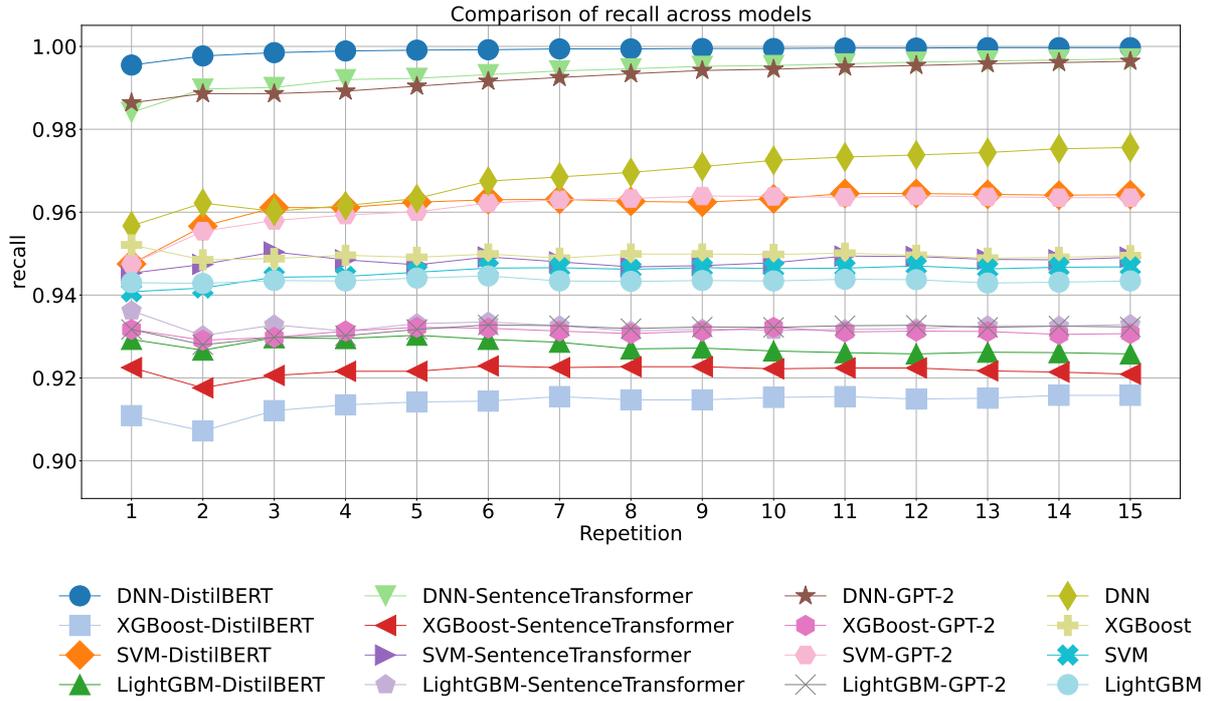

Supplementary Fig.9. Comparison of recall across models for droplet regime

Supplementary Fig.9 presents a detailed comparison of 16 models evaluated over 15 iterations, using recall as the performance metric. The integration of the deep neural network (DNN) with DistilBERT leads with a recall value near 1. DNN models paired with SentenceTransformer and GPT-2 follow closely, achieving recall values around 0.98. The combination of XGBoost with DistilBERT exhibits the lowest recall, approximately 0.92.

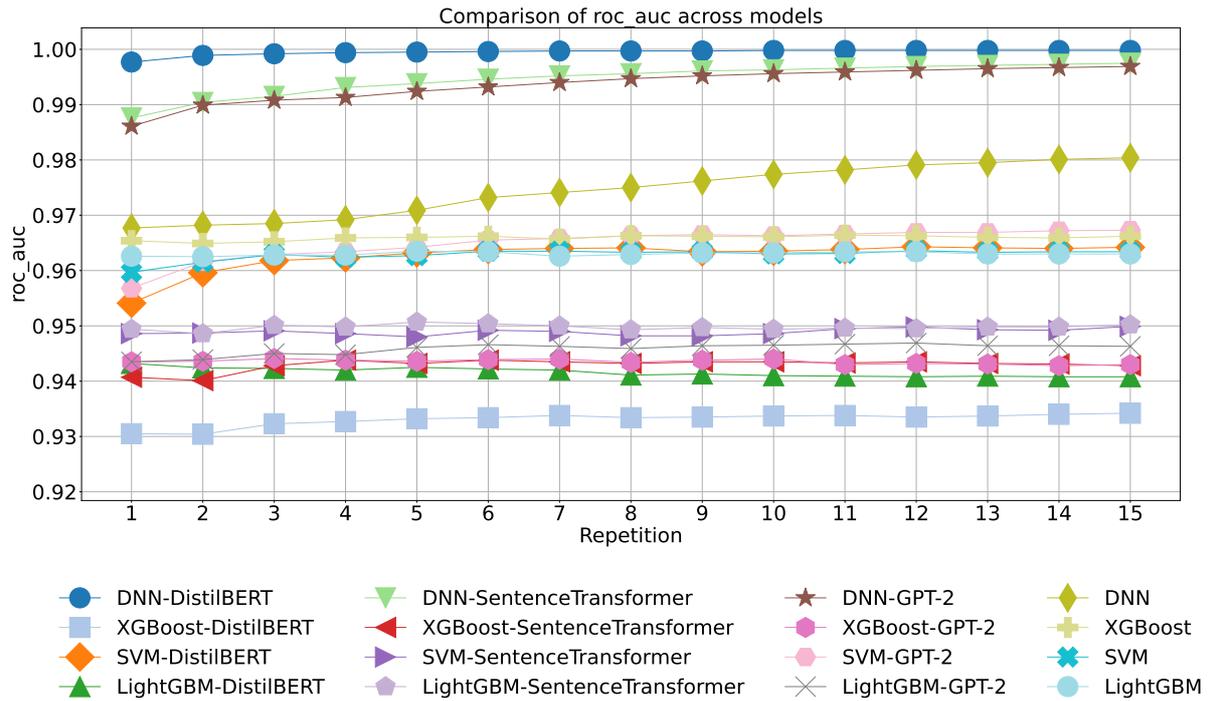

Supplementary Fig.10. Comparison of ROC AUC across models for droplet regime

Supplementary Fig.10 presents a detailed comparison of 16 models evaluated over 15 iterations, using ROC AUC as the performance metric. The deep neural network (DNN) paired with DistilBERT achieves the highest ROC AUC, indicating excellent ability to distinguish between classes. DNN combined with GPT-2 and SentenceTransformer perform similarly well, with scores around 0.98. XGBoost integrated with DistilBERT exhibits the lowest ROC AUC, around 0.93.

Overall, deep neural network (DNN) paired with DistilBERT consistently outperforms other models across all metrics, demonstrating superior predictive capabilities. XGBoost integrated with DistilBERT, on the other hand, shows the least effective performance, particularly in precision and recall.

2. Design Automation

2.1 Droplet capillary number

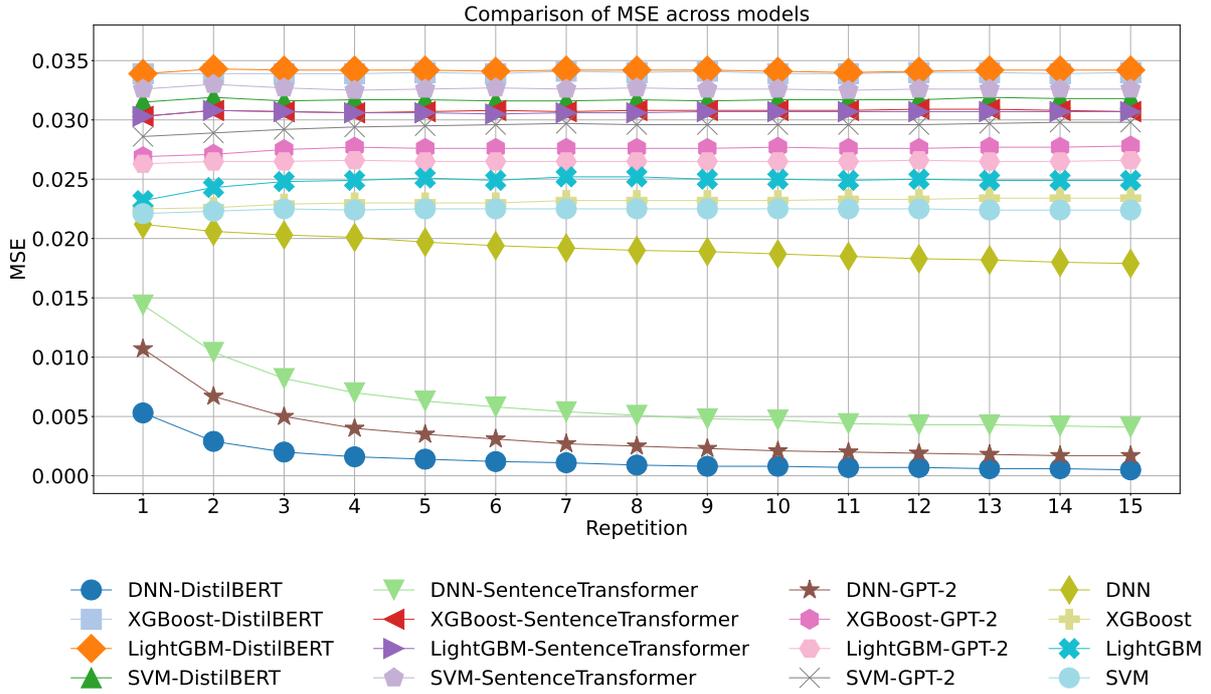

Supplementary Fig.11. Comparison of mean squared error (MSE) across models for droplet capillary number

Supplementary Fig.11 provides a detailed comparison of 16 models evaluated over 15 iterations, using mean squared error (MSE) as the performance metric. Combination of deep neural network (DNN) with DistilBERT shows the lowest MSE, decreasing consistently as repetitions increase, indicating high accuracy and improving performance. DNN paired with GPT-2 and SentenceTransformer also demonstrates low MSE values. DistilBERT paired with XGBoost and LightGBM yields the highest MSE values, indicating less accurate predictions.

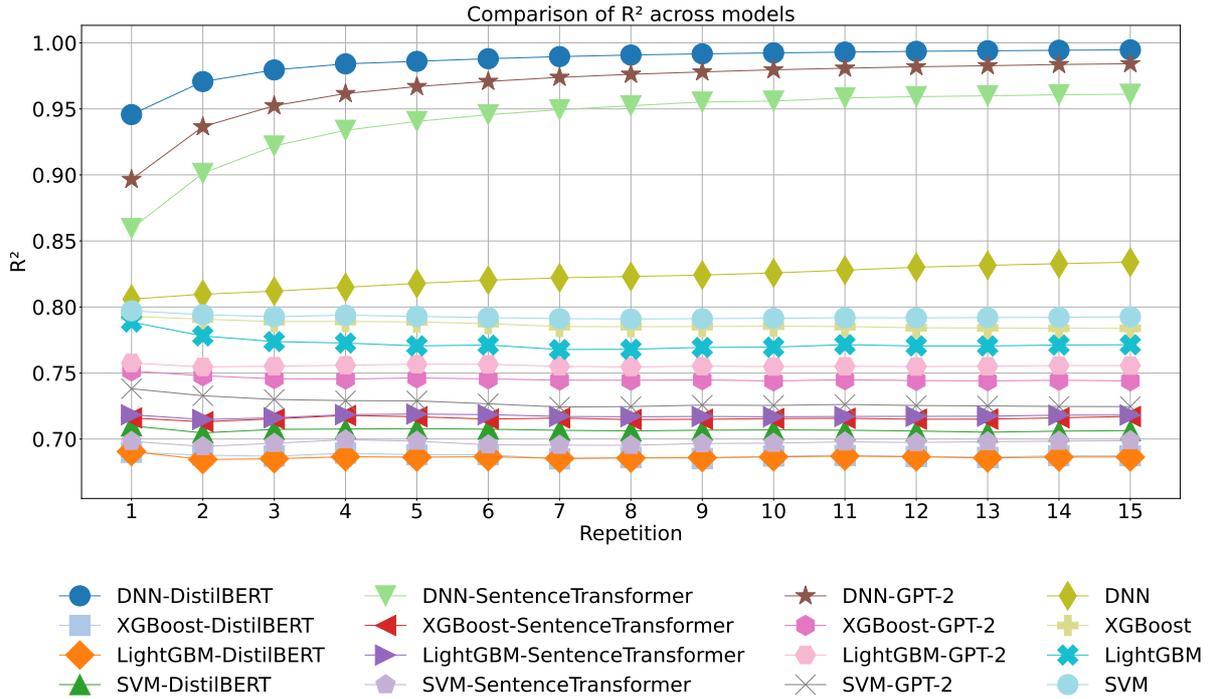

Supplementary Fig.12. Comparison of coefficient of determination (R²) across models for droplet capillary number

Supplementary Fig.12 provides a detailed comparison of 16 models evaluated over 15 iterations, using coefficient of determination ($R^2$) as the performance metric. Combination of deep neural network (DNN) with DistilBERT consistently achieves the highest $R^2$ values, close to 1, indicating excellent generalization. DNN paired with GPT-2 and SentenceTransformer also perform well with high $R^2$ values. DistilBERT paired with XGBoost and LightGBM yields lower $R^2$ values, suggesting poorer model generalization.

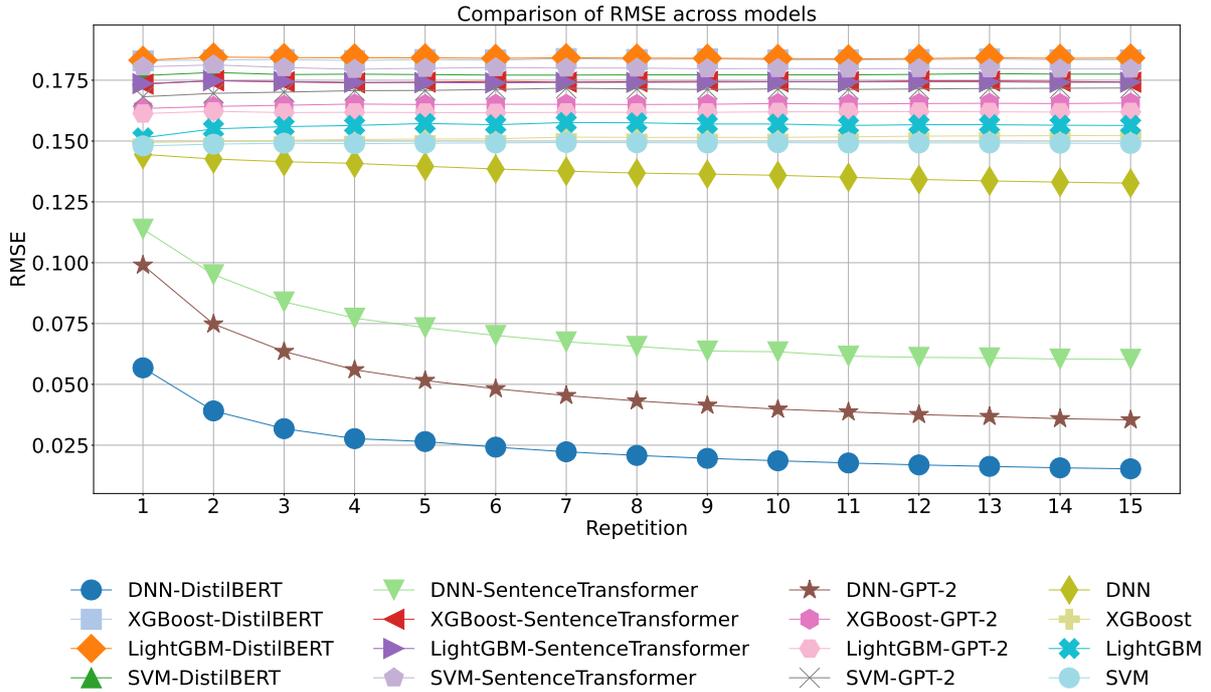

Supplementary Fig.13. Comparison of root mean squared error (RMSE) across models for droplet capillary number

Supplementary Fig.13 provides a detailed comparison of 16 models evaluated over 15 iterations, using root mean squared error (RMSE) as the performance metric. Combination of deep neural network (DNN) with DistilBERT has the lowest RMSE, confirming its superior predictive accuracy. Similar trends are observed with combination of deep neural network (DNN) with GPT-2 and SentenceTransformer, showing competitive RMSE values. DistilBERT paired with XGBoost and LightGBM exhibits higher RMSE values, reflecting less precision.

Overall, DNN integrated with DistilBERT demonstrates exceptional performance across all metrics. In contrast, DistilBERT paired with XGBoost and LightGBM appears to be the least effective, with higher error rates.

2.2 Droplet aspect ratio

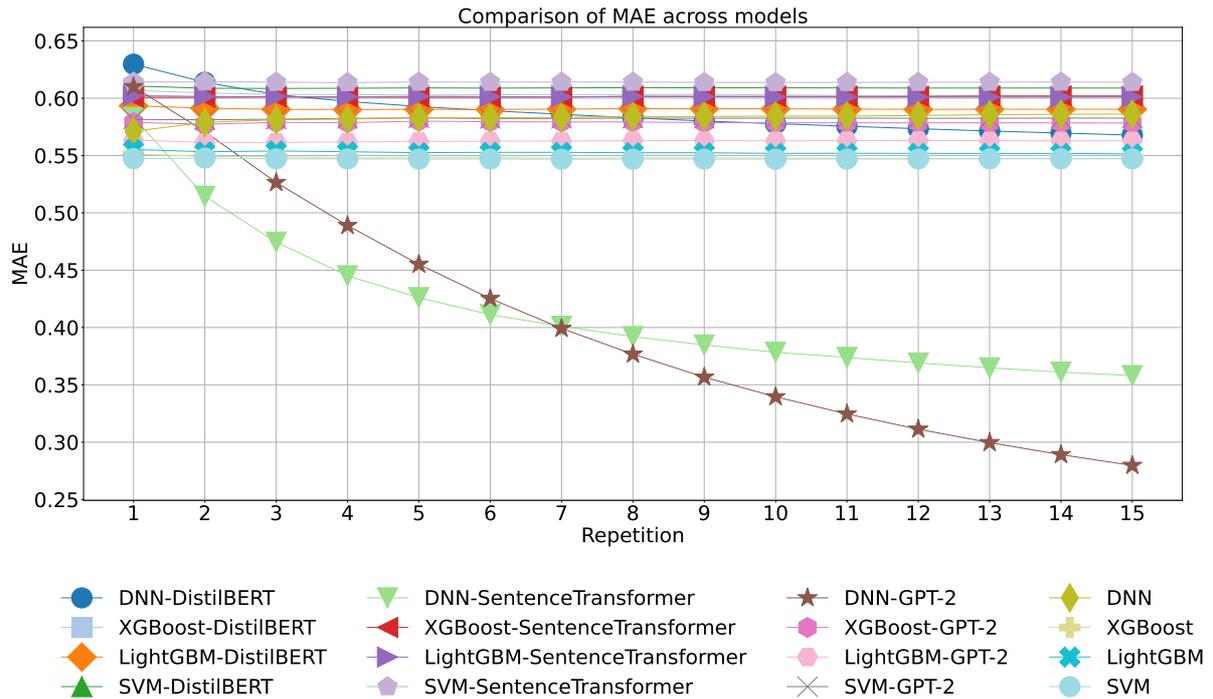

Supplementary Fig.14. Comparison of mean absolute error (MAE) across models for droplet aspect ratio

Supplementary Fig.14 provides a detailed comparison of 16 models evaluated over 15 iterations, using mean absolute error (MAE) as the performance metric. The deep neural network (DNN) integrated with SentenceTransformer and GPT-2 exhibits a clear decreasing trend in mean absolute error (MAE) over multiple repetitions, indicating improving accuracy. In contrast, the DNN paired with DistilBERT begins with a higher MAE but stabilizes quickly. Meanwhile, DistilBERT combined with XGBoost and LightGBM maintains consistent performance without notable improvement.

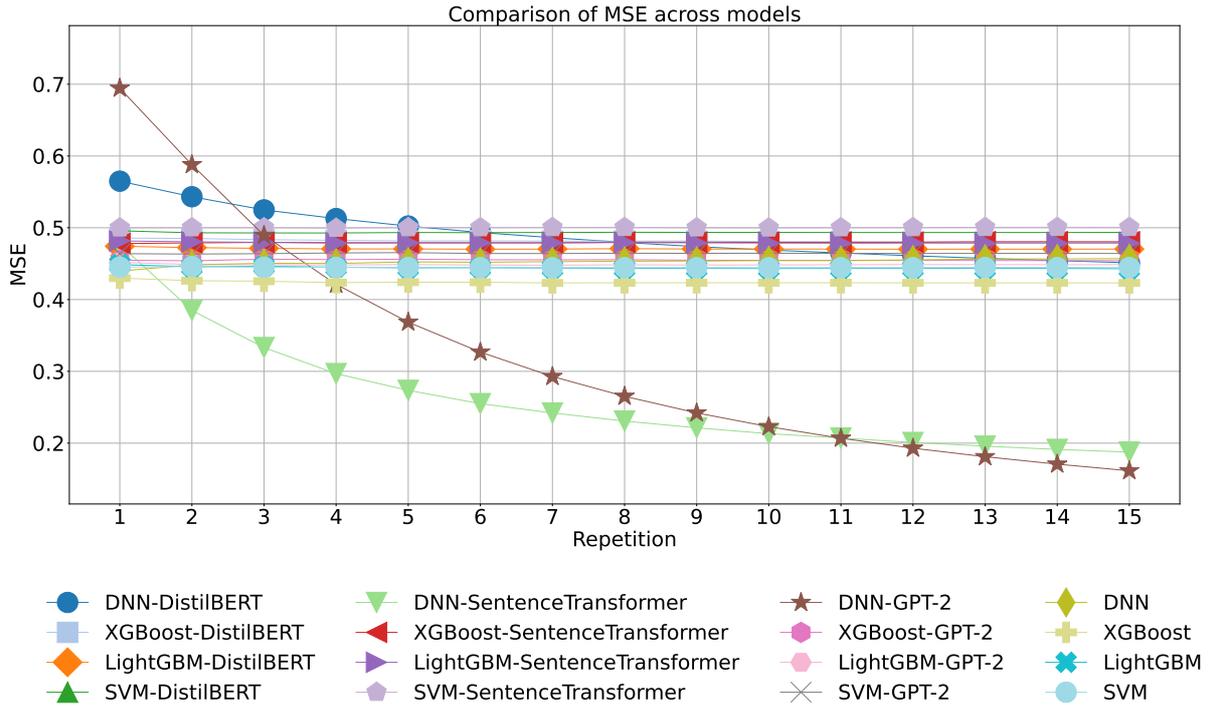

Supplementary Fig.15. Comparison of mean squared error (MSE) across models for droplet aspect ratio

Supplementary Fig.15 presents a detailed comparison of 16 models evaluated over 15 iterations, using mean squared error (MSE) as the performance metric. Deep neural network (DNN) paired with SentenceTransformer and GPT-2 demonstrate a significant reduction in MSE, reflecting enhanced predictive accuracy over time. Other models maintain consistent MSE values.

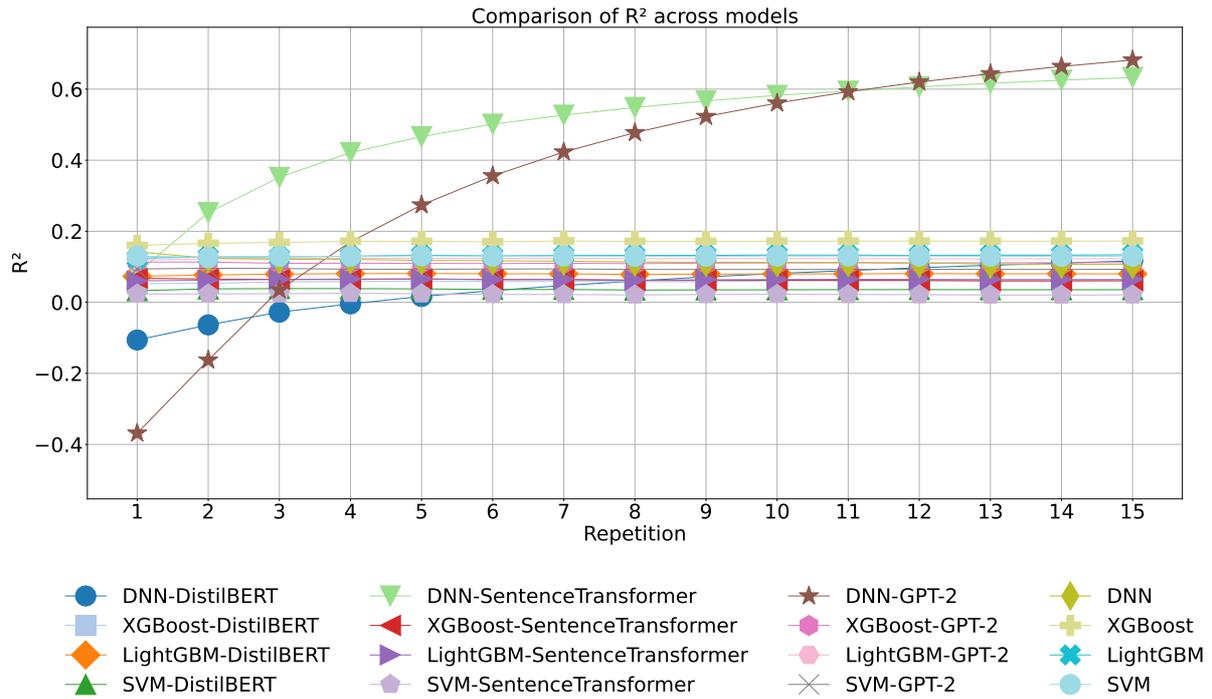

Supplementary Fig.16. Comparison of coefficient of determination (R²) across models for droplet aspect ratio

Supplementary Fig.16 presents a detailed comparison of 16 models evaluated over 15 iterations, using coefficient of determination (R²) as the performance metric. The deep neural network (DNN) combined with SentenceTransformer and GPT-2 exhibit a marked increase in R² values, indicating better model fit as repetitions increase. DNN paired with DistilBERT shows a gradual increase, suggesting improving performance. The remaining models have relatively stable R² values, indicating a consistent but less optimal fit.

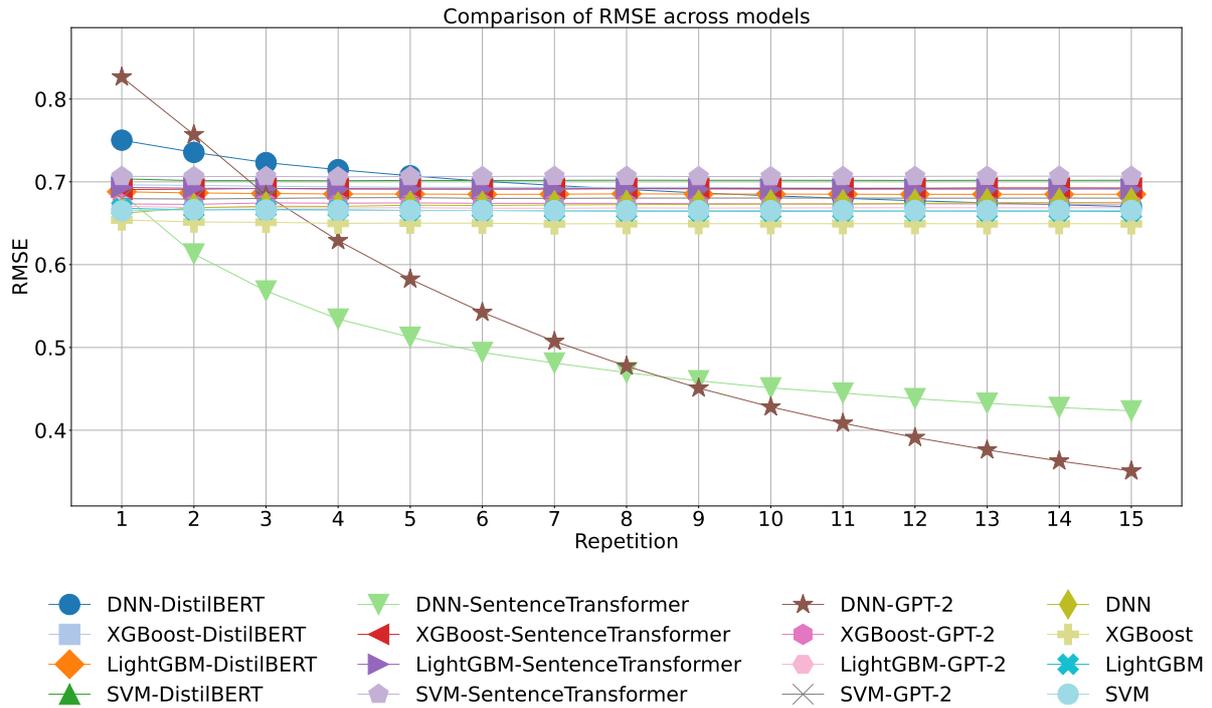

Supplementary Fig.17. Comparison of root mean squared error (RMSE) across models for droplet aspect ratio

Supplementary Fig.17 presents a detailed comparison of 16 models evaluated over 15 iterations, using root mean squared error (RMSE) as the performance metric. The deep neural network (DNN) combined with SentenceTransformer and GPT-2 continue to show decreasing RMSE, aligning with their improving MSE and MAE. DNN combined with DistilBERT also improves, albeit at a slower pace. Other models remain stable, indicating less change in predictive error.

Overall, DNN combined with SentenceTransformer and GPT-2 show the most significant improvements across all error metrics. DNN combined with DistilBERT improves steadily, while other models remain more constant in their performance.

| Droplet Aspect Ratio | | | | |
|---|---|---|---|---|
| Model | Metrics | | | |
| | MAE | MSE | RMSE | R² |
| DNN | 0.5857 ± 0.0031 | 0.452 ± 0.0058 | 0.6739 ± 0.003 | 0.1078 ± 0.0052 |
| DNN-DistilBERT | 0.6297 ± 0.0134 | 0.5432 ± 0.0152 | 0.7356 ± 0.0101 | 0.0165 ± 0.0196 |
| DNN-OpenGPT-2 | 0.5708 ± 0.0149 | 0.4894 ± 0.0384 | 0.757 ± 0.0272 | 0.0349 ± 0.0805 |
| DNN-SentenceTransformer | 0.5834 ± 0.0245 | 0.4767 ± 0.0355 | 0.6858 ± 0.0254 | 0.2525 ± 0.0469 |
| LightGBM | 0.5519 ± 0.0038 | 0.4439 ± 0.0066 | 0.6647 ± 0.0046 | 0.1327 ± 0.0126 |
| LightGBM-DistilBERT | 0.5905 ± 0.004 | 0.4705 ± 0.0062 | 0.6853 ± 0.0046 | 0.0802 ± 0.0053 |
| LightGBM-OpenGPT-2 | 0.5629 ± 0.0031 | 0.4481 ± 0.0067 | 0.6687 ± 0.0034 | 0.1223 ± 0.0078 |
| LightGBM-SentenceTransformer | 0.6012 ± 0.0033 | 0.4792 ± 0.0062 | 0.6917 ± 0.0035 | 0.0641 ± 0.0049 |
| SVM | 0.547 ± 0.0036 | 0.4445 ± 0.0064 | 0.6653 ± 0.005 | 0.1296 ± 0.0111 |
| SVM-DistilBERT | 0.6091 ± 0.0035 | 0.4935 ± 0.0042 | 0.7019 ± 0.003 | 0.0353 ± 0.0044 |
| SVM-OpenGPT-2 | 0.5826 ± 0.0041 | 0.4649 ± 0.0086 | 0.681 ± 0.0058 | 0.0931 ± 0.0067 |
| SVM-SentenceTransformer | 0.6138 ± 0.0037 | 0.5005 ± 0.0055 | 0.7065 ± 0.0044 | 0.0234 ± 0.0189 |
| XGBoost | 0.548 ± 0.0038 | 0.4236 ± 0.0072 | 0.6499 ± 0.0056 | 0.1726 ± 0.0065 |
| XGBoost-DistilBERT | 0.6032 ± 0.0054 | 0.4825 ± 0.0063 | 0.694 ± 0.0046 | 0.0523 ± 0.0139 |
| XGBoost-OpenGPT-2 | 0.5785 ± 0.003 | 0.4563 ± 0.0065 | 0.6741 ± 0.0044 | 0.1134 ± 0.0226 |
| XGBoost-SentenceTransformer | 0.6016 ± 0.0036 | 0.4789 ± 0.006 | 0.692 ± 0.0036 | 0.0644 ± 0.0067 |

Supplementary table 1. Metrics evaluation across models for droplet aspect ratio

Supplementary table 1 presents a comparative analysis of various machine learning models, including deep neural networks (DNN), LightGBM, support vector machines (SVM), and XGBoost, along with their combinations with natural language processing (NLP) models such as DistilBERT, OpenGPT-2, and SentenceTransformer. The models are evaluated using standard regression metrics: mean absolute error (MAE), mean squared error (MSE), root mean squared error (RMSE), and R². XGBoost achieves R² of 0.1726 and the lowest MSE of 0.4236. DNN paired with SentenceTransformer marginally achieves the highest R² (0.2525). However, some models incorporating advanced transformers, such as DNN-DistilBERT, exhibit lower R² values. These results highlight the importance of selecting appropriate embeddings to enhance model performance.

2.3 Droplet Expansion Ratio

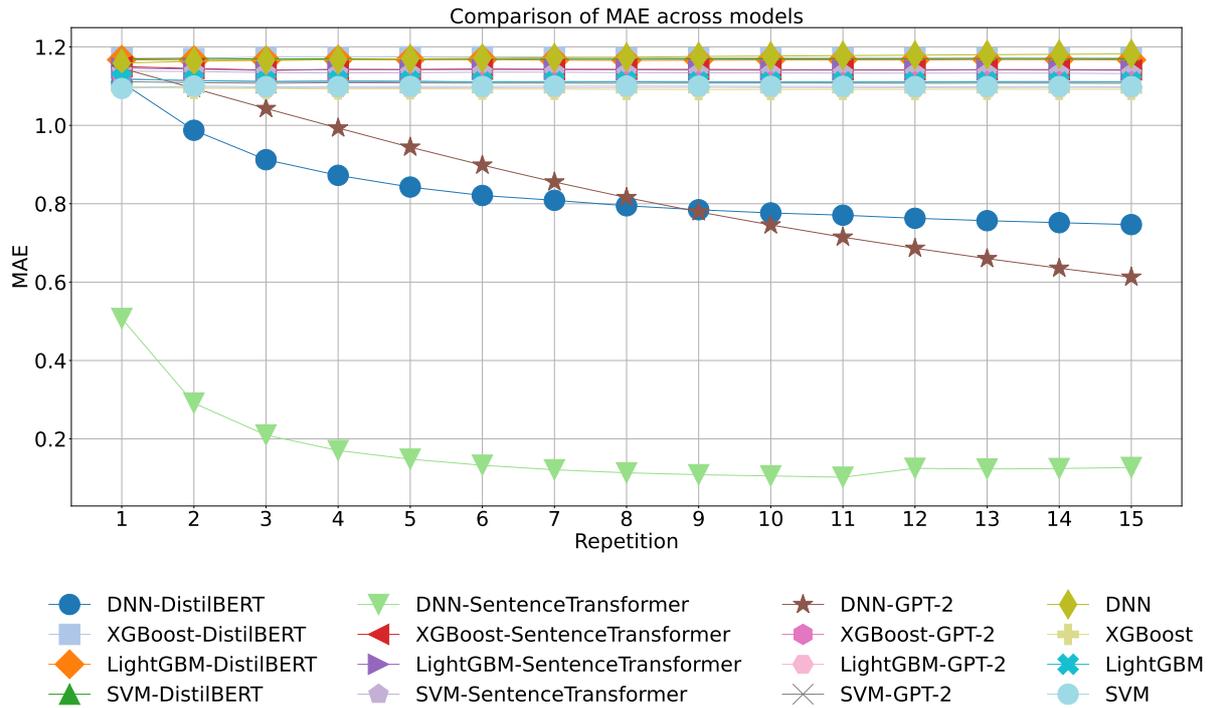

Supplementary Fig.18. Comparison of mean absolute error (MAE) across models for droplet expansion ratio

Supplementary Fig.18 presents a detailed comparison of 16 models evaluated over 15 iterations, using mean absolute error (MAE) as the performance metric. The deep neural network (DNN) paired with SentenceTransformer shows a significant decrease in MAE, indicating improved accuracy over repetitions. DNN paired with DistilBERT and GPT-2 also decreases steadily but at a slower rate. Others remain relatively stable, indicating consistent performance without much improvement.

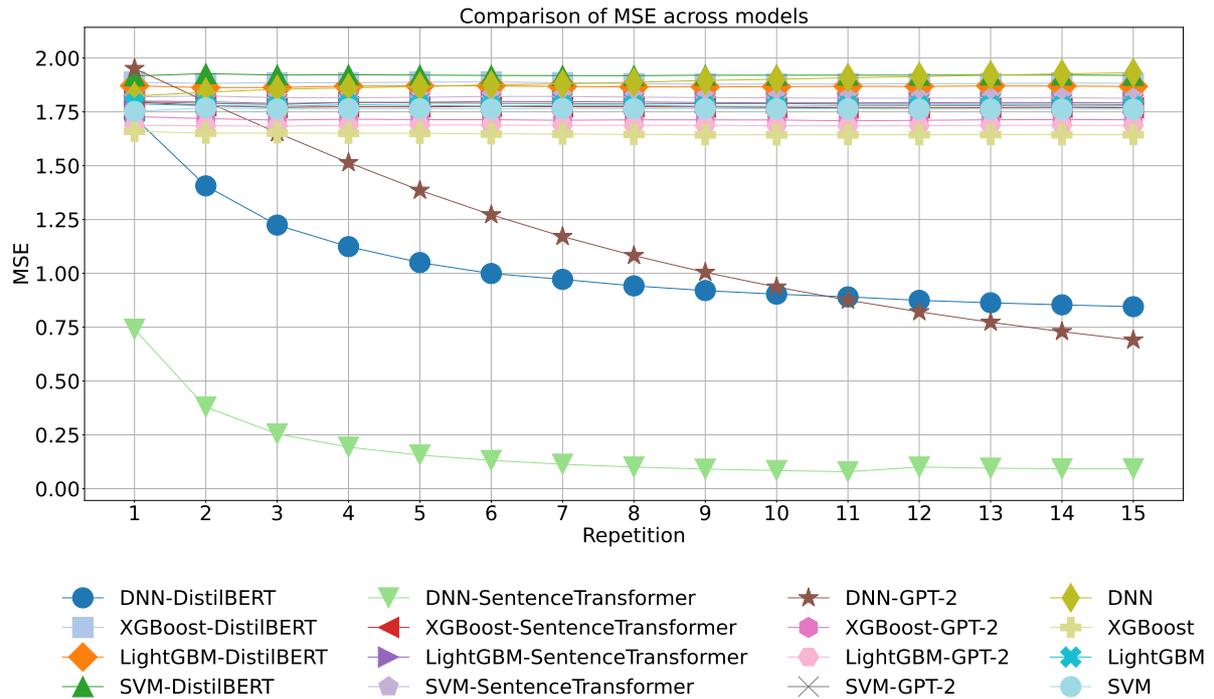

Supplementary Fig.19. Comparison of mean squared error (MSE) across models for droplet expansion ratio

Supplementary Fig.19 presents a detailed comparison of 16 models evaluated over 15 iterations, using mean squared error (MSE) as the performance metric. The deep neural network (DNN) paired with SentenceTransformer again demonstrates the largest reduction in MSE, suggesting enhanced predictive accuracy. DNN paired with DistilBERT and GPT-2 shows improvement. Other models have higher and more stable MSE values, indicating less change over repetitions.

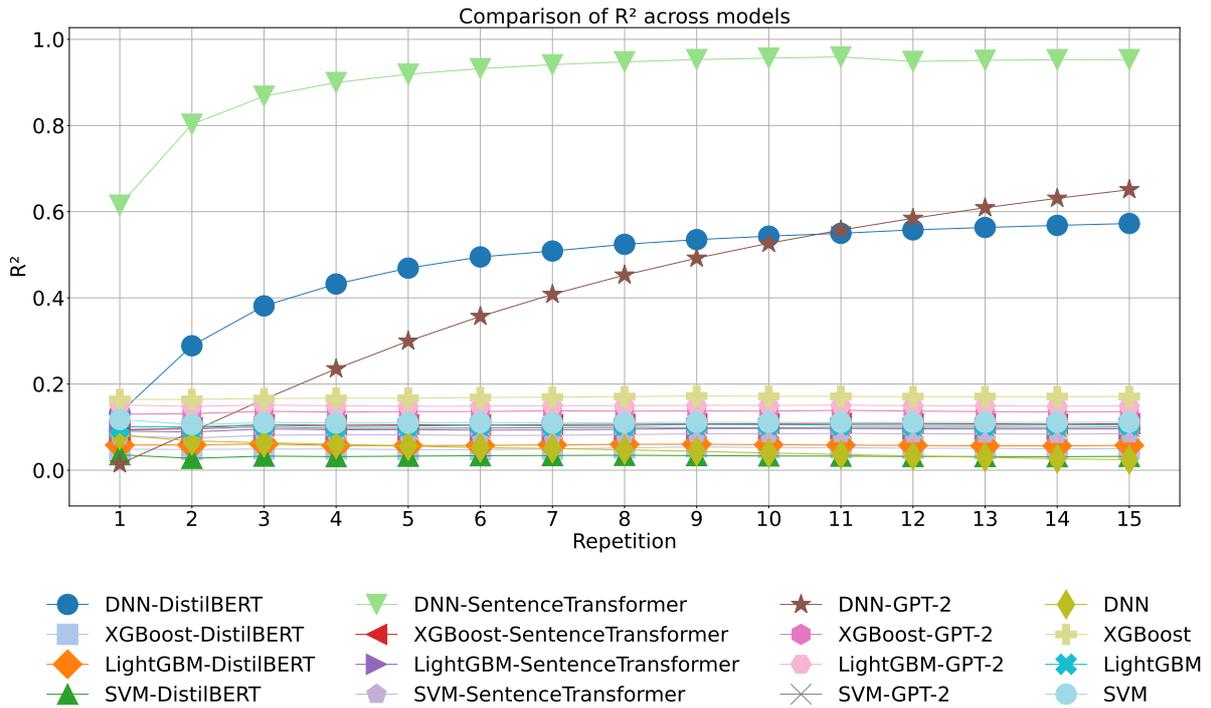

Supplementary Fig.20. Comparison of coefficient of determination (R²) across models for droplet expansion ratio

Supplementary Fig.20 presents a detailed comparison of 16 models evaluated over 15 iterations, using coefficient of determination ($R^2$) as the performance metric. The deep neural network (DNN) combined with SentenceTransformer exhibits the highest increase in $R^2$ values, indicating a better fit with increased repetitions. DNN paired with DistilBERT and GPT-2 shows moderate improvement in $R^2$. Other models maintain relatively low $R^2$ values, indicating less effective.

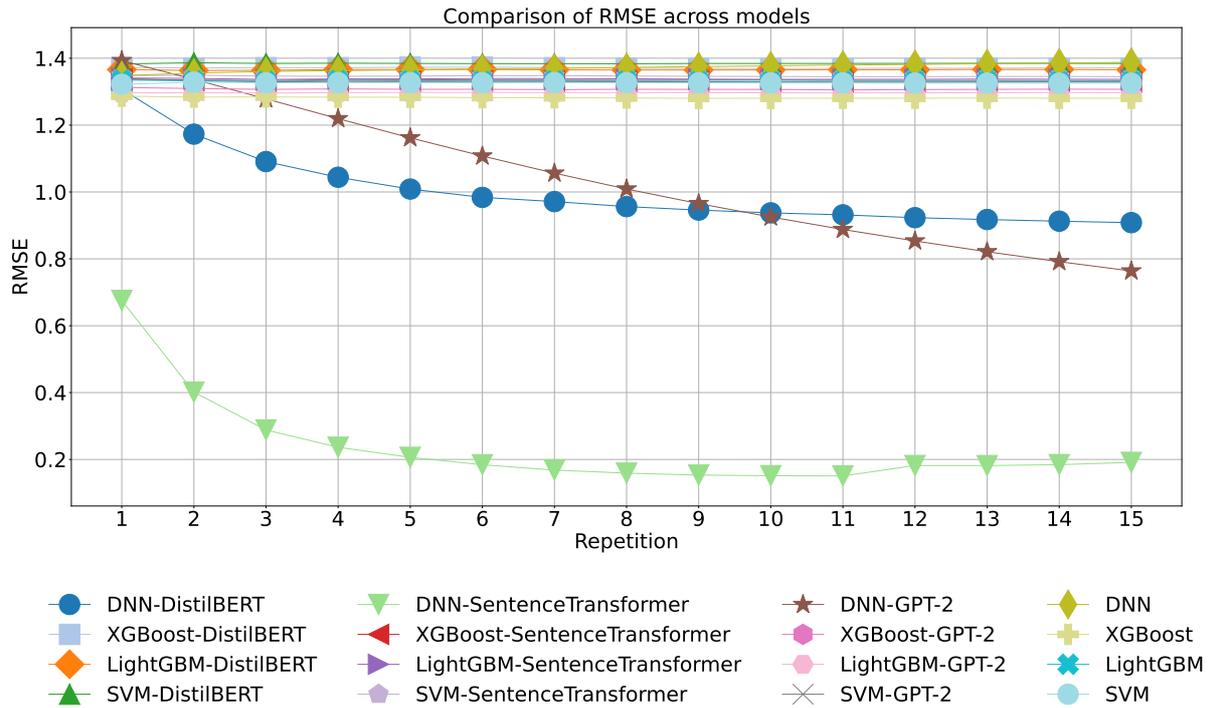

Supplementary Fig.21. Comparison of root mean squared error (RMSE) across models for droplet expansion ratio

Supplementary Fig.21 presents a detailed comparison of 16 models evaluated over 15 iterations, using root mean squared error (RMSE) as the performance metric. The deep neural network (DNN) combined with SentenceTransformer shows the greatest decrease in RMSE, confirming its improving accuracy. DNN paired with DistilBERT and GPT-2 follows with a slower decrease. Other models remain stable with higher RMSE values.

Overall, DNN combined with SentenceTransformer consistently outperforms other models across all metrics, showing significant improvement with more repetitions. DNN paired with DistilBERT also improves. The remaining models show minimal change, indicating stable but less optimal performance.

| Droplet Expansion Ratio | | | | |
|---|---|---|---|---|
| Model | Metrics | | | |
| | MAE | MSE | RMSE | R² |
| DNN | 1.1633 ± 0.0139 | 1.8233 ± 0.0648 | 1.3629 ± 0.0100 | 0.0639 ± 0.0139 |
| DNN-DistilBERT | 0.7467 ± 0.0104 | 0.8452 ± 0.0240 | 1.0909 ± 0.0338 | 0.1329 ± 0.0423 |
| DNN-OpenGPT-2 | 0.8984 ± 0.0236 | 1.7985 ± 0.0643 | 1.2786 ± 0.0244 | 0.2354 ± 0.0321 |
| DNN-SentenceTransformer | 0.1021 ± 0.0179 | 0.0786 ± 0.0335 | 0.1506 ± 0.0225 | 0.9592 ± 0.0175 |
| LightGBM | 1.1109 ± 0.0061 | 1.7767 ± 0.0162 | 1.3308 ± 0.0061 | 0.1007 ± 0.0155 |
| LightGBM-DistilBERT | 1.1674 ± 0.0103 | 1.8706 ± 0.0242 | 1.3663 ± 0.0088 | 0.0600 ± 0.0053 |
| LightGBM-OpenGPT-2 | 1.0968 ± 0.0078 | 1.6855 ± 0.0158 | 1.2974 ± 0.0073 | 0.1494 ± 0.0098 |
| LightGBM-SentenceTransformer | 1.1462 ± 0.0155 | 1.8000 ± 0.0408 | 1.3408 ± 0.0152 | 0.0954 ± 0.0056 |
| SVM | 1.0991 ± 0.0067 | 1.7628 ± 0.0169 | 1.3257 ± 0.0064 | 0.1103 ± 0.0116 |
| SVM-DistilBERT | 1.1697 ± 0.0122 | 1.9209 ± 0.0252 | 1.3845 ± 0.0091 | 0.0328 ± 0.0061 |
| SVM-OpenGPT-2 | 1.1072 ± 0.0061 | 1.7918 ± 0.0327 | 1.3358 ± 0.0122 | 0.0997 ± 0.0088 |
| SVM-SentenceTransformer | 1.1376 ± 0.0126 | 1.8161 ± 0.0261 | 1.3466 ± 0.0098 | 0.0809 ± 0.0094 |
| XGBoost | 1.0923 ± 0.0081 | 1.6471 ± 0.0189 | 1.2823 ± 0.0074 | 0.1672 ± 0.0083 |
| XGBoost-DistilBERT | 1.1742 ± 0.0104 | 1.8856 ± 0.0291 | 1.3719 ± 0.0106 | 0.0524 ± 0.0054 |
| XGBoost-OpenGPT-2 | 1.1114 ± 0.0057 | 1.7135 ± 0.0234 | 1.3094 ± 0.0144 | 0.1391 ± 0.0063 |
| XGBoost-SentenceTransformer | 1.1410 ± 0.0093 | 1.7705 ± 0.0233 | 1.3297 ± 0.0087 | 0.1058 ± 0.0068 |

Supplementary table 2. Metrics evaluation across models for droplet expansion ratio

Supplementary table 2 provides a detailed performance comparison of various machine learning models (DNN, LightGBM, SVM, XGBoost) combined with natural language processing models (DistilBERT, OpenGPT-2, SentenceTransformer) across regression tasks, using metrics such as mean absolute error (MAE), mean squared error (MSE), root mean squared error (RMSE), and R². Notably, DNN-SentenceTransformer exhibits the highest performance, with a remarkable R² of 0.9592, significantly outperforming other combinations. This suggests that SentenceTransformer integrates more effectively with DNN for this task, producing highly accurate predictions (MAE = 0.1021). On the other hand, most models, especially those incorporating DistilBERT and OpenGPT-2, tend to underperform. For instance, XGBoost-DistilBERT shows a relatively high error (MAE = 1.1742, RMSE = 1.3719) and a poor R² (0.0524), indicating that these combinations do not significantly improve model performance. Interestingly, the base versions of LightGBM and XGBoost also perform competitively, with XGBoost achieving a balanced performance (MAE = 1.0923, R² = 0.1672). Overall, the table highlights that while some NLP models such as SentenceTransformer provide a substantial boost to performance, others may introduce complexity without enhancing predictive accuracy, underscoring the importance of careful model selection and optimization.

## 2.4 Droplet Flow Rate Ratio

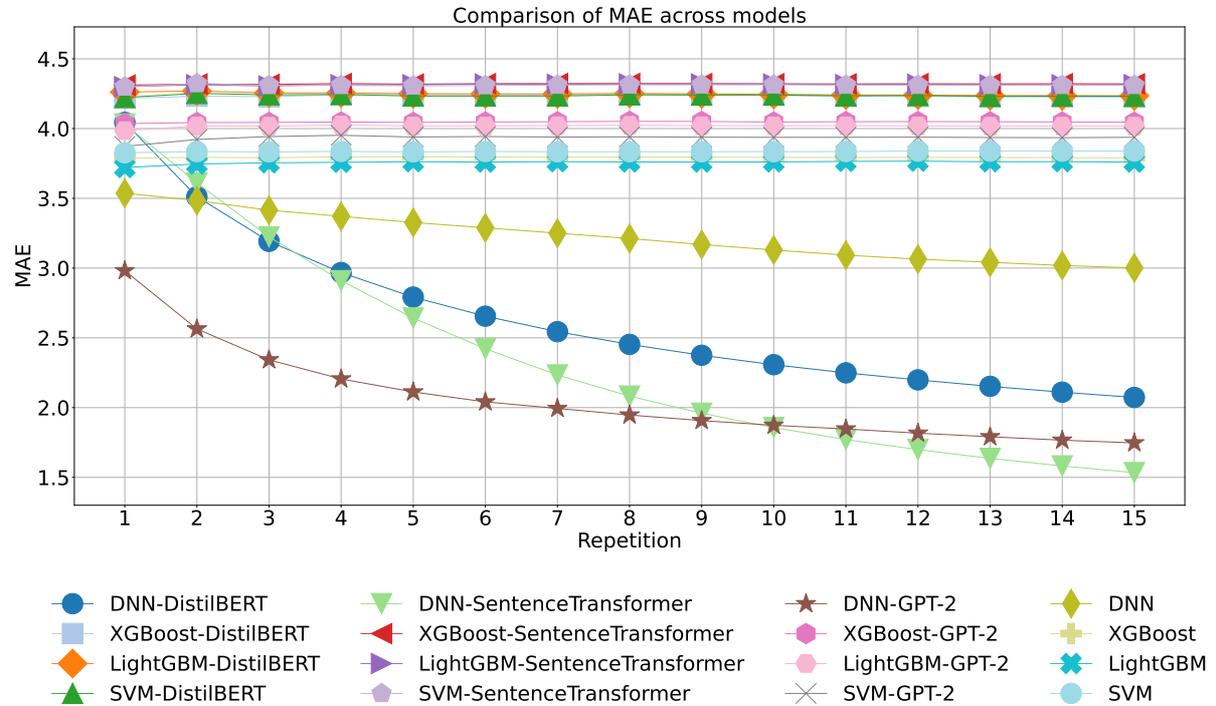

Supplementary Fig.22. Comparison of mean absolute error (MAE) across models for droplet flow rate ratio

Supplementary Fig.22 presents a detailed comparison of 16 models evaluated over 15 iterations, using mean absolute error (MAE) as the performance metric. The deep neural network paired with GPT-2 shows a significant decrease in MAE over repetitions, indicating improved accuracy. DNN combined with SentenceTransformer also shows improvement. The remaining models show minimal change.

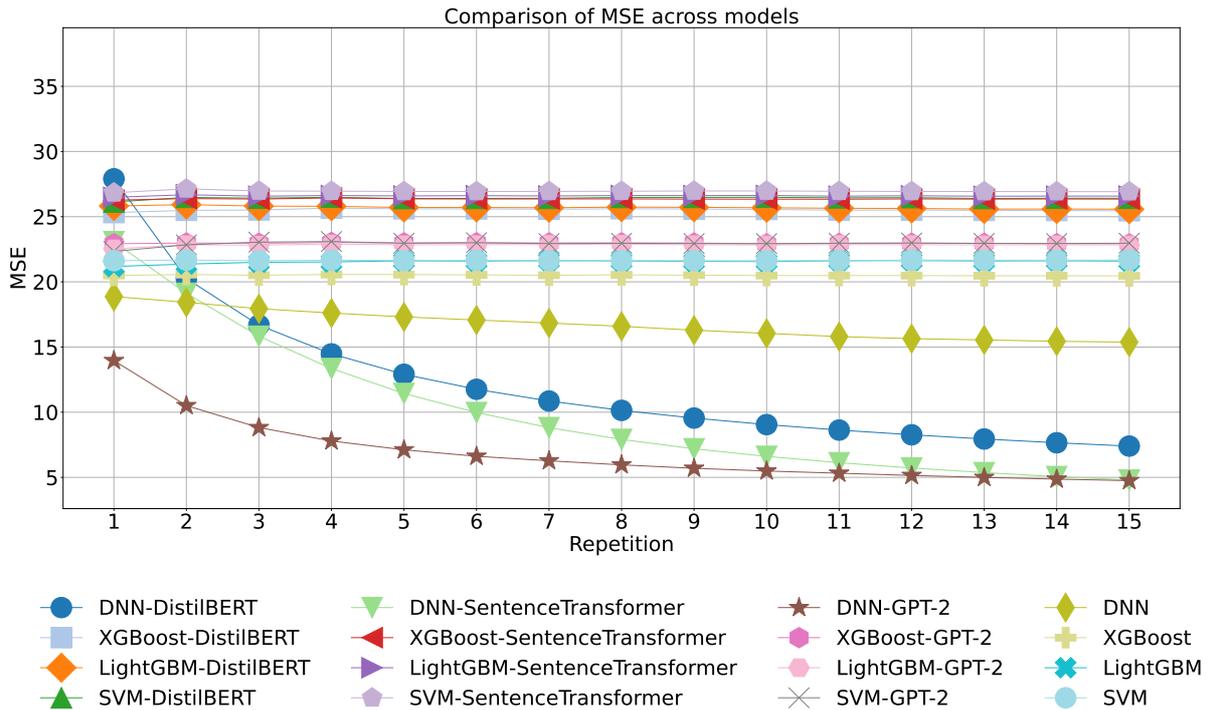

Supplementary Fig.23. Comparison of mean squared error (MSE) across models for droplet flow rate ratio

Supplementary Fig.23 presents a detailed comparison of 16 models evaluated over 15 iterations, using mean squared error (MSE) as the performance metric. Combination of deep neural network (DNN) with GPT-2 and SentenceTransformer demonstrates the most substantial reduction in MSE, suggesting enhanced predictive accuracy. DNN paired with DistilBERT follows with a noticeable decrease. Other models show higher and more stable MSE values, indicating less change over repetitions.

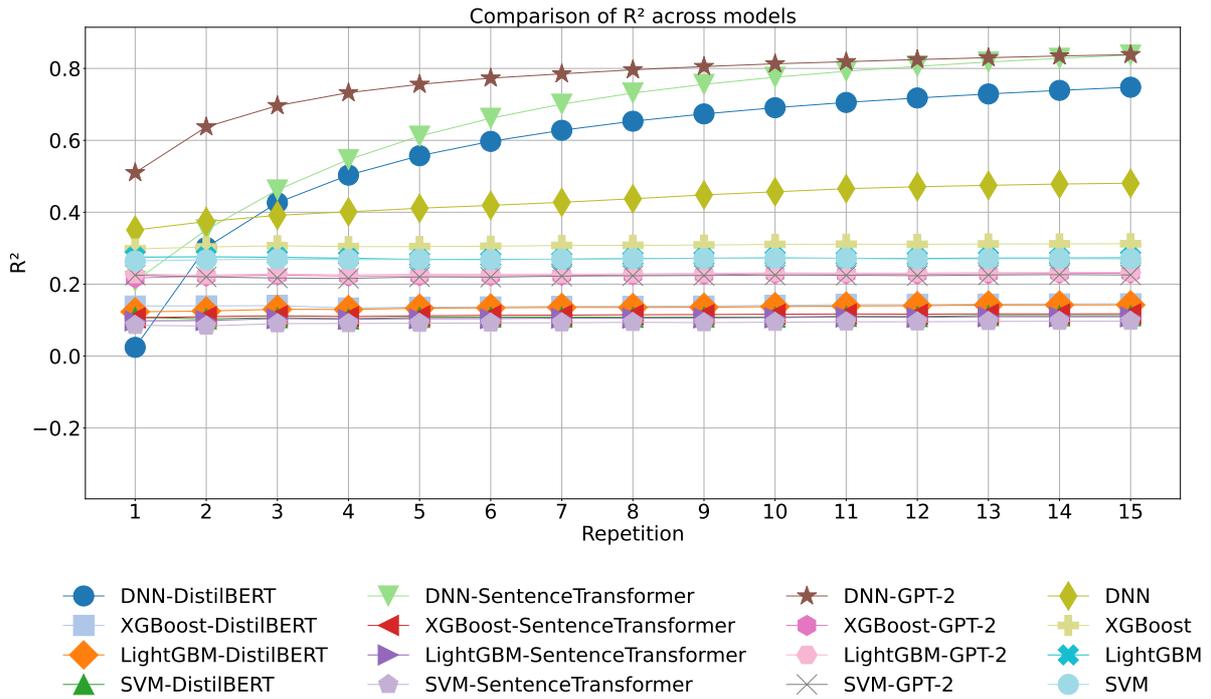

Supplementary Fig.24. Comparison of coefficient of determination (R²) across models for droplet flow rate ratio

Supplementary Fig.24 presents a detailed comparison of 16 models evaluated over 15 iterations, using coefficient of determination (R²) as the performance metric. Combination of deep neural network (DNN) with GPT-2 and SentenceTransformer exhibits the highest increase in R², signaling a better fit with more repetitions. DNN paired with DistilBERT also shows improvement. Others maintain relatively low R² values, indicating less effective.

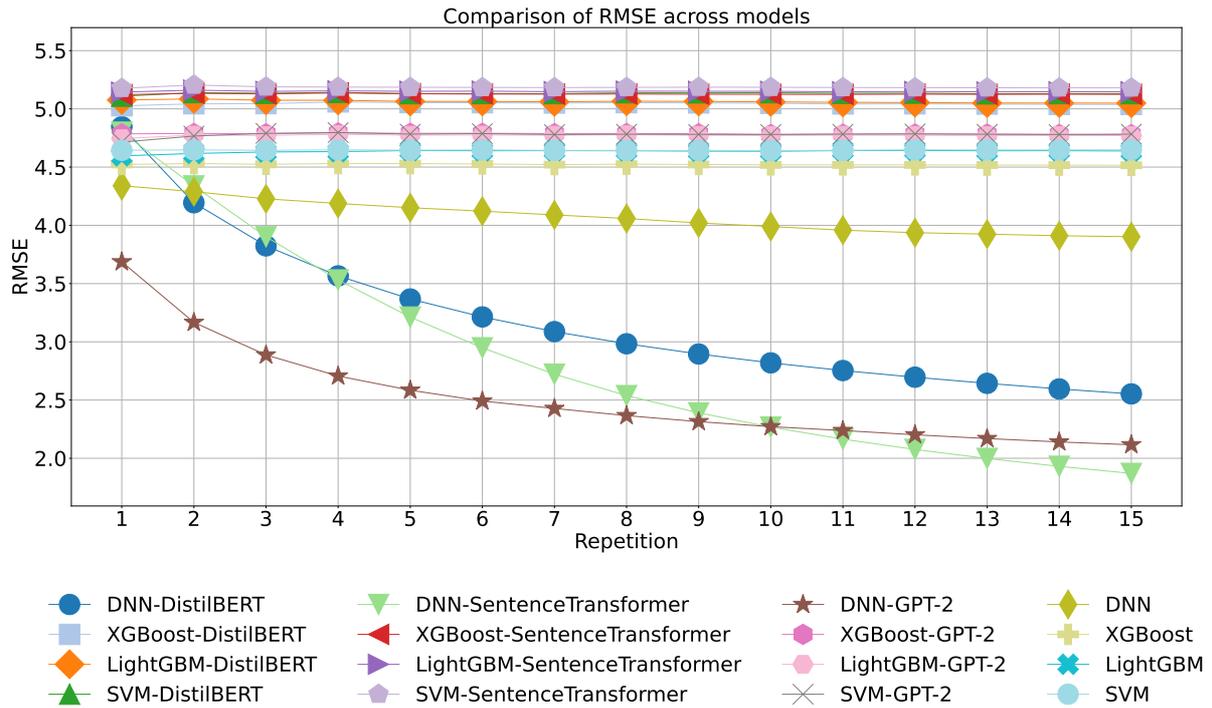

Supplementary Fig.25. Comparison of root mean squared error (RMSE) across models for droplet flow rate ratio

Supplementary Fig.25 presents a detailed comparison of 16 models evaluated over 15 iterations, using root mean squared error (RMSE) as the performance metric. The deep neural network (DNN) paired with SentenceTransformer shows the greatest decrease in RMSE, confirming its improving accuracy. DNN paired with DistilBERT and GPT-2 also decrease steadily. Other models remain stable with higher RMSE values.

Overall, combination of deep neural network (DNN) with GPT-2 and SentenceTransformer consistently outperforms other models across all metrics, showing significant improvement with more repetitions. DNN paired with DistilBERT improves as well, but not as dramatically. The remaining models show minimal change, indicating stable but less optimal performance.

| Droplet Flow Rate Ratio | | | | |
|---|---|---|---|---|
| Model | Metrics | | | |
| | MAE | MSE | RMSE | $R^2$ |
| DNN | 3.3707 ± 0.0378 | 17.3107 ± 0.3242 | 4.1514 ± 0.0392 | 0.4277 ± 0.0143 |
| DNN-DistilBERT | 2.792 ± 0.146 | 7.4 ± 0.8237 | 3.2137 ± 0.1543 | 0.7477 ± 0.0297 |
| DNN-OpenGPT-2 | 1.8469 ± 0.0416 | 4.7521 ± 0.2412 | 2.1161 ± 0.0428 | 0.8384 ± 0.0089 |
| DNN-SentenceTransformer | 4.0332 ± 0.0532 | 23.0771 ± 0.571 | 4.8002 ± 0.0592 | 0.2084 ± 0.0441 |
| LightGBM | 3.7646 ± 0.0234 | 21.1782 ± 0.6478 | 4.6299 ± 0.0477 | 0.272 ± 0.0091 |
| LightGBM-DistilBERT | 4.2453 ± 0.0271 | 25.9238 ± 0.5759 | 5.0854 ± 0.0558 | 0.1425 ± 0.0062 |
| LightGBM-OpenGPT-2 | 3.9849 ± 0.0405 | 22.8623 ± 0.294 | 4.7746 ± 0.0305 | 0.2316 ± 0.0086 |
| LightGBM-SentenceTransformer | 4.3172 ± 0.0328 | 26.5776 ± 0.2527 | 5.1607 ± 0.05 | 0.1062 ± 0.0087 |
| SVM | 3.8355 ± 0.022 | 21.629 ± 0.292 | 4.6488 ± 0.0209 | 0.2646 ± 0.0317 |
| SVM-DistilBERT | 4.2366 ± 0.0267 | 26.4881 ± 0.3944 | 5.1301 ± 0.0364 | 0.1079 ± 0.0124 |
| SVM-OpenGPT-2 | 3.9425 ± 0.0515 | 22.9605 ± 0.3874 | 4.7823 ± 0.033 | 0.2224 ± 0.0158 |
| SVM-SentenceTransformer | 4.3133 ± 0.0279 | 26.931 ± 0.3149 | 5.1841 ± 0.0306 | 0.0907 ± 0.0158 |
| XGBoost | 3.7952 ± 0.0235 | 20.4703 ± 0.1959 | 4.5171 ± 0.0212 | 0.3109 ± 0.0072 |
| XGBoost-DistilBERT | 4.2383 ± 0.0273 | 25.5001 ± 0.2716 | 5.0418 ± 0.027 | 0.1456 ± 0.007 |
| XGBoost-OpenGPT-2 | 4.0458 ± 0.0209 | 22.9177 ± 0.2481 | 4.7885 ± 0.0269 | 0.228 ± 0.0078 |
| XGBoost-SentenceTransformer | 4.3213 ± 0.0256 | 26.3464 ± 0.2421 | 5.1247 ± 0.0236 | 0.1154 ± 0.0075 |

Supplementary table 3. Metrics evaluation across models for droplet flow rate ratio

Supplementary table 3 presents a comprehensive evaluation of machine learning models (DNN, LightGBM, SVM, XGBoost) paired with various natural language processing models (DistilBERT, OpenGPT-2, SentenceTransformer), based on performance metrics such as mean absolute error (MAE), mean squared error (MSE), root mean squared error (RMSE), and $R^2$. Among the models, DNN-OpenGPT-2 exhibits the best overall performance, with the lowest MAE (1.8469), MSE (4.7521), and RMSE (2.1161), along with a high $R^2$ value (0.8384), indicating strong predictive power. In contrast, models like DNN-SentenceTransformer show poor performance with a high MSE (23.0771) and RMSE (4.8002), as well as an $R^2$ of 0.2084, reflecting weak predictive capabilities. Interestingly, adding NLP models like DistilBERT and SentenceTransformer to models such as LightGBM and SVM tends to degrade performance, with higher error rates and lower $R^2$ values, indicating that these combinations may not be optimal for enhancing predictive accuracy. For example, SVM-DistilBERT yields a high MAE (4.2366) and a low $R^2$ (0.1079). These results highlight the importance of carefully selecting and optimizing both machine learning models and NLP components to achieve the best results.

## 2.5 Droplet Normalized Oil Inlet

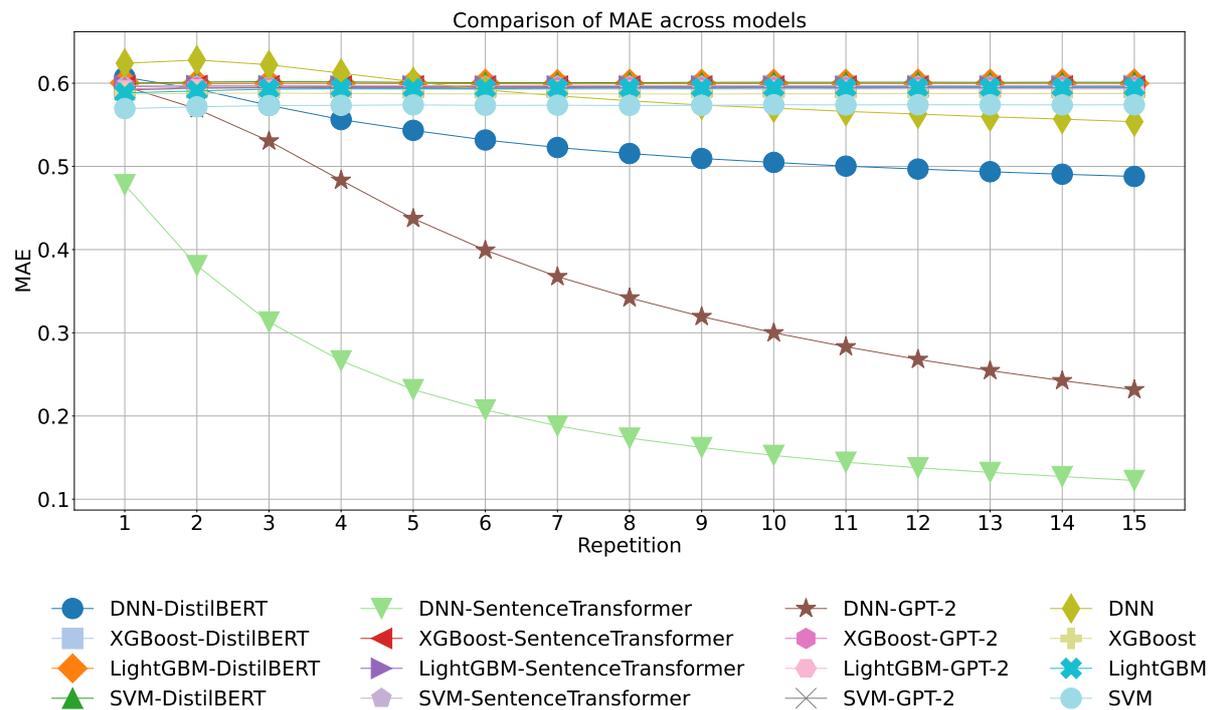

Supplementary Fig.26. Comparison of mean absolute error (MAE) across models for droplet normalized oil inlet

Supplementary Fig.26 presents a detailed comparison of 16 models evaluated over 15 iterations, using mean absolute error (MAE) as the performance metric. Combination of deep neural network (DNN) with SentenceTransformer has the lowest MAE, decreasing from around 0.5 to 0.1 as repetitions increase. DNN paired with GPT-2 starts at about 0.5 and decreases to about 0.3, showing improvement. DNN paired with DistilBERT remains relatively stable, starting at about 0.6 and slightly decreasing to below 0.5.

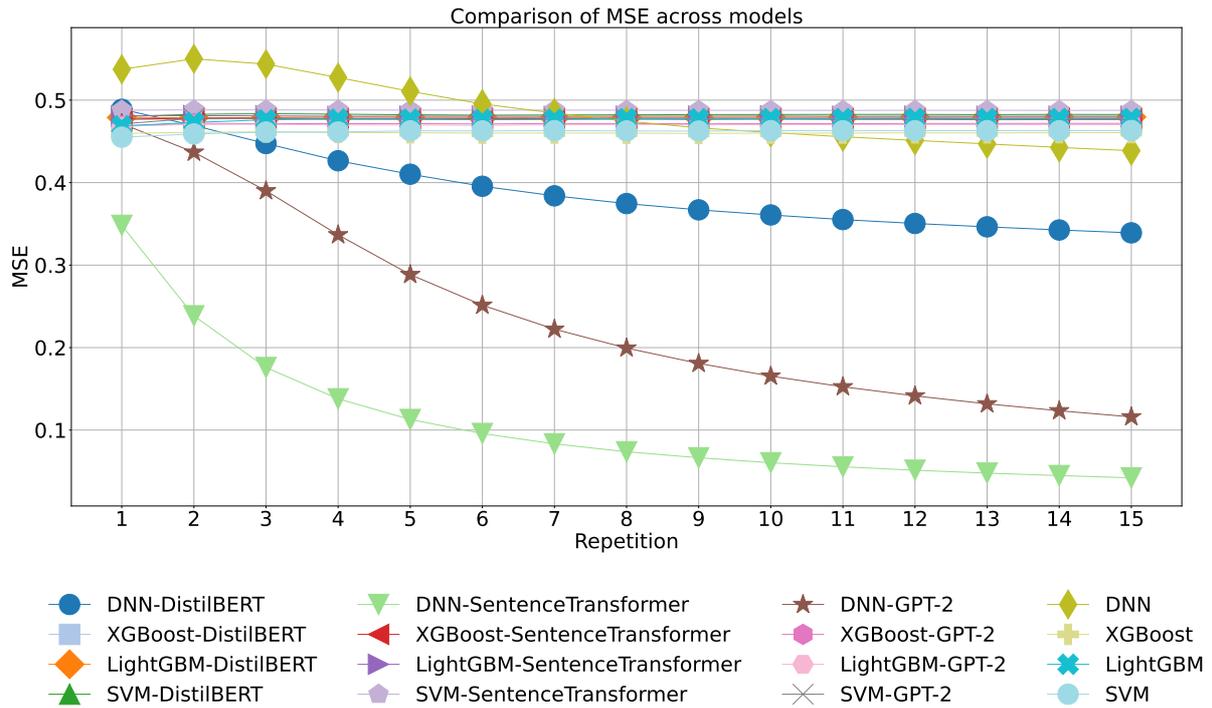

Supplementary Fig.27. Comparison of mean squared error (MSE) across models for droplet normalized oil inlet.

Supplementary Fig.27 presents a detailed comparison of 16 models evaluated over 15 iterations, using mean squared error (MSE) as the performance metric. Combination of deep neural network (DNN) with SentenceTransformer shows a significant reduction from approximately 0.3 to 0.05, indicating strong improvement. DNN paired with GPT-2 decreases from around 0.4 to 0.1. DNN paired with DistilBERT shows a slight decrease from about 0.5 to 0.3.

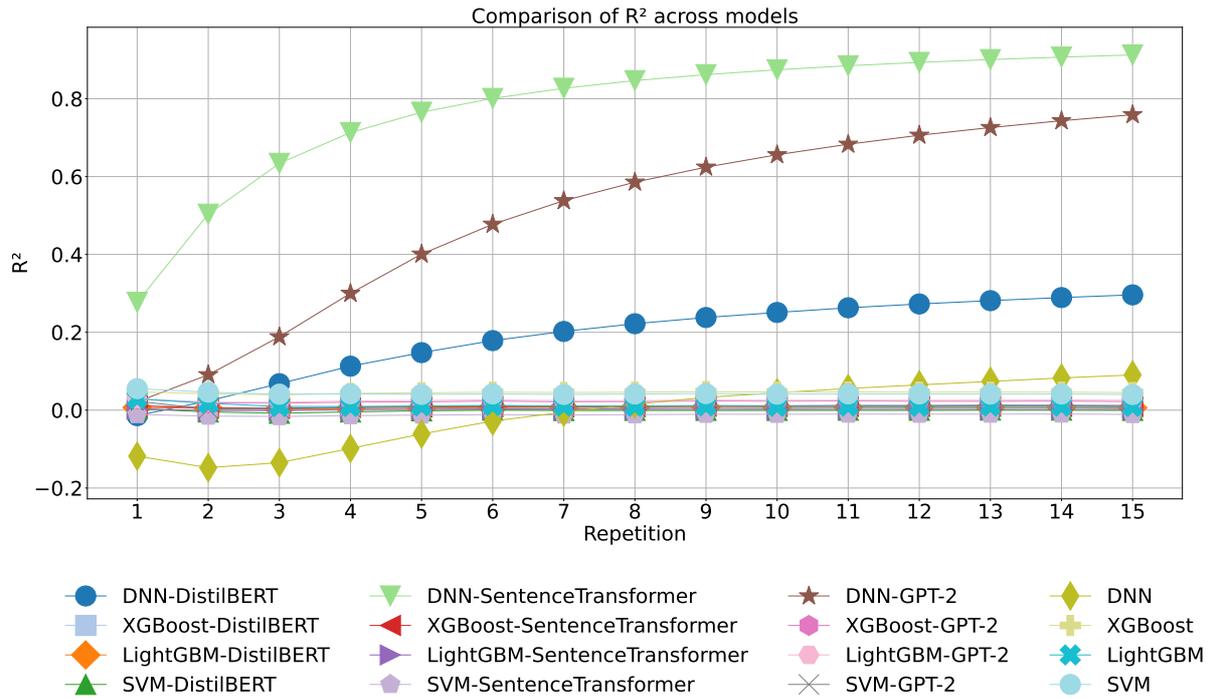

Supplementary Fig.28. Comparison of coefficient of determination ($R^2$) across models for droplet normalized oil inlet

Supplementary Fig.28 presents a detailed comparison of 16 models evaluated over 15 iterations, using coefficient of determination ($R^2$) as the performance metric. Combination of deep neural network (DNN) with SentenceTransformer achieves the highest $R^2$, increasing from about 0.3 to 0.9. DNN paired with GPT-2 shows improvement from around 0.2 to above 0.8. DNN paired with DistilBERT remains low.

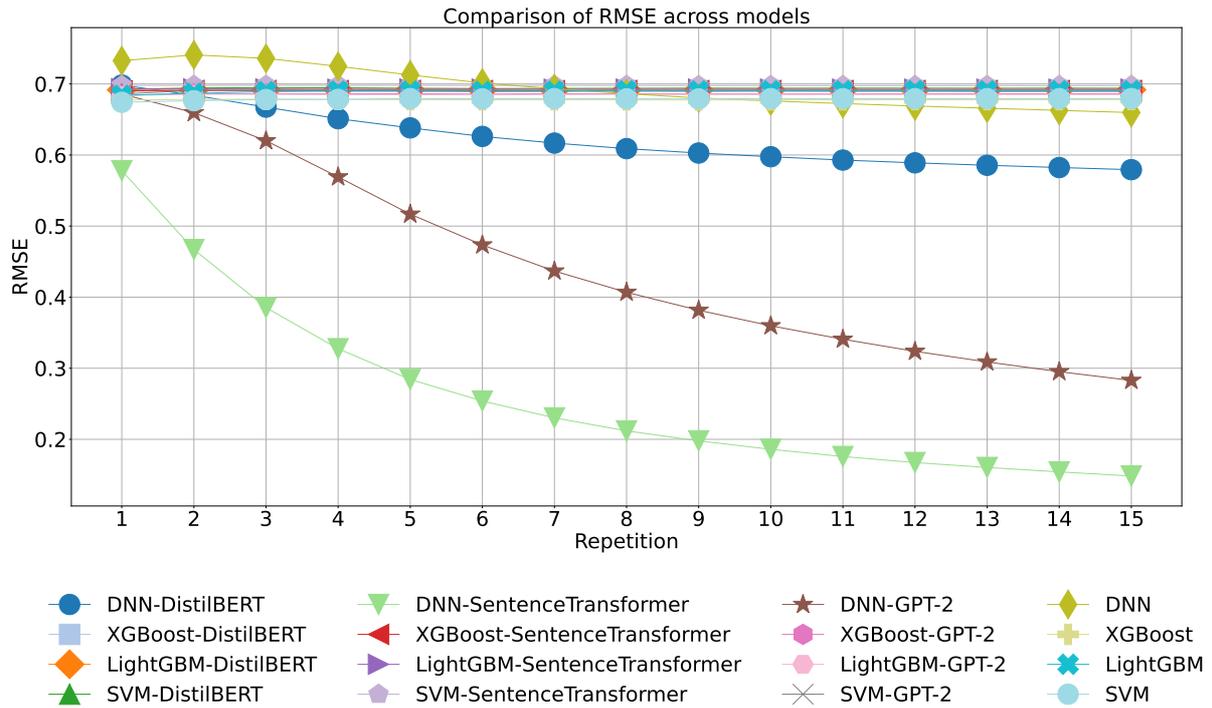

Supplementary Fig.29. Comparison of root mean squared error (RMSE) across models for droplet normalized oil inlet

Supplementary Fig.29 presents a detailed comparison of 16 models evaluated over 15 iterations, using root mean squared error (RMSE) as the performance metric. Combination of deep neural network (DNN) with SentenceTransformer decreases from approximately 0.6 to 0.15, confirming its improving accuracy. DNN paired with GPT-2 reduces from about 0.7 to around 0.3. DNN combined with DistilBERT remains relatively stable, with a slight decrease from 0.7 to 0.6.

Overall, combination of deep neural network (DNN) with SentenceTransformer consistently outperforms other models across all metrics, showing significant improvement with more repetitions. DNN paired with GPT-2 also improves but to a lesser extent. DNN combined with DistilBERT shows minimal change, indicating stable but less optimal performance.

| Droplet Normalized Oil Inlet | | | | |
|---|---|---|---|---|
| Model | Metrics | | | |
| | MAE | MSE | RMSE | R² |
| DNN | 0.592 ± 0.007 | 0.5374 ± 0.0104 | 0.7327 ± 0.0071 | -0.0052 ± 0.0194 |
| DNN-DistilBERT | 0.5433 ± 0.0076 | 0.4102 ± 0.0099 | 0.6511 ± 0.008 | -0.0141 ± 0.0247 |
| DNN-OpenGPT-2 | 0.5691 ± 0.0093 | 0.4369 ± 0.0125 | 0.6596 ± 0.0094 | 0.0909 ± 0.0222 |
| DNN-SentenceTransformer | 0.4779 ± 0.0322 | 0.3482 ± 0.0441 | 0.578 ± 0.0375 | 0.2771 ± 0.0878 |
| LightGBM | 0.5934 ± 0.0043 | 0.4688 ± 0.0135 | 0.6904 ± 0.0032 | 0.01 ± 0.0056 |
| LightGBM-DistilBERT | 0.5995 ± 0.0042 | 0.4792 ± 0.0042 | 0.6916 ± 0.003 | 0.0049 ± 0.0079 |
| LightGBM-OpenGPT-2 | 0.5926 ± 0.0038 | 0.4704 ± 0.0035 | 0.6851 ± 0.0025 | 0.0201 ± 0.0066 |
| LightGBM-SentenceTransformer | 0.5962 ± 0.0027 | 0.4805 ± 0.0035 | 0.6926 ± 0.0026 | 0.0027 ± 0.003 |
| SVM | 0.5695 ± 0.0061 | 0.4551 ± 0.0098 | 0.6742 ± 0.0073 | 0.0425 ± 0.0072 |
| SVM-DistilBERT | 0.601 ± 0.0032 | 0.4818 ± 0.0048 | 0.694 ± 0.0035 | -0.0006 ± 0.0052 |
| SVM-OpenGPT-2 | 0.5944 ± 0.0028 | 0.4764 ± 0.0035 | 0.6895 ± 0.0025 | 0.0229 ± 0.0132 |
| SVM-SentenceTransformer | 0.597 ± 0.0039 | 0.488 ± 0.0055 | 0.6981 ± 0.004 | -0.011 ± 0.0053 |
| XGBoost | 0.5872 ± 0.0031 | 0.4601 ± 0.0038 | 0.6782 ± 0.007 | 0.0463 ± 0.0033 |
| XGBoost-DistilBERT | 0.5972 ± 0.0032 | 0.4784 ± 0.0034 | 0.6907 ± 0.004 | 0.0085 ± 0.0021 |
| XGBoost-OpenGPT-2 | 0.5948 ± 0.003 | 0.4719 ± 0.0034 | 0.6863 ± 0.0025 | 0.0283 ± 0.0114 |
| XGBoost-SentenceTransformer | 0.5964 ± 0.0031 | 0.477 ± 0.0097 | 0.6903 ± 0.0071 | 0.0103 ± 0.0022 |

Supplementary table 4. Metrics evaluation across models for droplet normalized oil inlet

Supplementary table 4 presents a comprehensive evaluation of machine learning models (DNN, LightGBM, SVM, XGBoost) paired with various natural language processing models (DistilBERT, OpenGPT-2, SentenceTransformer), based on performance metrics such as mean absolute error (MAE), mean squared error (MSE), root mean squared error (RMSE), and R². Notably, the DNN-SentenceTransformer combination achieves the lowest MAE (0.4779) and the highest R² (0.2771), indicating superior performance in terms of accuracy and variance explanation. Conversely, the DNN model shows a negative R² (-0.0052), suggesting poor predictive capability. While LightGBM and XGBoost models generally perform well, the addition of embeddings does not consistently enhance their predictive power. The standard errors across models remain relatively low, indicating reliable estimates. However, the negative and low R² values in several configurations, such as SVM-DistilBERT.

2.6 Droplet Normalized Orifice Length

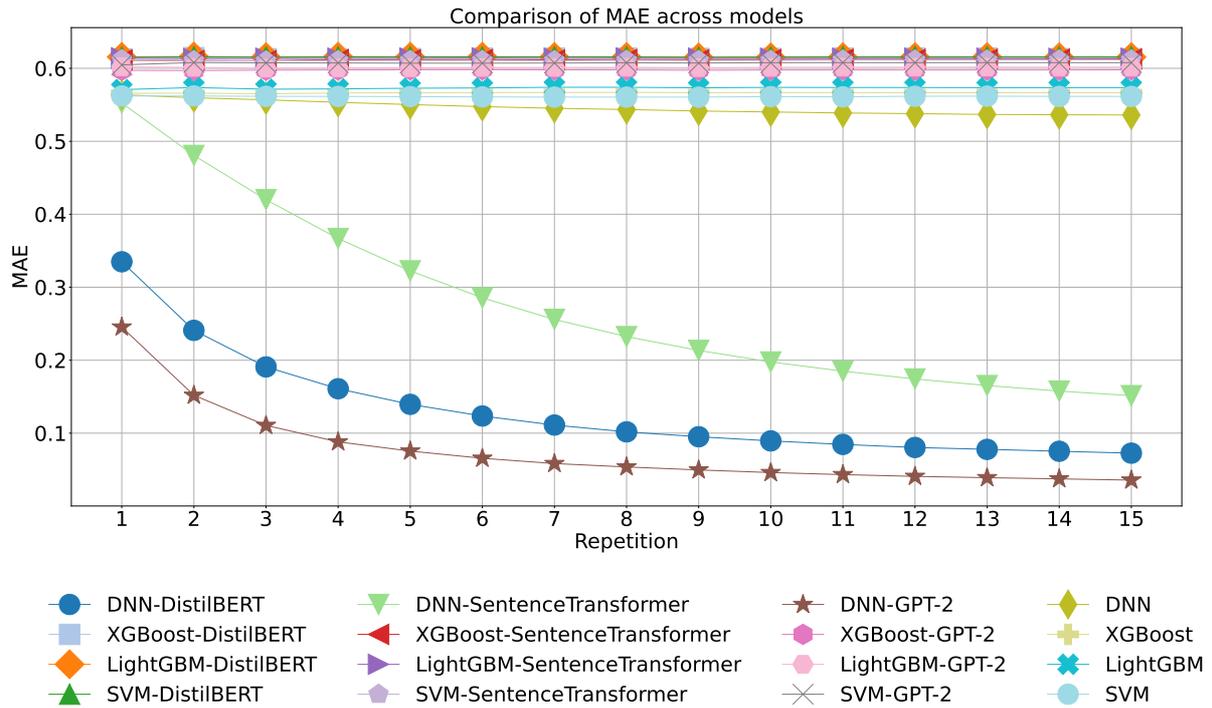

Supplementary Fig.30. Comparison of mean absolute error (MAE) across models for droplet normalized orifice length

Supplementary Fig.30 presents a detailed comparison of 16 models evaluated over 15 iterations, using mean absolute error (MAE) as the performance metric. Combination of deep neural network (DNN) with SentenceTransformer starts at approximately 0.6 and decreases to around 0.1 by the 15th repetition. DNN paired with GPT-2 begins at about 0.3 and reduces to just below 0.1. DNN integrated with DistilBERT decreases from around 0.35 to 0.15.

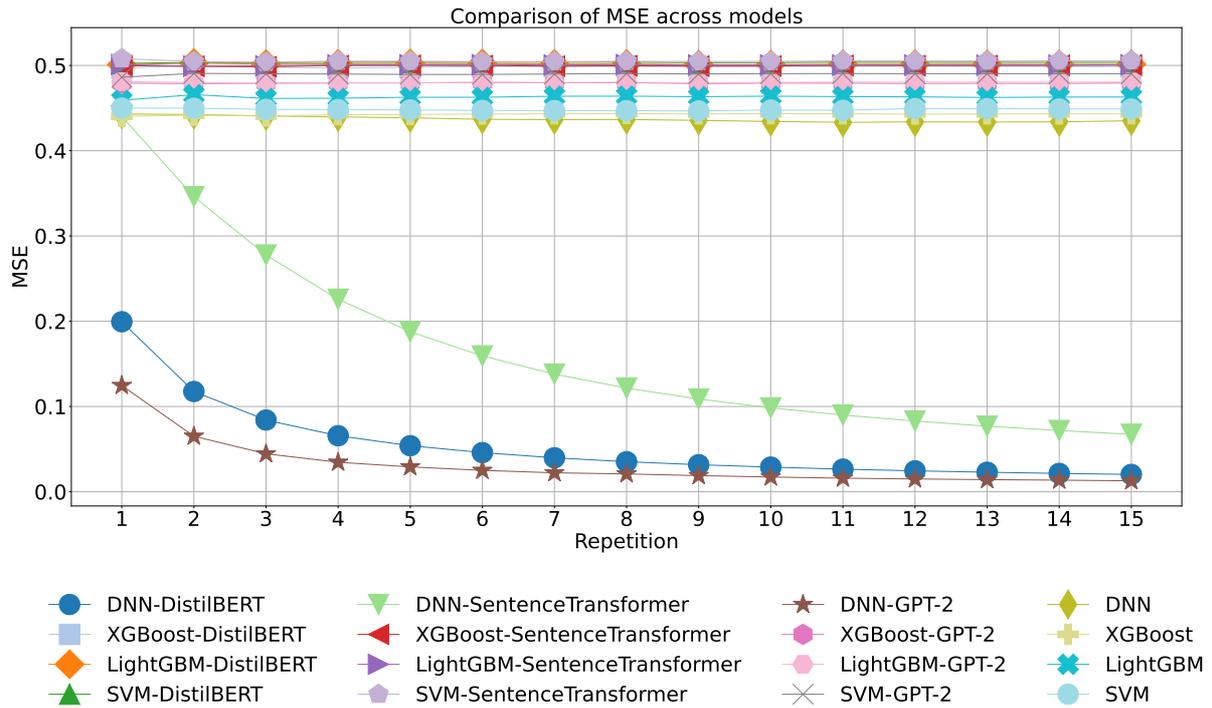

Supplementary Fig.31. Comparison of mean squared error (MSE) across models for droplet normalized orifice length

Supplementary Fig.31 presents a detailed comparison of 16 models evaluated over 15 iterations, using mean squared error (MSE) as the performance metric. Combination of deep neural network (DNN) with SentenceTransformer shows a significant drop from about 0.45 to below 0.1. DNN integrated with DistilBERT reduces from approximately 0.2 to below 0.05. DNN paired with GPT-2 shows a decrease from around 0.15 to below 0.05.

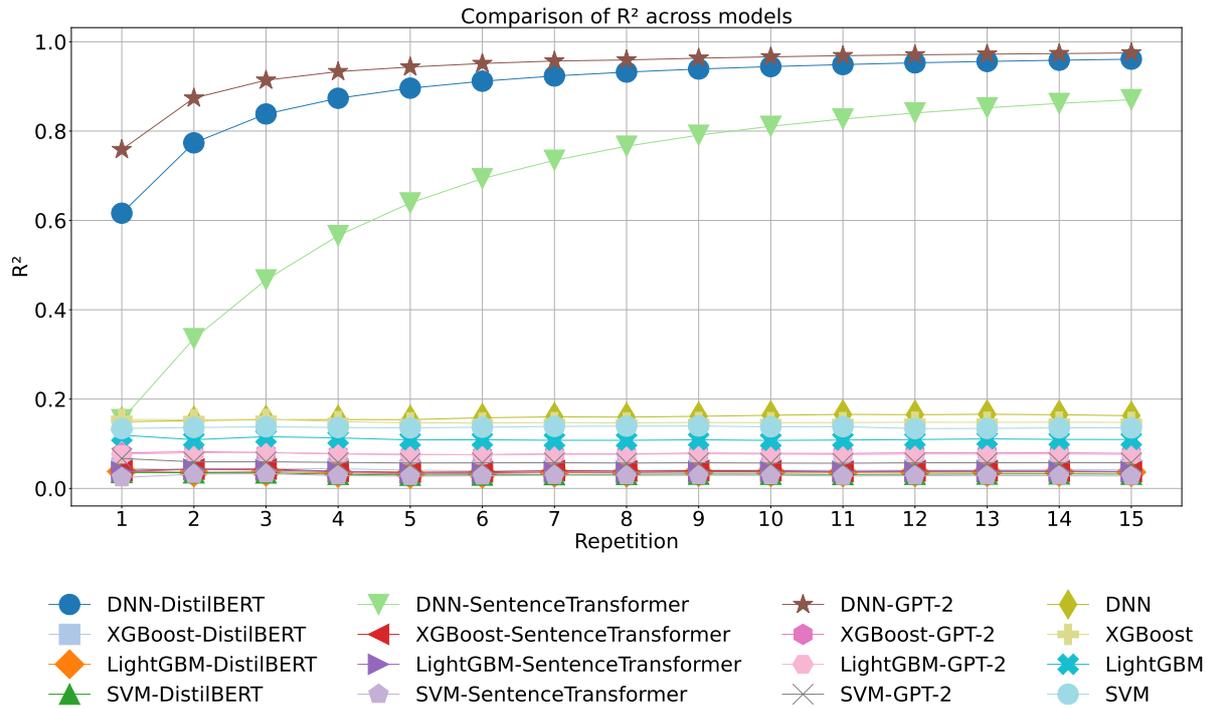

Supplementary Fig.32. Comparison of coefficient of determination (R²) across models for droplet normalized orifice length

Supplementary Fig.32 presents a detailed comparison of 16 models evaluated over 15 iterations, using coefficient of determination ($R^2$) as the performance metric. The deep neural network (DNN) paired with GPT-2 achieves the highest $R^2$, increasing from around 0.8 to nearly 1.0. DNN integrated with DistilBERT improves from about 0.6 to approximately 1. DNN integrated with SentenceTransformer increases from approximately 0.2 to around 0.9.

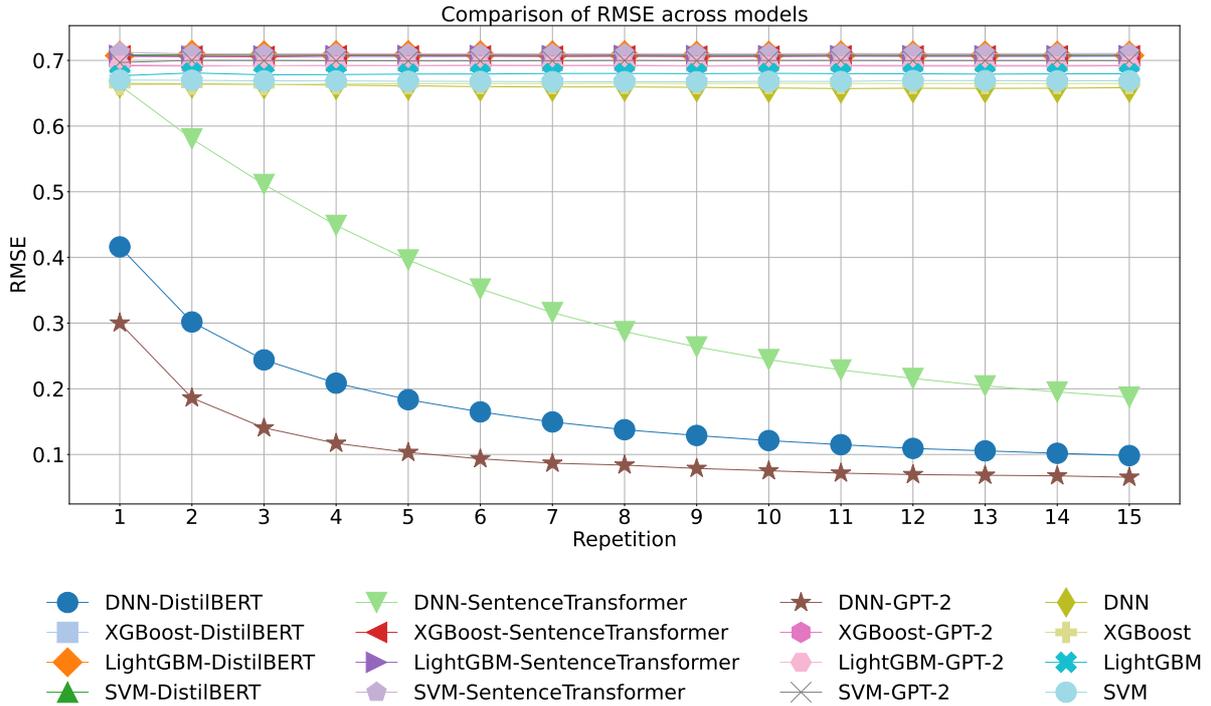

Supplementary Fig.33. Comparison of root mean squared error (RMSE) across models for droplet normalized orifice length

Supplementary Fig.33 presents a detailed comparison of 16 models evaluated over 15 iterations, using root mean squared error (RMSE) as the performance metric. The deep neural network (DNN) paired with DistilBERT decreases from around 0.4 to 0.1. DNN combined with GPT-2 shows a reduction from about 0.3 to below 0.1. DNN combined with SentenceTransformer declines from around 0.65 to around 0.2.

Overall, DNN combined with GPT-2 consistently outperforms other models across all metrics, demonstrating substantial improvements with increased repetitions. While the DNN paired with DistilBERT also shows significant enhancement, it is less pronounced compared to the performance of DNN with GPT-2. The DNN integrated with SentenceTransformer exhibits steady improvement, though it remains less optimal compared to the DNN models paired with GPT-2 and DistilBERT.

| Droplet Normalized Orifice Length | | | | |
|---|---|---|---|---|
| Model | Metrics | | | |
| | MAE | MSE | RMSE | R² |
| DNN | 0.5475 ± 0.0044 | 0.4336 ± 0.0043 | 0.6645 ± 0.0063 | 0.166 ± 0.0076 |
| DNN-DistilBERT | 0.0726 ± 0.0068 | 0.0203 ± 0.0051 | 0.1018 ± 0.0089 | 0.961 ± 0.0099 |
| DNN-OpenGPT-2 | 0.0357 ± 0.0057 | 0.0128 ± 0.004 | 0.0653 ± 0.0076 | 0.975 ± 0.008 |
| DNN-SentenceTransformer | 0.4806 ± 0.0188 | 0.346 ± 0.0246 | 0.5805 ± 0.0213 | 0.3353 ± 0.0471 |
| LightGBM | 0.5732 ± 0.0035 | 0.4632 ± 0.0043 | 0.6767 ± 0.0111 | 0.1075 ± 0.0081 |
| LightGBM-DistilBERT | 0.6153 ± 0.0027 | 0.5013 ± 0.0034 | 0.7075 ± 0.0024 | 0.0329 ± 0.0058 |
| LightGBM-OpenGPT-2 | 0.6017 ± 0.0027 | 0.4806 ± 0.0031 | 0.6927 ± 0.0023 | 0.0795 ± 0.0091 |
| LightGBM-SentenceTransformer | 0.6135 ± 0.0025 | 0.5006 ± 0.0042 | 0.7071 ± 0.0024 | 0.0373 ± 0.0044 |
| SVM | 0.5608 ± 0.0039 | 0.4478 ± 0.0064 | 0.6694 ± 0.0033 | 0.137 ± 0.0112 |
| SVM-DistilBERT | 0.6155 ± 0.0029 | 0.5036 ± 0.0034 | 0.709 ± 0.0024 | 0.0325 ± 0.0037 |
| SVM-OpenGPT-2 | 0.6076 ± 0.0031 | 0.4895 ± 0.005 | 0.6995 ± 0.0036 | 0.0567 ± 0.0048 |
| SVM-SentenceTransformer | 0.611 ± 0.0031 | 0.5057 ± 0.0032 | 0.7106 ± 0.0023 | 0.0291 ± 0.0055 |
| XGBoost | 0.5667 ± 0.0031 | 0.443 ± 0.0041 | 0.6649 ± 0.0031 | 0.1474 ± 0.0069 |
| XGBoost-DistilBERT | 0.6133 ± 0.0027 | 0.4986 ± 0.0034 | 0.7056 ± 0.0024 | 0.0441 ± 0.0064 |
| XGBoost-OpenGPT-2 | 0.5978 ± 0.0034 | 0.4795 ± 0.0048 | 0.6921 ± 0.0035 | 0.0794 ± 0.0048 |
| XGBoost-SentenceTransformer | 0.6106 ± 0.0047 | 0.5006 ± 0.0031 | 0.7068 ± 0.0063 | 0.0385 ± 0.0037 |

Supplementary table 5. Metrics evaluation across models for droplet normalized orifice length

Supplementary table 5 presents a comprehensive evaluation of machine learning models (DNN, LightGBM, SVM, XGBoost) paired with various natural language processing models (DistilBERT, OpenGPT-2, SentenceTransformer), based on performance metrics such as mean absolute error (MAE), mean squared error (MSE), root mean squared error (RMSE), and R². The DNN-OpenGPT-2 model exhibits exceptional performance, achieving the lowest MAE (0.0357) and the highest R² (0.975), indicating a strong predictive capability and minimal error. In contrast, LightGBM and SVM models with embeddings show varied results, often with higher errors and lower R² values.

2.7 Droplet Normalized Water Inlet

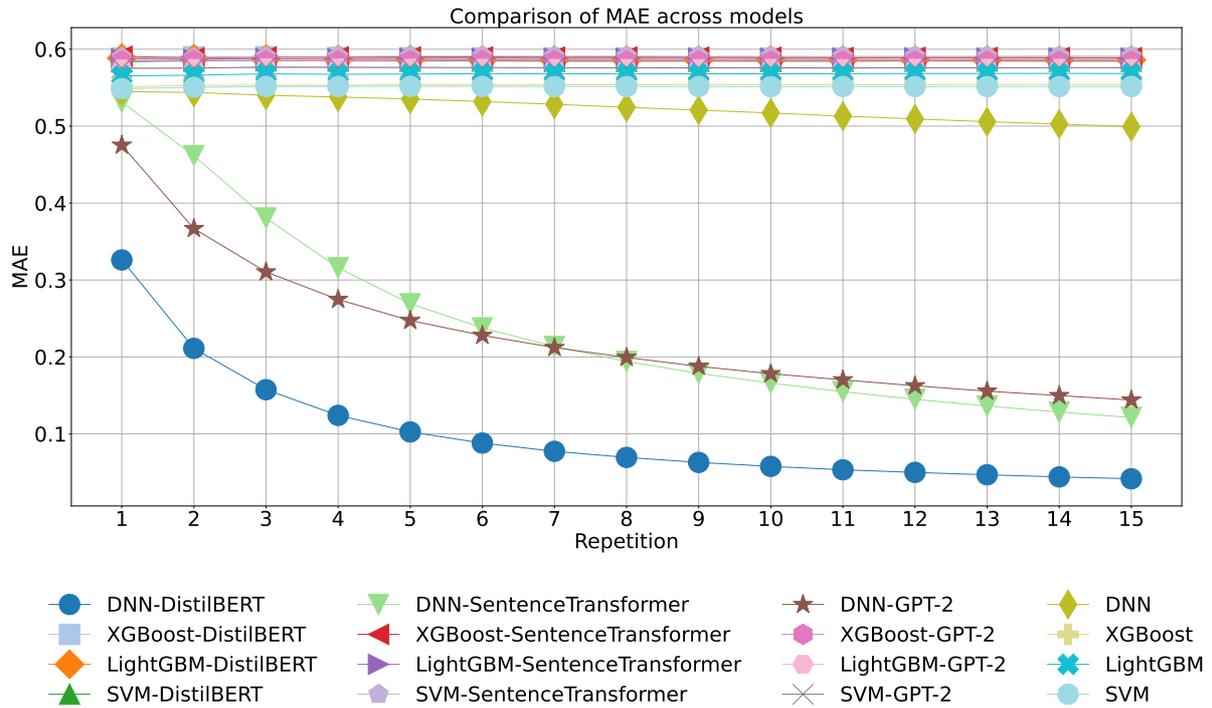

Supplementary Fig.34. Comparison of mean absolute error (MAE) across models for droplet normalized water inlet

Supplementary Fig.34 presents a detailed comparison of 16 models evaluated over 15 iterations, using mean absolute error (MAE) as the performance metric. The deep neural network (DNN) paired with DistilBERT shows a significant decrease in MAE from 0.3 to about 0.05 over the repetitions, indicating high accuracy and improvement. DNN combined with GPT-2 and SentenceTransformer also show a decrease. Other models remain relatively stable with higher MAE values around 0.6, indicating less improvement.

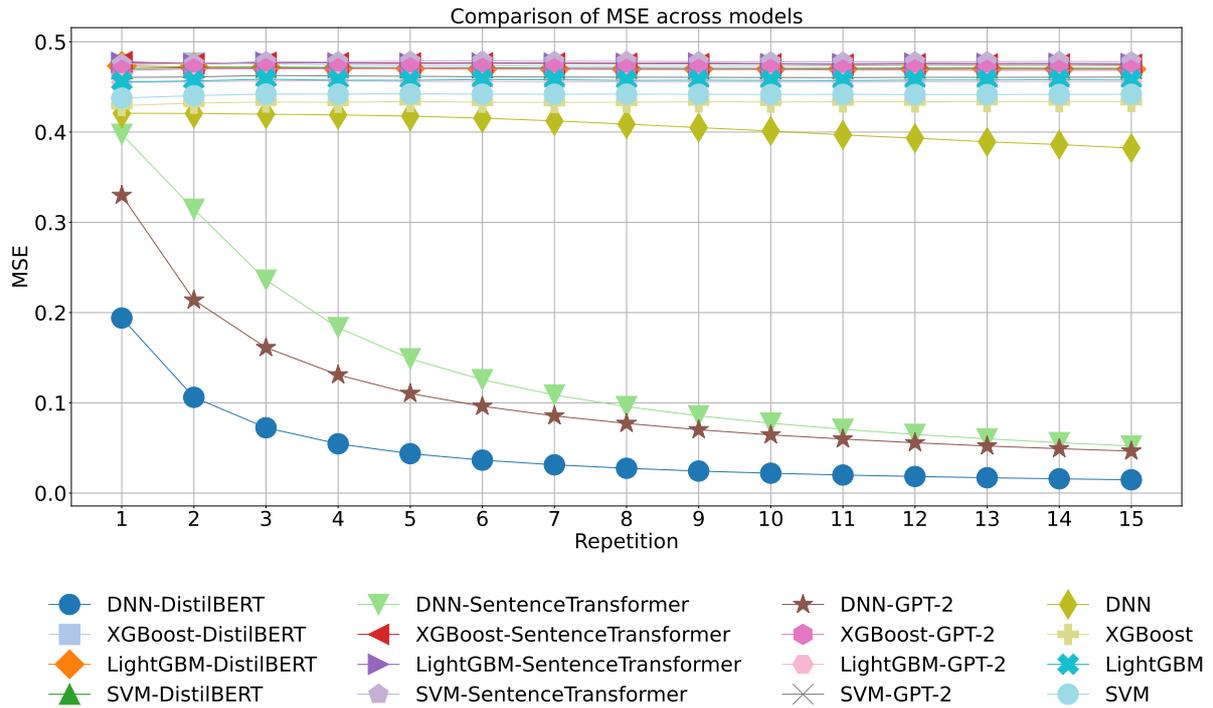

Supplementary Fig.35. Comparison of mean squared error (MSE) across models for droplet normalized water inlet

Supplementary Fig.35 presents a detailed comparison of 16 models evaluated over 15 iterations, using mean squared error (MSE) as the performance metric. Combination of deep neural network (DNN) with DistilBERT starts at 0.2 and decreases to approximately 0.02, highlighting strong performance and learning. DNN integrated with GPT-2 and SentenceTransformer show a gradual decrease to below 0.1, indicating moderate improvement. The rest of the models maintain higher MSE values between 0.4 and 0.5.

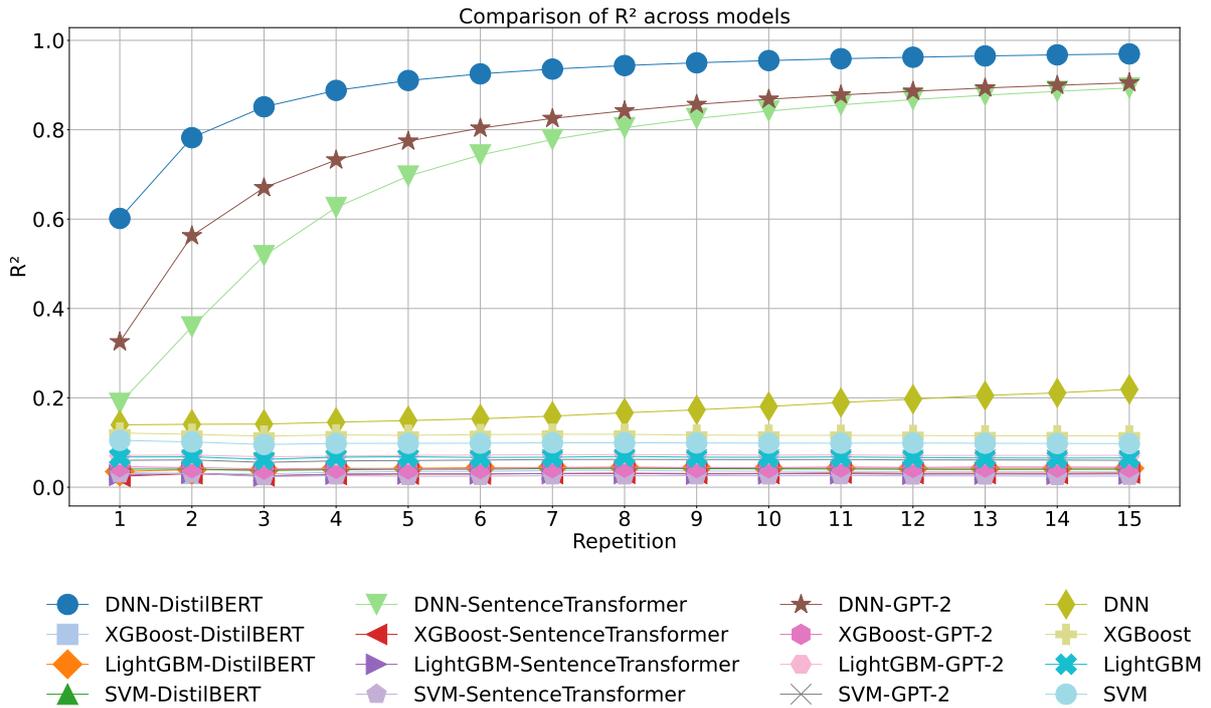

Supplementary Fig.36. Comparison of coefficient of determination (R²) across models for droplet normalized water inlet

Supplementary Fig.36 presents a detailed comparison of 16 models evaluated over 15 iterations, using coefficient of determination ($R^2$) as the performance metric. Combination of deep neural network (DNN) with DistilBERT shows a consistent increase from about 0.6 to nearly 1.0, indicating an excellent generalization over repetitions. DNN integrated with GPT-2 and SentenceTransformer increase to around 0.9, reflecting good predictive capability. Other models exhibit minimal change, maintaining lower $R^2$ values below 0.3.

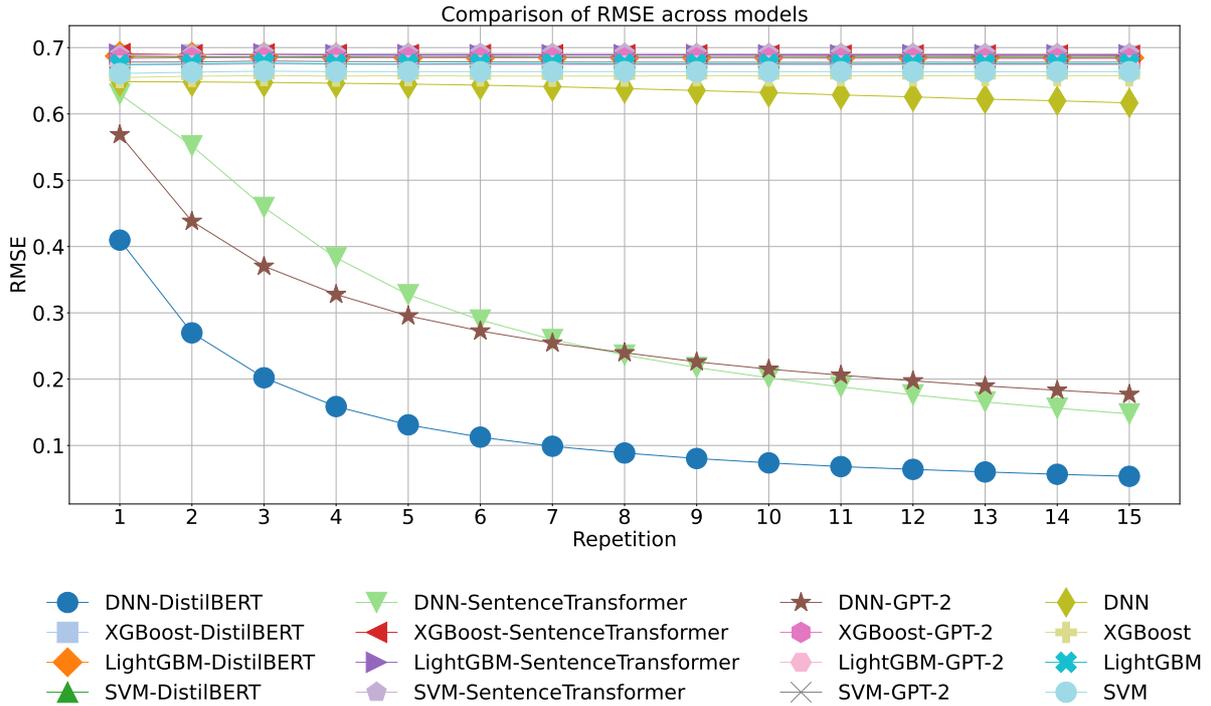

Supplementary Fig.37. Comparison of root mean squared error (RMSE) across models for droplet normalized water inlet

Supplementary Fig.37 presents a detailed comparison of 16 models evaluated over 15 iterations, using root mean squared error (RMSE) as the performance metric. Combination of deep neural network (DNN) with DistilBERT decreases from around 0.4 to about 0.05, showing strong predictive accuracy. DNN integrated with GPT-2 and SentenceTransformer decrease to approximately 0.2, reflecting moderate accuracy improvement. Remaining models stay from 0.6 to 0.7, indicating less accuracy.

Overall, combination of DNN with DistilBERT consistently outperforms others across all metrics, showing significant improvement and accuracy. DNN integrated with GPT-2 and SentenceTransformer show moderate improvements, while other models remain relatively stable with less accuracy.

| Droplet Normalized Water Inlet | | | | |
|---|---|---|---|---|
| Model | Metrics | | | |
| | MAE | MSE | RMSE | R² |
| DNN | 0.5245 ± 0.0044 | 0.405 ± 0.0051 | 0.6482 ± 0.005 | 0.1897 ± 0.0096 |
| DNN-DistilBERT | 0.0416 ± 0.0071 | 0.0147 ± 0.0049 | 0.0534 ± 0.0089 | 0.9697 ± 0.0102 |
| DNN-OpenGPT-2 | 0.4752 ± 0.0228 | 0.3299 ± 0.0279 | 0.5686 ± 0.0257 | 0.3254 ± 0.0574 |
| DNN-SentenceTransformer | 0.5314 ± 0.0093 | 0.3975 ± 0.0127 | 0.6297 ± 0.0103 | 0.3588 ± 0.0467 |
| LightGBM | 0.5683 ± 0.0034 | 0.4555 ± 0.011 | 0.6744 ± 0.0083 | 0.0689 ± 0.0086 |
| LightGBM-DistilBERT | 0.5876 ± 0.0088 | 0.4696 ± 0.0035 | 0.6847 ± 0.0027 | 0.0424 ± 0.0058 |
| LightGBM-OpenGPT-2 | 0.5753 ± 0.003 | 0.456 ± 0.0062 | 0.6744 ± 0.0041 | 0.0713 ± 0.0039 |
| LightGBM-SentenceTransformer | 0.5883 ± 0.0031 | 0.4758 ± 0.0039 | 0.6891 ± 0.003 | 0.0297 ± 0.0051 |
| SVM | 0.5514 ± 0.0032 | 0.4418 ± 0.0043 | 0.6637 ± 0.0031 | 0.0956 ± 0.0131 |
| SVM-DistilBERT | 0.5845 ± 0.0031 | 0.4708 ± 0.0037 | 0.6854 ± 0.0027 | 0.0401 ± 0.0036 |
| SVM-OpenGPT-2 | 0.5753 ± 0.0087 | 0.4606 ± 0.0108 | 0.6782 ± 0.008 | 0.0612 ± 0.0086 |
| SVM-SentenceTransformer | 0.5905 ± 0.0033 | 0.478 ± 0.007 | 0.691 ± 0.0048 | 0.0249 ± 0.0042 |
| XGBoost | 0.554 ± 0.003 | 0.4338 ± 0.0067 | 0.657 ± 0.0041 | 0.1158 ± 0.0065 |
| XGBoost-DistilBERT | 0.5894 ± 0.0063 | 0.4734 ± 0.0038 | 0.6874 ± 0.0028 | 0.0327 ± 0.0116 |
| XGBoost-OpenGPT-2 | 0.5847 ± 0.0101 | 0.4683 ± 0.0139 | 0.6836 ± 0.0101 | 0.0424 ± 0.0088 |
| XGBoost-SentenceTransformer | 0.5892 ± 0.0032 | 0.4749 ± 0.0038 | 0.6883 ± 0.0028 | 0.0312 ± 0.0061 |

Supplementary table 6. Metrics evaluation across models for droplet normalized water inlet

Supplementary table 6 presents a comprehensive evaluation of machine learning models (DNN, LightGBM, SVM, XGBoost) paired with various natural language processing models (DistilBERT, OpenGPT-2, SentenceTransformer), based on performance metrics such as mean absolute error (MAE), mean squared error (MSE), root mean squared error (RMSE), and R². The DNN-DistilBERT model stands out with significantly lower MAE (0.0416) and a high R² (0.9697), indicating excellent predictive performance and minimal error. The standard errors are generally low, ensuring the reliability of these measurements. DNN-OpenGPT-2 and DNN-SentenceTransformer configurations reveal moderate performance. These results highlight the importance of selecting appropriate embeddings to enhance model performance.

2.8 Droplet Orifice Width

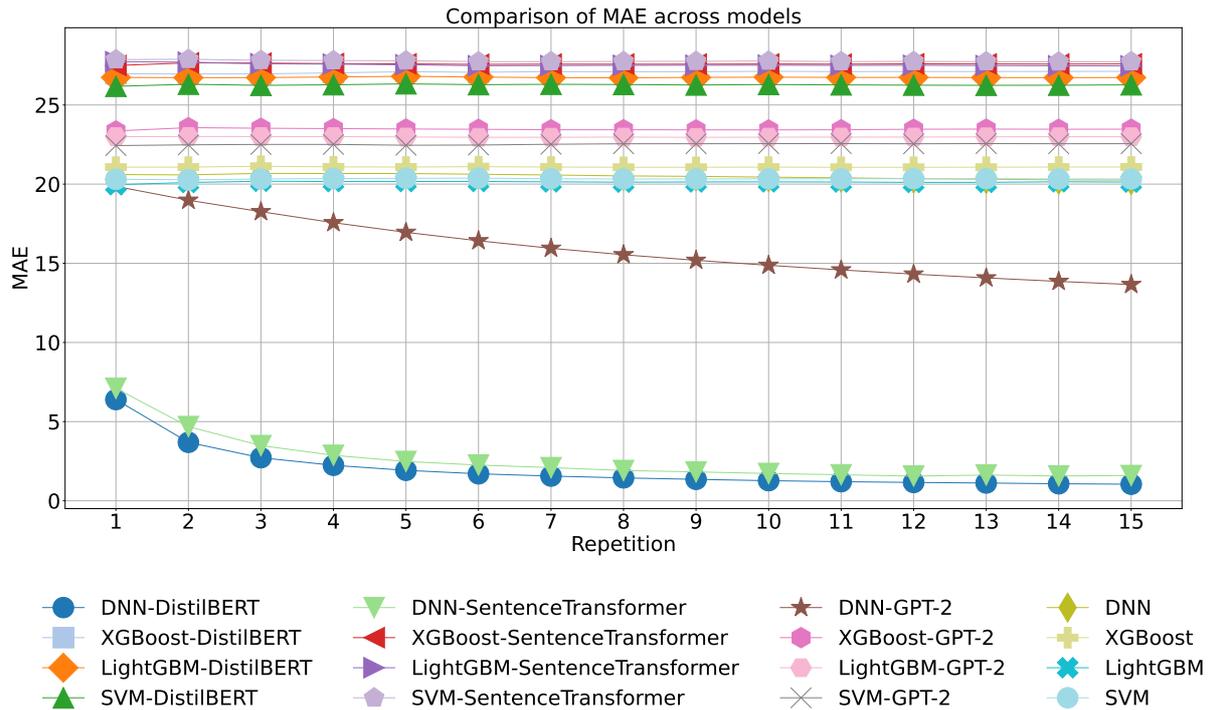

Supplementary Fig.38. Comparison of mean absolute error (MAE) across models for droplet orifice width

Supplementary Fig.38 presents a detailed comparison of 16 models evaluated over 15 iterations, using mean absolute error (MAE) as the performance metric. The deep neural network (DNN) paired with DistilBERT and SentenceTransformer starts at approximately 6 and decreases to about 2 by the 15th repetition, indicating significant improvement. DNN paired with GPT-2 starts at 20 and decreases to below 15, showing moderate improvement. Other models remain relatively stable above 20, suggesting poorer performance.

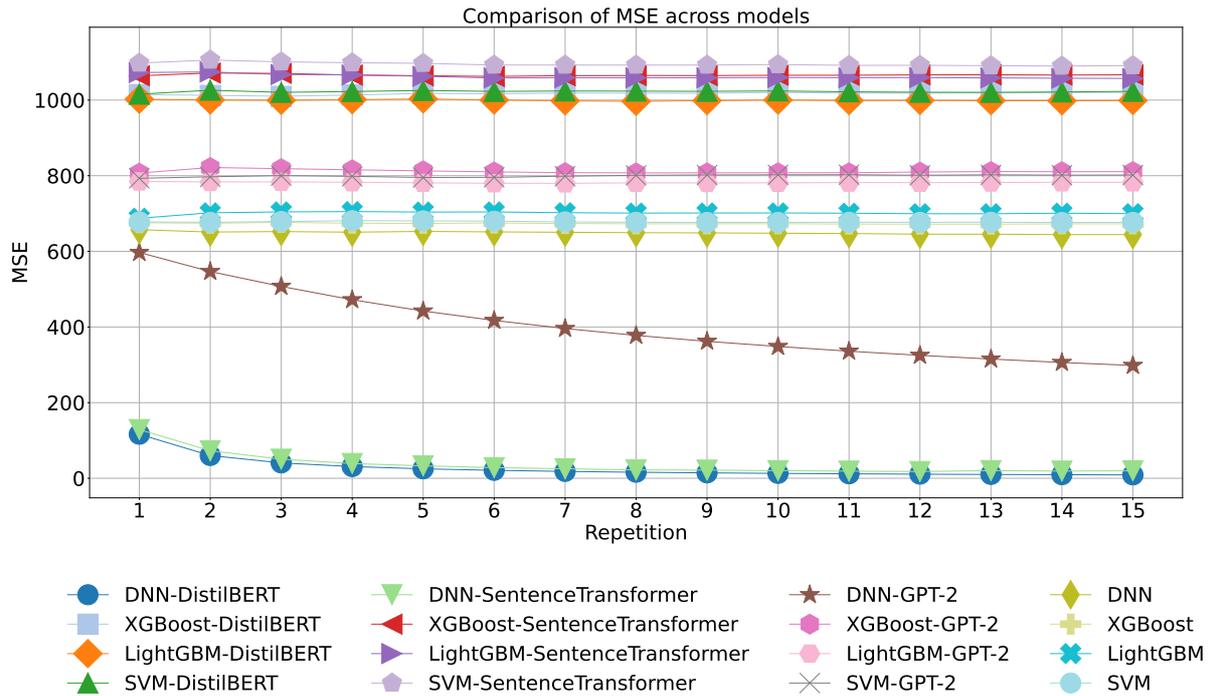

Supplementary Fig.39. Comparison of mean squared error (MSE) across models for droplet orifice width

Supplementary Fig.39 presents a detailed comparison of 16 models evaluated over 15 iterations, using mean squared error (MSE) as the performance metric. The deep neural network (DNN) paired with DistilBERT and SentenceTransformer begins at around 100 and decreases to approximately 20, showing strong performance improvement. DNN paired with GPT-2 starts around 600 and decreases to 300, reflecting moderate improvement. Other models remain stable around above 600, showing little improvement.

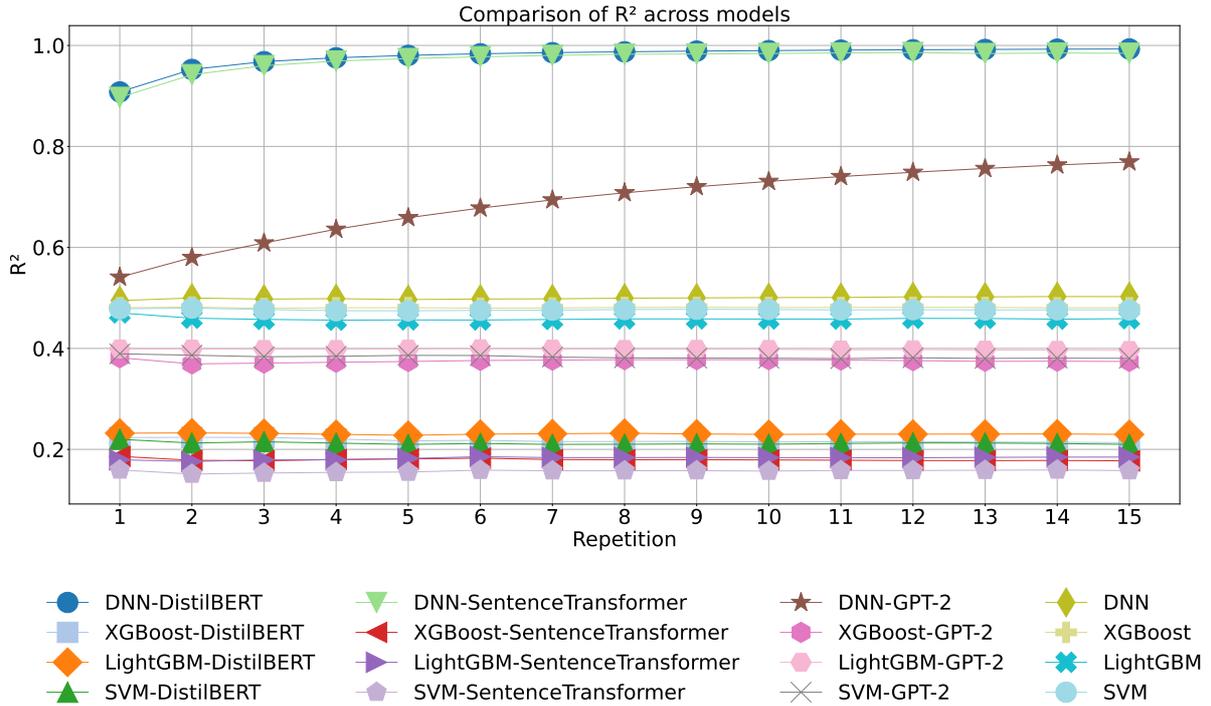

Supplementary Fig.40. Comparison of coefficient of determination (R²) across models for droplet orifice width

Supplementary Fig.40 presents a detailed comparison of 16 models evaluated over 15 iterations, using coefficient of determination ($R^2$) as the performance metric. The deep neural network (DNN) paired with DistilBERT and SentenceTransformer increases from about 0.9 to nearly 1.0, indicating a strong fit over repetitions. DNN paired with GPT-2 increases from around 0.5 to below 0.8, showing moderate improvement. Other models remain relatively stable around 0.2 to 0.5, indicating limited generalization capability.

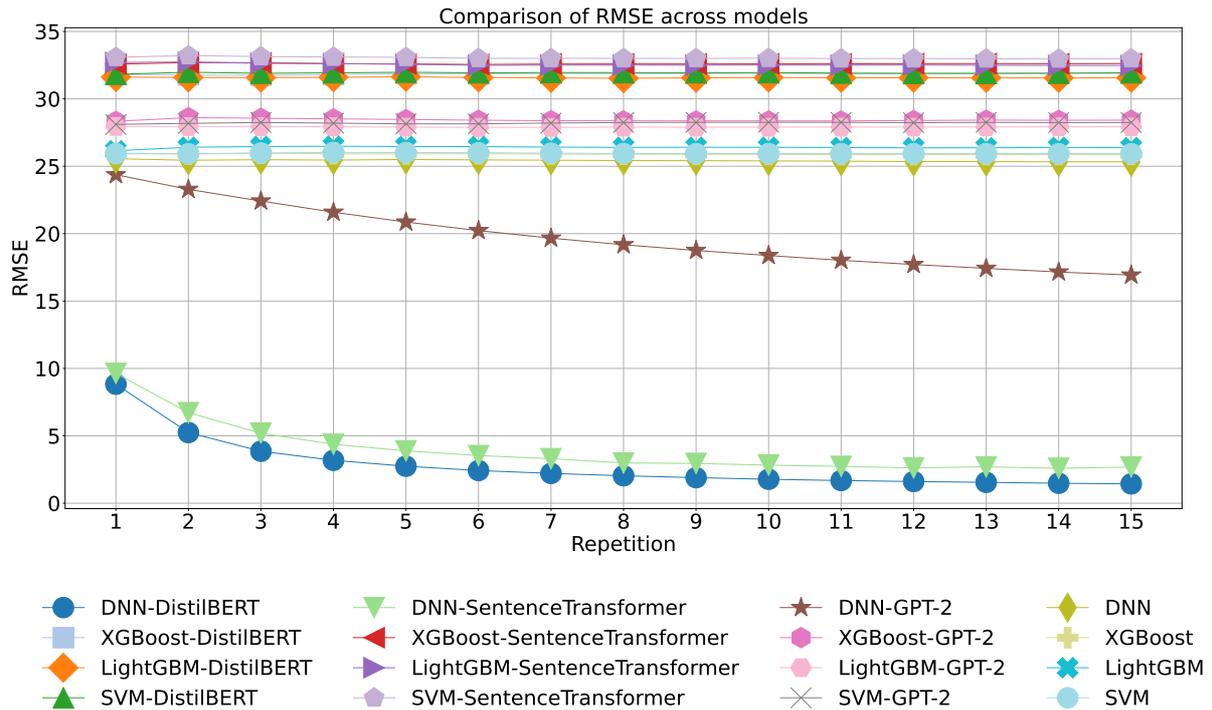

Supplementary Fig.41. Comparison of root mean squared error (RMSE) across models for droplet orifice width

Supplementary Fig.41 presents a detailed comparison of 16 models evaluated over 15 iterations, using root mean squared error (RMSE) as the performance metric. The deep neural network (DNN) paired with DistilBERT and SentenceTransformer decreases from around 10 to about 2, showing high predictive accuracy improvement. DNN paired with GPT-2 decreases from 25 to below 20, indicating moderate accuracy improvement. Other models remain around 30, showing less improvement.

Overall, DNN paired with DistilBERT and SentenceTransformer shows the best performance across all metrics. DNN paired with GPT-2 shows moderate enhancement. Other models remain less accurate.

| Droplet Orifice Width | | | | |
|---|---|---|---|---|
| Model | Metrics | | | |
| | MAE | MSE | RMSE | R² |
| DNN | 20.6718 ± 0.187 | 644.365 ± 7.0814 | 25.4474 ± 0.2044 | 0.4977 ± 0.0122 |
| DNN-DistilBERT | 1.0451 ± 0.159 | 8.8175 ± 4.0448 | 1.4343 ± 0.2123 | 0.993 ± 0.0032 |
| DNN-OpenGPT-2 | 15.54 ± 0.2969 | 546.5903 ± 21.9155 | 23.2854 ± 0.468 | 0.58 ± 0.0167 |
| DNN-SentenceTransformer | 1.6437 ± 0.2239 | 17.8457 ± 5.7285 | 3.3031 ± 0.4495 | 0.9859 ± 0.0046 |
| LightGBM | 20.1259 ± 0.1836 | 701.2015 ± 12.0877 | 26.4633 ± 0.2712 | 0.4585 ± 0.0092 |
| LightGBM-DistilBERT | 26.8097 ± 0.215 | 999.263 ± 18.4334 | 31.5847 ± 0.2024 | 0.2297 ± 0.0055 |
| LightGBM-OpenGPT-2 | 23.007 ± 0.3073 | 779.87 ± 9.9273 | 27.9572 ± 0.5699 | 0.3977 ± 0.0059 |
| LightGBM-SentenceTransformer | 27.5022 ± 0.1702 | 1059.647 ± 11.5529 | 32.5039 ± 0.1772 | 0.1847 ± 0.006 |
| SVM | 20.3261 ± 0.1363 | 677.1053 ± 9.1912 | 25.9802 ± 0.3626 | 0.4775 ± 0.0087 |
| SVM-DistilBERT | 26.2719 ± 0.1412 | 1024.2027 ± 12.0929 | 31.8909 ± 0.1632 | 0.2118 ± 0.0111 |
| SVM-OpenGPT-2 | 22.5483 ± 0.1602 | 798.3176 ± 16.118 | 28.2438 ± 0.2892 | 0.3815 ± 0.0083 |
| SVM-SentenceTransformer | 27.7382 ± 0.2257 | 1105.8276 ± 25.6404 | 32.9875 ± 0.1775 | 0.158 ± 0.0072 |
| XGBoost | 21.0796 ± 0.1929 | 674.9371 ± 11.185 | 25.9121 ± 0.2222 | 0.4811 ± 0.007 |
| XGBoost-DistilBERT | 26.9642 ± 0.2995 | 1020.3485 ± 8.439 | 31.8687 ± 0.2112 | 0.2235 ± 0.0148 |
| XGBoost-OpenGPT-2 | 23.4776 ± 0.1405 | 807.3917 ± 10.2222 | 28.3806 ± 0.1931 | 0.3762 ± 0.0085 |
| XGBoost-SentenceTransformer | 27.5833 ± 0.1428 | 1064.3121 ± 36.653 | 32.5731 ± 0.575 | 0.1787 ± 0.0073 |

Supplementary table 7. Metrics evaluation across models for droplet orifice width

Supplementary table 7 presents a comprehensive evaluation of machine learning models (DNN, LightGBM, SVM, XGBoost) paired with various natural language processing models (DistilBERT, OpenGPT-2, SentenceTransformer), based on performance metrics such as mean absolute error (MAE), mean squared error (MSE), root mean squared error (RMSE), and R². Notably, the DNN-DistilBERT model demonstrates superior performance with a remarkably low MAE of 1.0451 and an R² of 0.993, indicating high predictive accuracy. In contrast, other models, especially when enhanced with SentenceTransformer embeddings, exhibit higher error rates and significantly lower R² values. The LightGBM and SVM models, when combined with DistilBERT, show suboptimal results, with high MSE and RMSE values. These results highlight the importance of selecting appropriate embeddings to enhance model performance.